\documentclass[11pt,letterpaper]{article}

\usepackage[T1]{fontenc}
\usepackage{newtxtext}
\usepackage[scaled=0.92]{helvet}
\usepackage{newtxmath}
\usepackage[margin=1in]{geometry}
\usepackage{microtype}
\usepackage{amsmath}
\usepackage{graphicx}
\usepackage{float}
\usepackage{booktabs}
\usepackage{xcolor}
\usepackage[numbers,sort&compress]{natbib}
\usepackage[font=small,labelfont=bf,labelsep=period]{caption}
\usepackage{tikz}
\usetikzlibrary{arrows.meta,positioning,fit,calc}
\usepackage[colorlinks=true,linkcolor={blue!50!black},citecolor={blue!50!black},urlcolor={blue!50!black}]{hyperref}

\graphicspath{{figures/}}
\input{generated/numbers.tex}

\newcommand{\costmark}{\ensuremath{\mathsf{C}}}
\newcommand{\speedmark}{\ensuremath{\mathsf{S}}}
\newcommand{\bestmark}{\ensuremath{\star}}
\newcommand{\reftier}{\ensuremath{^{\dagger}}}

\title{\Large\textbf{Watermarks and Fingerprints as Soft Bindings for Content Provenance:}\\[2pt] \large\textbf{An Open-Licence Benchmark for Images, Audio and Video}}
\author{Seyedmahdi Kazempourradi\thanks{Corresponding author: \href{mailto:mahdi@originalpictures.com}{\texttt{mahdi@originalpictures.com}}} \and Ramtin Mojtahedi \and Behrang Mohseni\\[4pt]
\normalsize Original Pictures Technologies, Inc., Delaware, USA}
\date{October 2026}

\begin{document}
\maketitle

\begin{abstract}
Content-provenance standards such as C2PA let a platform recover a stripped manifest through a soft binding: an invisible watermark read from the content, or a fingerprint looked up in a registry. We benchmarked both families under one protocol, restricted to openly available models whose licences we audited, on public media, with false-match rates calibrated on held-out negatives and source-level bootstrap intervals for performance estimates. For watermarking we evaluated \val{wm:count/image} image, \val{wm:count/audio} audio and \val{wm:count/video} video configurations from \val{wm:count/families} methods on perceptual quality, robustness, false positives and cost; for fingerprinting, \val{fp:count/methods} methods on registries of up to \val{fp:count/image/nref} images, partial edits and adversarial attacks. PixelSeal gave the best balance for image and video watermarks and AudioSeal for audio, but the error-correcting detector of every TrustMark variant fired on \val{wm:tm/blind/min} to \val{wm:tm/blind/max}\% of unmarked images, so a verifier should test the expected payload. Among fingerprints, copy detectors trained on non-commercial data detected up to \val{fp:image/ISC21-1st/TPR@pair@1e-07/pct}\% of transformed images at a pair-level false-match rate of $10^{-7}$ and DINOv2, the best permissively licensed method, \val{fp:image/DINOv2-B/TPR@pair@1e-07/pct}\%; near-copies in a product catalogue dominated the false matches, and no detection improvement from geometric verification was observed under a matched calibration false-binding constraint in the evaluated image pipelines. On the same attacked copies the two families failed differently: the union of watermark and fingerprint successes covered \val{fp:wm/image/DINOv2-S/either/pct}\% of image copies; fingerprints covered more audio queries, whereas expected-key watermark verification covered more video queries. Embedding a watermark moved the ISCC code of \val{fp:wm/image/ISCC-Image-64/drift/pct}\% of images past its match threshold, and platform-dependent colour conversion changed watermark bits between x86 and ARM hosts.

\end{abstract}


\section{Introduction}
\label{sec:introduction}
Content provenance standards attach a signed record of origin and edits to a media file. The C2PA specification \cite{c2pa2026spec} binds that record to the file cryptographically, so the binding holds only as long as the bytes do. Re-encoding, resizing, a screenshot or a platform's upload pipeline strips or invalidates the manifest, and the content then circulates without its history. C2PA therefore defines \emph{soft bindings}: identifiers derived from the content that let anyone holding a copy look the manifest up again. A soft binding is either an invisible \emph{watermark}, which carries a key inside the pixels or samples, or a \emph{fingerprint}, which is computed from the content and matched against a registry. Regulation adds a second reason to put a signal into the content itself: Article~50(2) of the EU Artificial Intelligence Act requires providers of generative systems to mark synthetic audio, images, video and text in a machine-readable, detectable form, with solutions that are effective, interoperable, robust and reliable as far as technically feasible \cite{eu2024aiact, eu2026omnibus}. Neither the standard nor the regulation prescribes a method or a performance threshold.

The two families are complementary. A watermark answers directly and needs no index, but it must be embedded before release, it can be removed, and it says nothing about content that was never marked. A fingerprint needs no change to the content and applies to an existing archive, but every lookup risks binding a query to the wrong asset, and that risk grows with the registry: a service holding millions of assets needs false-match rates per comparison of $10^{-7}$ or lower. For both families, the rate at which unmarked or unregistered content is reported as a match matters as much as robustness, because a false match attaches someone else's provenance to a piece of content. That rate depends on the decision rule a verifier applies, not only on the model.

Choosing either kind of soft binding is therefore an empirical question, and the published evidence is fragmented. Each new watermark is reported against baselines of its authors' choosing, at its own embedding strength and with its own metrics \cite{bui2025trustmark, xu2025invismark, soucek2025pixelseal, sanroman2024audioseal}; watermark benchmarks cover one modality each \cite{an2024waves, lu2025vine, liu2024audiomarkbench, jiang2025videomarkbench}. Copy-detection benchmarks rank learned descriptors by average precision without a fixed false-match rate \cite{douze2021isc}, perceptual-hash studies measure robustness to benign edits or, separately, to adversaries \cite{jain2022adversarial, prokos2023squint}, and the one open fingerprint on the C2PA soft-binding list, the ISCC of ISO 24138 \cite{iso24138}, has not been compared with either. Licences decide the matter for many adopters and differ between a paper and its release: the strongest learned image descriptors ship with weights trained on non-commercial data, and some watermark checkpoints carry no licence at all. Finally, the two families are evaluated separately, although a provenance service would deploy them together.

This paper benchmarks openly available watermarks and fingerprints for images, audio and video under one protocol, designed around the decisions a provenance service must make. Our contributions are as follows.
\begin{itemize}
\item A watermarking track of \val{wm:count/image} image, \val{wm:count/audio} audio and \val{wm:count/video} video configurations from \val{wm:count/families} methods on public, licence-audited media, measuring perceptual quality with display-aware metrics, robustness to \val{wm:count/transforms} routine transformations with worst-subset reporting, payload recovery, false positives under blind and expected-payload decision rules, and embedding latency and estimated cloud cost (Sections~\ref{sec:wm-method} and~\ref{sec:wm-results}).
\item A fingerprinting track of \val{fp:count/image/methods} image, \val{fp:count/audio/methods} audio and \val{fp:count/video/methods} video methods with \val{fp:count/attacks} transformations, a registry of about a hundred thousand images, and thresholds fixed on held-out negatives at query-level and registry-size-independent pair-level false-match rates down to $10^{-7}$; a partial-edit track with localization; a test of second-stage geometric verification; and white-box, transfer, forgery and inversion attacks (Sections~\ref{sec:fp-method} and~\ref{sec:fp-results}).
\item A measurement of how the two families complement each other on identical media and attacks, and of how embedding a watermark moves fingerprints (Section~\ref{sec:together}).
\item Statistics that treat the source item, not the transformed copy, as the independent unit, and two reproducibility findings: detector outputs that change between x86 and ARM hosts because colour conversion differs, and benchmark inputs that must be pinned beyond the corpus itself.
\end{itemize}
Section~\ref{sec:related} reviews related work, Section~\ref{sec:design} describes the design shared by both tracks and each track's protocol, Sections~\ref{sec:wm-results}--\ref{sec:together} report the results, and Section~\ref{sec:discussion} discusses what they imply for a C2PA soft-binding service.

\section{Background and Related Work}
\label{sec:related}

\subsection{Soft Bindings, Provenance Standards and Regulation}
The C2PA Technical Specification 2.4 \cite{c2pa2026spec} binds a manifest to its asset with a cryptographic \emph{hard binding} and optionally with one or more \emph{soft bindings}: content identifiers that are either not statistically unique, such as a fingerprint, or embedded as an invisible watermark (Sec.~2.3.13). A fingerprint is defined as ``a set of inherent properties computable from digital content that identifies the content or near duplicates of it'' (Sec.~2.4.2). A soft binding is recorded in a \texttt{c2pa.soft-binding} assertion carrying an algorithm identifier, the value and optional scope (Sec.~18.10), and it lets a validator that holds content without a manifest query a repository through the Soft Binding Resolution API \cite{c2pa2026softbindingapi}. Version 2.4 states that the algorithm \emph{should} be one named in the C2PA soft-binding algorithm list (Sec.~9.3.2) and that a soft binding shall not be used as a hard binding (Sec.~9.3.1). At the snapshot we used (commit \texttt{5c864efb}, 16 September 2026) the list held \val{fp:c2pa/list_total} entries, \val{fp:c2pa/list_watermark} watermarks and \val{fp:c2pa/list_fingerprint} fingerprints \cite{c2pasoftbindinglist}. Of the fingerprints, only ISCC (\texttt{io.iscc.v0}) is an open, multi-modal standard \cite{iso24138}; the others are proprietary services or, in one case, a cryptographic digest registered as a soft binding. Collomosse and Parsons describe the combination of manifest, watermark and fingerprint as the three pillars of durable provenance \cite{collomosse2024durable}, and Adobe's deep image comparator \cite{black2021dic} and SImProv \cite{black2022simprov} combine retrieval with localization of the edited region; neither is released as code.

Article~50(2) of the EU Artificial Intelligence Act requires providers of systems that generate synthetic audio, image, video or text to ensure that outputs are marked in a machine-readable format and detectable as artificially generated or manipulated. The technical solutions must be effective, interoperable, robust and reliable as far as technically feasible \cite{eu2024aiact}. The Digital Omnibus on AI, in force since 27 July 2026, postponed this marking obligation to 2 December 2026 for systems placed on the market before 2 August 2026 \cite{eu2026omnibus}. The Act does not name a technique or set thresholds. Imperceptibility, survival under processing, false-positive rate and cost are therefore left to evidence of the kind a comparative benchmark provides.

\subsection{Invisible Watermarking}
\paragraph{Images.} Cox et al.\ introduced spread-spectrum watermarking for multimedia, spreading a pseudo-random sequence over perceptually significant transform coefficients so that the mark cannot be removed without degrading the host \cite{cox1997secure}. Later transform-domain schemes embedded the mark in the DCT domain with a detector that does not need the original image \cite{barni1998dct}, or combined the discrete wavelet transform with singular value decomposition \cite{ganic2004dwtsvd}. The DWT-DCT and DWT-DCT-SVD encoders of the invisible-watermark library \cite{invisiblewatermark} are open implementations in this family. Their documentation warns that decoding is not guaranteed even on unmodified images.

Deep watermarking replaced hand-designed transforms with learned encoders and decoders. HiDDeN trained both networks jointly and showed that the encodings can be made robust to noise applied between them \cite{zhu2018hidden}. StegaStamp extended the idea to printed and re-photographed images \cite{tancik2020stegastamp}, and MBRS handled non-differentiable JPEG compression by mixing real and simulated compression within each mini-batch \cite{jia2021mbrs}. RivaGAN added an attention-based embedder and two adversarial critics, one for visual quality and one for robustness, and was designed for video \cite{zhang2019rivagan}. RoSteALS moved embedding into the latent space of a frozen pretrained autoencoder \cite{bui2023rosteals}. TrustMark, a GAN-based method with spatio-spectral losses, targets images of arbitrary resolution and adds a network that removes its own watermark for re-watermarking \cite{bui2025trustmark}. InvisMark was designed for high-resolution generated images and reports a PSNR near 51\,dB, an SSIM near 0.998 and 256-bit payloads \cite{xu2025invismark}. The Watermark Anything Model (WAM) segments an image into watermarked and unwatermarked regions and recovers several 32-bit messages from small areas \cite{sander2025wam}. Pixel Seal replaces pixel-wise perceptual losses with adversarial-only training, stabilizes it with a three-stage schedule, and applies just-noticeable-difference attenuation at high resolution \cite{soucek2025pixelseal}. Petrov et al.\ derived upper bounds on watermark capacity under PSNR and linear robustness constraints and trained ChunkySeal, a scaled version of Video Seal that carries 1024 bits \cite{petrov2025chunkyseal}.

A separate line embeds the mark during generation. Stable Signature, for example, fine-tunes the latent decoder of a diffusion model, conditioned on a binary signature, so that every generated image carries it \cite{fernandez2023stable}. Such methods cannot mark existing content, so our study covers post-hoc methods only. Each of the papers above evaluates its method against baselines of its own choosing, on its own images, at its own embedding strength and with its own selection of metrics, mostly PSNR and SSIM. Their reported numbers are therefore not directly comparable.

\paragraph{Video.} Hartung and Girod applied spread-spectrum embedding to both uncompressed and compressed video \cite{hartung1998video}. Among learned methods, DVMark uses a multiscale network for video watermarking \cite{luo2025dvmark}, and RivaGAN operates on frame sequences \cite{zhang2019rivagan}. Video Seal trains an embedder and an extractor with video codecs among the training transformations. It also propagates the watermark temporally, so that an image model can be applied without watermarking every high-resolution frame \cite{fernandez2024videoseal}. Pixel Seal adapts to video by pooling the watermark over time \cite{soucek2025pixelseal}. In deployment, video is usually re-encoded with H.264/AVC \cite{wiegand2003h264} or H.265/HEVC \cite{sullivan2012hevc}, and inter-frame prediction together with coarse quantization can attenuate low-energy residuals. Few video watermarking models have openly licensed weights, and independent evidence is thin: VideoMarkBench, the dedicated video benchmark discussed below, uses clips produced by video generators rather than camera footage \cite{jiang2025videomarkbench}.

\paragraph{Audio.} Bender et al.\ surveyed early data-hiding techniques for images and audio \cite{bender1996techniques}, and Kirovski and Malvar adapted spread-spectrum watermarking to audio signals \cite{kirovski2003spread}. WavMark encodes up to 32 bits in each second of audio and reports a mean bit error rate of 0.48\% over ten attacks on 10-20\,s clips \cite{chen2023wavmark}. AudioSeal trains a generator and a detector with a localization loss for sample-level detection and uses a perceptual loss inspired by auditory masking; its single-pass detector is intended for large-scale screening \cite{sanroman2024audioseal}. SilentCipher adds psychoacoustic thresholding and pseudo-differentiable compression layers and operates on 44.1\,kHz audio \cite{singh2024silentcipher}. Perth is an open library whose main watermarker is a neural implicit model. Its documentation claims survival of compression and resampling, but no peer-reviewed evaluation is available \cite{perth2025}. WavMark and AudioSeal were motivated by voice cloning and evaluated mainly on speech, and the four methods use different native sample rates, so a shared protocol has to fix resampling and codec settings before their numbers can be compared.

\subsection{Content Fingerprinting}
\paragraph{Perceptual hashes.} Classical perceptual hashes reduce an image to a short binary code that changes little under common processing. The average and difference hashes compare block means \cite{krawetz2011looks, krawetz2013kind}; pHash thresholds low-frequency DCT coefficients at their median \cite{zauner2010phash}; the Marr--Hildreth and block-mean hashes use edge and block statistics \cite{zauner2010phash, yang2006blockhash}, and implementations are collected in imagehash \cite{buchner_imagehash}, Blockhash \cite{commonsmachinery_blockhash} and OpenCV's \texttt{img\_hash} module \cite{opencv_imghash}. Meta's PDQ applies a DCT to a filtered 64$\times$64 luminance image and keeps 256 bits together with a quality score \cite{meta2019pdq}; it is deployed in industry hash-sharing programmes. The ISCC Image-Code of ISO 24138 is a DCT hash of a 32$\times$32 greyscale image after trimming uniform borders \cite{iso24138}. The ISCC benchmark TwinSpect \cite{iscc_twinspect} evaluates such compact codes, but to our knowledge no study has compared them with learned descriptors under a common false-positive calibration.

\paragraph{Learned copy-detection descriptors.} The Image Similarity Challenge 2021 and its DISC21 dataset established copy detection as a retrieval problem with a micro-average precision metric over one million references \cite{douze2021isc}. SSCD trains a descriptor with self-supervised contrastive learning, entropy regularisation and heavy augmentation, and outperformed classification-oriented self-supervised features by a wide margin \cite{pizzi2022sscd}; the winning descriptor-track entry used a large memory bank and negative-embedding subtraction \cite{yokoo2021isc}. General-purpose embeddings such as DINOv2 \cite{oquab2024dinov2} and CLIP \cite{radford2021clip, cherti2023openclip} capture semantics rather than copy identity, which helps under heavy geometric change but risks matching different photographs of the same object. ISCC's experimental semantic image code binarises a descriptor derived from the ISC21 winner \cite{iscc_sci}, and DINOHash adversarially fine-tunes DINOv2 into a 96-bit hash \cite{singhi2025dinohash}. Local features with geometric verification, from SIFT \cite{lowe2004sift} and ORB \cite{rublee2011orb} with RANSAC \cite{fischler1981ransac} to DISK \cite{tyszkiewicz2020disk} with LightGlue \cite{lindenberger2023lightglue}, remain the standard way to confirm a partial copy and to register a query onto its source.

\paragraph{Video and audio fingerprints.} For video, the MPEG-7 video signature \cite{paschalakis2012mpeg7} underlies the ISCC Video-Code, Meta's TMK+PDQF pools per-frame PDQ features with a temporal match kernel \cite{poullot2015tmk, meta2019pdq}, and vPDQ matches clips by the share of frames with a close PDQ hash \cite{meta_vpdq}. The 2023 Video Similarity Challenge showed that frame-level learned descriptors with temporal alignment dominate for copied segments \cite{pizzi2024vsc}. Audio identification is mature: landmark hashing \cite{wang2003shazam} and sub-band energy fingerprints \cite{haitsma2002robust} underlie tools such as audfprint \cite{ellis_audfprint}, Olaf \cite{six2023olaf}, Panako \cite{six2022panako} and Chromaprint \cite{lalinsky_chromaprint}, on which the ISCC Audio-Code is built. Neural audio fingerprints learn segment embeddings with contrastive objectives \cite{chang2021nafp, araz2025nmfp}.

\subsection{Benchmarks, Attacks and Security}
Petitcolas et al.\ catalogued attacks on early copyright marking systems and argued that such systems need common evaluation \cite{petitcolas1998attacks}. Recent benchmarks were motivated by generative models. WAVES evaluates image watermarks on detection and identification under conventional distortions, regeneration by diffusion models and adversarial attacks \cite{an2024waves}. W-Bench tests eleven image watermarking methods against regeneration, global and local editing and image-to-video generation, and proposes VINE, a watermark built on a pretrained diffusion model \cite{lu2025vine}. MarkDiffusion is a toolkit for in-generation watermarking of latent diffusion models, with shared evaluation pipelines for detectability, robustness and output quality \cite{pan2025markdiffusion}. AudioMarkBench evaluates three audio watermarking methods against 15 perturbations on a multilingual speech dataset under no-box, black-box and white-box threat models \cite{liu2024audiomarkbench}. VideoMarkBench evaluates four video watermarking methods against 12 perturbations on clips produced by three video generators \cite{jiang2025videomarkbench}.

Attack papers bound what any benchmark can claim. Zhao et al.\ proved that regeneration attacks, which add noise and then reconstruct the image, remove pixel-level invisible watermarks \cite{zhao2024removable}. Jiang et al.\ showed that small adversarial perturbations evade watermark detection with less visible change than JPEG compression or blurring \cite{jiang2023evading}. Our watermarking track measures robustness to routine processing (codecs, resizing, cropping, filtering and resampling), that is, survival of ordinary distribution rather than security against a motivated adversary; the fingerprinting track adds white-box, transfer, forgery and inversion attacks (Section~\ref{sec:fp-res-security}).

Perceptual hashes were designed for benign processing, not for an adversary. Gradient attacks against Apple's NeuralHash produced collisions and evasions with imperceptible perturbations \cite{struppek2022neuralhash}; black-box attacks evaded PDQ and PhotoDNA and forged targeted collisions \cite{jain2022adversarial, prokos2023squint, dolhansky2020collision}; and inversion attacks reconstruct recognisable images from PDQ, PhotoDNA and NeuralHash codes \cite{hawkes2025inversion, madden2024robustness}. The last result matters for provenance: a soft-binding value is written into the manifest, so a descriptor that can be inverted discloses content to anyone who reads the manifest.

Following standard practice for binomial rates and resampled statistics, we report exact Clopper-Pearson intervals \cite{clopper1934} and bootstrap intervals \cite{efron1979bootstrap}.

\subsection{Watermarks and Fingerprints Together}
Invisible watermarks \cite{bui2025trustmark, sanroman2024audioseal, fernandez2024videoseal} carry a key that points to the manifest directly, while a fingerprint needs a registry lookup but survives when the watermark is removed or was never embedded. Watermarks that localize edits \cite{zhang2024editguard, zhang2025omniguard} and blind forensic detectors \cite{guillaro2023trufor} address tampering without a reference. Prior benchmarks evaluate each family on its own; we measure their interaction on the same media and attacks.

\subsection{Perceptual Quality Metrics}
Most watermarking papers report PSNR and SSIM \cite{wang2004ssim}, and some add MS-SSIM \cite{wang2003msssim} or the learned LPIPS distance \cite{zhang2018lpips}. These metrics summarize an average difference. Watermark residuals, however, are often structured patterns that a viewer notices in flat regions or when switching between the original and the marked image. FLIP models this second case: it estimates the differences an observer perceives when flipping between two images \cite{andersson2020flip}. ColorVideoVDP models spatial and temporal vision for luminance and colour and accounts for viewing conditions and display characteristics \cite{mantiuk2024cvvdp}. VMAF is a perceptual video quality metric in common use for encoder tuning \cite{rassool2017vmaf}. The Pixel Seal authors argue that proxy losses such as MSE and LPIPS fail to predict visible artefacts \cite{soucek2025pixelseal}, which is one reason to report display-aware metrics next to PSNR.

For audio, SNR and scale-invariant SNR \cite{leroux2019sdr} measure waveform error. PESQ \cite{rix2001pesq,itu2001p862} and its wideband extension \cite{itu2007p8622} predict speech quality. The ITU withdrew both recommendations in January 2024 in favour of P.863, but they are still used, which keeps results comparable with earlier watermarking papers. PESQ was designed for telephone speech and does not apply to music. STOI predicts intelligibility \cite{taal2011stoi}. Integrated loudness under ITU-R BS.1770 \cite{itu2023bs1770} and EBU R~128 \cite{ebu2023r128} detects level changes that waveform metrics do not isolate.

\section{Benchmark Design}
\label{sec:design}
The benchmark has two tracks, one per family of soft binding, that share their principles, their hosts and, for the combination experiment, their media. This section states what they share and then describes each track's protocol.

\subsection{Shared Principles}
\label{sec:principles}
\paragraph{Licence audit.} A provenance service can deploy only models whose code \emph{and} weights permit commercial use, and the two licences differ in several projects; the weights may also have been trained on data restricted to non-commercial use. We recorded all three for every method. Methods that fail the audit were measured as reference points and are marked \dag\ (watermarks) or \reftier\ (fingerprints) in every table; they are excluded from rankings and recommendations.

\paragraph{Public, pinned media.} All media are public and were fetched from their distribution points; every file is pinned by SHA-256 in a manifest. No private or personal media were used. All randomness (payloads, transformations, splits) is seeded, so every method sees byte-identical inputs.

\paragraph{False matches are controlled, not just reported.} A soft binding is useful only if it rarely points unmarked or unregistered content at someone else's manifest. Both tracks therefore fix the decision rule before measuring robustness. For watermarks, the rule is either the model's blind detector or an expected-payload test that requires $k$ of $n$ decoded bits to match the key the verifier expects, with $k$ chosen for a per-key false-match probability of at most $10^{-6}$ under independent fair bits. For fingerprints, thresholds are fixed on one half of the never-registered negatives, at query-level and at registry-size-independent pair-level false-match rates, and evaluated on the other half; a target is reported only if the calibration set resolves it. Detection is always reported at such an operating point, next to threshold-free measures.

\paragraph{The source is the unit.} Every image, clip or track contributes one query per transformation, and the transformed copies of one source succeed or fail together far more often than independent queries would. Intervals therefore come from bootstraps over sources (for fingerprint retrieval rates and paired differences, with the threshold recomputed from the resampled calibration set in every replicate), exact intervals for watermark false positives count sources rather than trials, and paired comparisons pair by source. For audio watermarks, whose clips were rendered at three sample rates, the clip is the source.

\paragraph{Hosts.} GPU work ran on NVIDIA A10G instances (AWS g5, AMD EPYC 7R32) and CPU work on the same processor family; latency is reported per network on an idle host. Because the watermarking track found host-dependent decoder outputs, both tracks were also run or decoded on an Apple M5 Pro (arm64) laptop to test determinism.

\subsection{Watermarking Protocol}
\label{sec:wm-method}
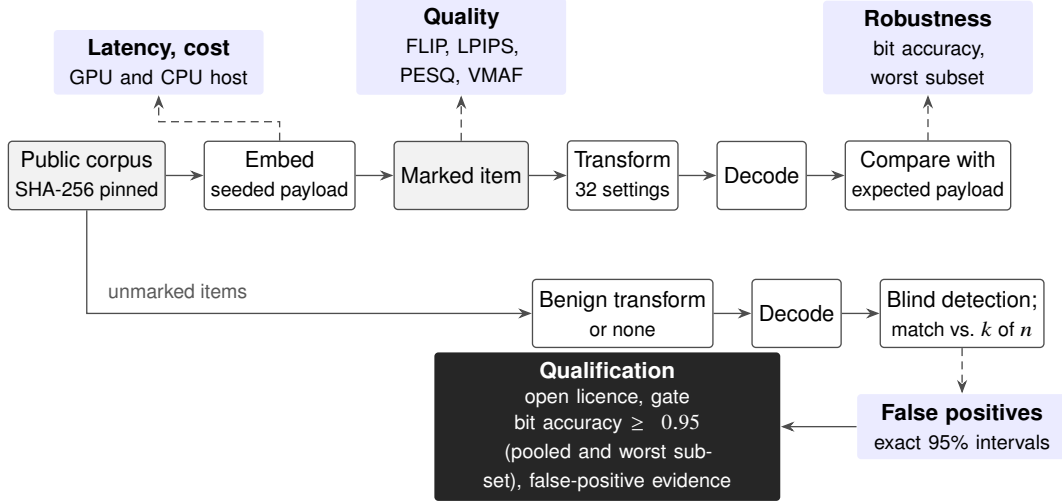
\begin{figure*}[!t]
\centering
\begin{tikzpicture}[
  font=\sffamily\footnotesize,
  node distance=4.5mm and 5mm,
  box/.style={draw=black!70, line width=0.5pt, rounded corners=1.5pt, align=center, minimum height=9mm,
              inner xsep=2.5pt, fill=white},
  data/.style={box, fill=black!5},
  metric/.style={box, draw=none, fill=blue!8, text width=26mm, minimum height=8mm},
  arr/.style={-{Stealth[length=1.8mm]}, line width=0.5pt, draw=black!70},
  lab/.style={font=\sffamily\scriptsize, text=black!65, align=center},
]
\node[data] (src) {Public corpus\\\scriptsize SHA-256 pinned};
\node[box, right=of src] (emb) {Embed\\\scriptsize seeded payload};
\node[data, right=of emb] (marked) {Marked item};
\node[box, right=of marked] (att) {Transform\\\scriptsize \val{wm:count/transforms} settings};
\node[box, right=of att] (dec) {Decode};
\node[box, right=of dec] (cmp) {Compare with\\\scriptsize expected payload};

\draw[arr] (src) -- (emb);
\draw[arr] (emb) -- (marked);
\draw[arr] (marked) -- (att);
\draw[arr] (att) -- (dec);
\draw[arr] (dec) -- (cmp);

\node[box, below=9mm of att] (att0) {Benign transform\\\scriptsize or none};
\node[box, right=of att0] (dec0) {Decode};
\node[box, right=of dec0] (fp) {Blind detection;\\\scriptsize match vs.\ $k$ of $n$};
\draw[arr] (src.south) |- (att0.west);
\draw[arr] (att0) -- (dec0);
\draw[arr] (dec0) -- (fp);
\node[lab, anchor=south west] at ($(src.south |- att0.west)+(1.5mm,0.6mm)$) {unmarked items};

\node[metric, above=6mm of marked] (mq) {\textbf{Quality}\\\scriptsize FLIP, LPIPS, PESQ, VMAF};
\node[metric, above=6mm of cmp] (mr) {\textbf{Robustness}\\\scriptsize bit accuracy, worst subset};
\node[metric, below=6mm of fp] (mf) {\textbf{False positives}\\\scriptsize exact 95\% intervals};
\node[metric, above=6mm of emb, xshift=-16mm] (mt) {\textbf{Latency, cost}\\\scriptsize GPU and CPU host};
\draw[arr, densely dashed] (marked) -- (mq);
\draw[arr, densely dashed] (cmp) -- (mr);
\draw[arr, densely dashed] (fp) -- (mf);
\draw[arr, densely dashed] (emb.north) -- ++(0,2.5mm) -| (mt.south);

\node[box, fill=black!85, text=white, draw=none, left=10mm of mf, text width=44mm] (q)
  {\textbf{Qualification}\\\scriptsize open licence, gate bit accuracy $\ge 0.95$\\\scriptsize (pooled and worst subset), false-positive evidence};
\draw[arr] (mf) -- (q);
\end{tikzpicture}
\caption{Benchmark protocol. Each of the \val{wm:count/image} image, \val{wm:count/audio} audio and \val{wm:count/video} video configurations embeds a seeded payload into every item. Quality is measured on the marked item; robustness on the decoder output after each transformation; false positives on unmarked items under the untouched condition and three benign transformations; latency and cost on a fixed GPU and CPU host. A configuration qualifies for ranking only if its licence, gate robustness and false-positive evidence all pass.}
\label{fig:wm-pipeline}
\end{figure*}

Fig.~\ref{fig:wm-pipeline} summarises the protocol. Each configuration embeds a pseudo-random payload into every corpus item, the marked item is measured against the original, a fixed set of transformations is applied, and the decoder output is compared with the embedded payload. Unmarked items pass through the same decoders to measure false positives. All randomness is seeded, and payloads are derived from a hash of a fixed seed and the item index, so every configuration of a given payload size receives the same message for the same item.

\subsubsection{Methods Under Test}
We selected watermarking methods that are distributed with trained weights and an inference implementation, and that target natural images, speech and music, or natural video. Table~\ref{tab:wm-models} lists them. For each method we recorded the licence of the code and, separately, of the weights, because the two differ in several projects. A configuration counts as \emph{open} only when both carry a licence that permits commercial use. Methods whose weights have no stated licence, or whose weight provenance we could not establish, were still measured and are reported as reference points, but they are excluded from every ranking. Several methods were excluded before measurement: VINE, whose code licence is non-commercial \cite{lu2025vine}; the WAM checkpoint trained on COCO, licensed CC BY-NC \cite{sander2025wam}; and DVMark, for which no weights are released \cite{luo2025dvmark}.

Strength was varied where a method exposes a single scalar that scales the embedded residual: the TrustMark strength factor, the blending factor of the PixelSeal, Video Seal and WAM embedders, and the AudioSeal output scale. We kept each method's default and added settings on both sides of it. For PixelSeal we added two settings between 0.1 and the default 0.2 after an initial pass showed that the robustness threshold is crossed in that interval. All other hyperparameters were left at the values published with the weights.

\begin{table}[!t]
\centering
\footnotesize
\caption{Methods, payload sizes and licences.}
\label{tab:wm-models}
\setlength{\tabcolsep}{3pt}
\begin{tabular}{llrll}
\toprule
Modality & Method & Bits & Code & Weights \\
\midrule
Image & TrustMark B, C, P, Q \cite{bui2025trustmark} & 100 & MIT & MIT \\
Image & PixelSeal \cite{soucek2025pixelseal} & 256 & MIT & MIT \\
Image, video & Video Seal 1.0 \cite{fernandez2024videoseal} & 256 & MIT & MIT \\
Image, video & ChunkySeal \cite{petrov2025chunkyseal} & 1024 & MIT & MIT \\
Image & WAM \cite{sander2025wam} & 32 & MIT & MIT$^{a}$ \\
Image & DWT-DCT, DWT-DCT-SVD \cite{invisiblewatermark} & 64 & MIT & n/a \\
Image & RivaGAN \cite{zhang2019rivagan} & 32 & MIT & unclear\textsuperscript{\dag} \\
Image & InvisMark \cite{xu2025invismark} & 100 & MIT & none stated\textsuperscript{\dag} \\
Audio & AudioSeal base, streaming \cite{sanroman2024audioseal} & 16 & MIT & MIT \\
Audio & WavMark \cite{chen2023wavmark} & 16 & MIT & MIT \\
Audio & SilentCipher \cite{singh2024silentcipher} & 40 & MIT & none stated\textsuperscript{\dag} \\
Audio & Perth \cite{perth2025} & 0 & MIT & MIT \\
\bottomrule
\end{tabular}
\par\vspace{2pt}\parbox{\linewidth}{\raggedright\scriptsize Bits: payload bits carried and compared. TrustMark transmits a 100-bit codeword; in the BCH\_5 mode used here it carries a 61-bit payload, 35 BCH parity bits and 4 version bits. Bit accuracy and the expected-payload test use the 100 transmitted bits; exact recovery compares the 61 decoded payload bits. $^{a}$The MIT-licensed checkpoint trained on SA-1B; the checkpoint trained on COCO is licensed CC BY-NC and was not used. \dag~Reported for reference only. Perth signals presence without a payload. Every checkpoint is pinned by SHA-256 in the lockfile kept with the scripts (Data and Code Availability).}
\end{table}

\subsubsection{Corpus}
\emph{Images} (\val{wm:image/n_items} items): the 24 Kodak images \cite{kodak}, 25 DIV2K validation images \cite{agustsson2017div2k} at about 2K resolution, 28 CLIC 2020 professional validation images \cite{clic}, and 8 CC0 high-dynamic-range panoramas from Poly Haven \cite{polyhaven}, exposure-normalised to a median luminance of 0.18 and stored as 16-bit sRGB TIFF. DIV2K images are used for measurement only and are not redistributed. FLIP was computed on the 60 Kodak, CLIC and HDR images and ColorVideoVDP on the 32 Kodak and HDR images, because both are costly at DIV2K resolution; the other metrics use all 85 images.

\emph{Audio} (\val{wm:audio/n_items} inputs from \val{wm:audio/n_sources} source clips): 40 LibriSpeech test-clean utterances from 40 different speakers \cite{panayotov2015librispeech} and 7 public-domain or CC BY music excerpts distributed with librosa \cite{mcfee2015librosa}, each resampled to 16, 44.1 and 48~kHz. One music excerpt failed in every configuration at all three rates; we treat a clip that fails for every method as a corpus defect and exclude it everywhere, leaving \val{wm:audio/n_sources} source clips (40 speech, 6 music) and 138 inputs, of which \val{wm:audio/n_speech} are speech and \val{wm:audio/n_music} music. The three renderings of a clip are not independent samples; Section~\ref{sec:wm-stats} explains how the statistics account for this.

\emph{Video} (\val{wm:video/n_items} clips): the first 2.5~s of six 1080p sequences from the Xiph test-media collection \cite{xiphderf} (\emph{aspen}, \emph{controlled burn}, \emph{red kayak}, \emph{speed bag}, \emph{crowd run} and the \emph{Sintel} trailer), covering slow and fast motion, fine texture and synthetic content.

\subsubsection{Transformations}
Image transformations (\val{wm:count/transforms/image}): JPEG at quality 90, 75, 60 and 50; WebP at 80 and 60; bicubic downscaling to 0.75 and 0.5; centre crops retaining 90, 70 and 50\% of each side; Gaussian blur with $\sigma=1$; and brightness scaling by 0.8 and 1.2. Audio transformations (\val{wm:count/transforms/audio}): MP3 at 320, 128 and 64~kb/s; AAC at 256, 128 and 64~kb/s; resampling round trips through 22.05 and 8~kHz; gain of $-6$~dB; and additive white noise at 30~dB SNR. Codec round trips are aligned to the original by cross-correlation before decoding, so that encoder delay does not register as watermark loss. A verifier that holds only the received file cannot do this, so a follow-up experiment decoded the same round trips without alignment (Section~\ref{sec:wm-res-audio}). Video transformations (\val{wm:count/transforms/video}): H.264 at CRF 18, 22, 23 and 28; H.265 at CRF 23 and 28; and H.264 at CRF 23 after 0.5 downscaling or a 75\% centre crop. Marked video is stored losslessly (FFV1, 4:4:4) before each transformation, so the transformation is the only lossy step.

One transformation per modality serves as the \emph{gate}: JPEG 75 for images, MP3 at 128~kb/s for audio and H.264 at CRF 23 for video. These are common default settings in sharing pipelines, and a watermark that does not survive them is of little use for provenance.

\subsubsection{Perceptual Quality}
Image quality was measured with PSNR, SSIM \cite{wang2004ssim}, MS-SSIM \cite{wang2003msssim}, LPIPS \cite{zhang2018lpips}, FLIP \cite{andersson2020flip} and ColorVideoVDP \cite{mantiuk2024cvvdp}. We use FLIP as the primary image metric because it models the visibility of differences when flipping between two images, which is the viewing condition under which an embedded pattern is most likely to be noticed. Audio quality was measured with SNR, SI-SNR \cite{leroux2019sdr}, wideband PESQ \cite{itu2007p8622}, STOI \cite{taal2011stoi}, the absolute change in integrated loudness following ITU-R BS.1770 \cite{itu2023bs1770}, and a log-spectral difference. PESQ is primary. It was designed and validated for speech, so the primary PESQ value is the mean over the \val{wm:audio/n_speech} speech inputs only; PESQ on the \val{wm:audio/n_music} music inputs is reported separately and not used for ranking. Video quality was measured with PSNR, SSIM, LPIPS and FLIP averaged over frames, and with VMAF \cite{rassool2017vmaf}, which is primary.

\subsubsection{Robustness}
Robustness is the fraction of payload bits recovered correctly after a transformation (bit accuracy), averaged over items. Because a verifier acts on a decision rather than on a bit rate, we also report two payload-level rates (Table~\ref{tab:wm-exact}): \emph{exact recovery}, the share of items whose decoded payload equals the embedded one, after BCH decoding of its 61-bit payload for TrustMark (the only method with an error-correcting layer) and with every bit correct for the others; and the share of items that pass the expected-payload test of Section~\ref{sec:wm-fp-method} ($\geq k$ of $n$ bits). For video, decoder logits are averaged over all frames of a clip before thresholding, which is the aggregation the Video Seal and PixelSeal decoders are designed for. We report the pooled mean and the mean on the worst content subset (image corpus, audio source and sample rate, or video clip), because a pooled mean can hide systematic failure on one kind of content.

\subsubsection{False Positives}
\label{sec:wm-fp-method}
We distinguish two decision rules. A \emph{blind} detector uses only the model's own output: TrustMark's error-correcting decoder reports whether a valid codeword was found, and the audio models expose a detection score, which we threshold at 0.8 for AudioSeal (whose score is the fraction of audio frames classified as watermarked) and at 0.5 for the others. An \emph{expected-payload} test instead declares a watermark when at least $k$ of $n$ decoded bits equal the payload the verifier expects. Under the null hypothesis that unmarked content yields independent fair bits, the per-key false-match probability is
\begin{equation}
P_{\mathrm{FM}}(k,n) = \sum_{i=k}^{n}\binom{n}{i}2^{-n},
\end{equation}
and we set $k$ to the smallest value with $P_{\mathrm{FM}}\le 10^{-6}$. For 16-bit payloads no such $k$ exists, since even an exact match has probability $2^{-16}\approx1.5\times10^{-5}$; for these methods only the blind detector can be tested.

False-positive trials run each decoder on the unmarked corpus under no transformation and three benign ones (JPEG 75, 0.5 downscale and 70\% crop for images; the untouched signal, MP3 at 128~kb/s and a 22.05~kHz round trip for audio; and, for video, the H.264 CRF~18 encode of each unmarked clip). A verifier checks one fixed expected payload, the one recorded for the asset; comparing each decode with the item's own payload is that check. We also compare each decode with 50 unrelated payloads (200 for video), one fixed list shared by all items. These comparisons test whether decoded bits agree with payloads the content was never marked with, conditional on the decodes we have; they reuse the same decodes and therefore add no independent media, and we report them as counts and as the largest number of matching bits observed, not as additional trials. The transformed copies of one item, and the three renderings of one audio clip, share their source, so observed rates and their exact Clopper-Pearson intervals \cite{clopper1934} use the unmarked source as the trial unit: \val{wm:fpr/image/PixelSeal/own_src_n} images, \val{wm:fpr/audio/WavMark/blind_src_n} audio clips and \val{wm:fpr/video/PixelSeal/own_src_n} video clips. All intervals are two-sided at 95\%. We also plot the distribution of matching bits against the binomial null.

A configuration's false-positive evidence is \emph{analytic} when an expected-payload test with $P_{\mathrm{FM}}\le10^{-6}$ exists and no unmarked trial reached $k$, and \emph{observed} when no such test exists but the blind detector produced no false detection. Our target is a rate of at most $10^{-3}$. Table~\ref{tab:wm-fpr} keeps the two kinds of evidence apart: the theoretical $P_{\mathrm{FM}}$ of each expected-payload test, and the observed counts with their source-level 95\% upper bounds. Analytic evidence meets the target by construction, provided unmarked content decodes to bits that are independent of the expected payload; the match distributions in Section~\ref{sec:wm-res-fpr} are consistent with that assumption but cannot prove it. Observation alone does not establish the target for any configuration: with no false positive, the upper bound falls below $10^{-3}$ only after \val{wm:fpr/sources_for_1e-3} independent sources, far more than our corpus holds.

\subsubsection{Latency and Cost}
\label{sec:wm-cost}
Embedding latency was measured on an AWS g5.xlarge instance (NVIDIA A10G, 24~GB; AMD EPYC 7R32, 4 vCPU; PyTorch 2.14, CUDA 13.0) as the median over all corpus items, excluding the first item of each run (which includes lazy initialisation), with device synchronisation before and after each call. We report milliseconds per megapixel for images and video frames, and milliseconds of compute per second of audio. CPU latency was measured on the same host with four threads, as the median over four images (Kodak and DIV2K) or audio items after one warm-up item. Model loading time is excluded.

Strength only scales the residual that the network adds, so settings of one network perform identical computation. Repeated settings nevertheless differed by up to about 20\% in measured latency, which is timing noise. We therefore assign each configuration the median latency of its network, and treat differences between networks of less than about 10\% as unresolved.

Cost is an estimate of on-demand compute price per unit of media:
\begin{equation}
C = \frac{t \cdot U}{3.6\times10^{6}}\, p,
\end{equation}
where $t$ is the latency in milliseconds per unit, $U$ the number of units (1{,}000 megapixels, or 1{,}000 hours of audio expressed in seconds), and $p$ the hourly price: \$\val{wm:price/gpu} for g5.xlarge and \$\val{wm:price/cpu} for m6a.xlarge, a 4-vCPU instance of the same processor family, both us-east-1 on-demand Linux prices effective \val{wm:price/date} and retrieved from the AWS Price List API. Each configuration is costed on the cheaper of the two. The estimate assumes a fully utilised instance and excludes decoding, storage, data transfer and idle capacity; it is intended for comparing methods, not for budgeting a deployment.

\subsubsection{Qualification and Combined Score}
A configuration \emph{qualifies} when (i) its code and weights are openly licensed, (ii) gate bit accuracy is at least 0.95 both pooled and on the worst subset, and (iii) its false-positive evidence is analytic or observed. The best-cost, best-speed and best-combined configurations in each table are chosen among qualifying configurations only: a fast method whose mark does not survive the gate transformation is not a usable operating point.

The combined score summarises four criteria as ratios to the best qualifying configuration, so that each lies in $(0,1]$ and equals 1 for the best:
\begin{align}
r_q &= \begin{cases} F^{*}/F & \text{image (FLIP } F\text{)}\\ (P-1)/(P^{*}-1) & \text{audio (PESQ } P\text{)} \\ V/V^{*} & \text{video (VMAF } V\text{)}\end{cases}\nonumber\\
r_b &= (a-0.5)/(a^{*}-0.5),\quad r_t = t^{*}/t,\quad r_c = C^{*}/C,\nonumber\\
S &= \exp\Big(\sum_{j} w_j \ln r_j\Big),\quad \sum_j w_j = 1,
\end{align}
where $a$ is worst-subset gate bit accuracy (0.5 is chance) and starred quantities are the best qualifying values. With equal weights, $S$ is the geometric mean of the four ratios, which rewards balance and penalises a configuration that is poor on any one criterion. Because speed and cost are strongly correlated, we also report a quality-weighted variant ($w_q=0.5$, $w_b=0.2$, $w_t=w_c=0.15$) and note where the two disagree.

\subsubsection{Statistics}
\label{sec:wm-stats}
The experimental unit is the source: an image, a video clip, or an audio clip together with its three sample-rate renderings. Means are over items, and their percentile bootstrap 95\% intervals come from 10{,}000 resamples of sources, so that the three renderings of an audio clip are drawn together and the intervals reflect \val{wm:audio/n_sources} independent audio sources rather than 138 inputs (Table~\ref{tab:wm-ci} in the Appendix). Planned comparisons between the leading configurations use two-sided Wilcoxon signed-rank tests on per-source differences \cite{wilcoxon1945}, averaging the renderings of an audio clip first, with Holm adjustment over the family of five tests \cite{holm1979}. Video results rest on six clips and are therefore descriptive; we do not claim significance for video differences.

\subsubsection{Cross-Device Determinism}
To test whether decoding is reproducible across hardware, we decoded the same compressed files on the benchmark host (x86-64, Linux) and on an Apple M-series machine (arm64, macOS), each with one and four CPU threads and with its accelerator (CUDA or MPS). Files were twelve Video Seal-marked clips (three clips, strengths 0.4 and 1.1, CRF 18 and 22) and one TrustMark-Q-marked image after JPEG 75. For each file we recorded the largest logit difference within one host and across hosts, the number of flipped bits, and whether decoded frames were byte-identical across hosts before and after conversion from YUV to RGB. To separate the colour conversion from the network, each host then converted the first \val{wm:rgb/nframes} frames of every clip to RGB once and saved the arrays, and both hosts decoded both hosts' arrays with the same TorchScript detector (a 2$\times$2 design: input host by decoding host). Both hosts used PyAV \val{wm:rgb/av} with libswscale \val{wm:rgb/swscale}.

\subsection{Fingerprinting Protocol}
\label{sec:fp-method}
\begin{figure*}[!t]
\centering
\begin{tikzpicture}[
  font=\sffamily\footnotesize,
  box/.style={draw=black!70, line width=0.5pt, rounded corners=1.5pt, align=center, minimum height=9mm,
              text width=30mm, inner xsep=2pt, fill=white},
  data/.style={box, fill=black!5},
  metric/.style={box, draw=none, fill=blue!8, minimum height=10mm, text width=33mm},
  arr/.style={-{Stealth[length=1.8mm]}, line width=0.5pt, draw=black!70},
  lab/.style={font=\sffamily\scriptsize, text=black!60},
]
\node[data] (reg) at (0,0) {Registered assets\\\scriptsize images, audio, video};
\node[box] (wm) at (3.85,0) {Optional watermark\\\scriptsize TrustMark, AudioSeal,\\[-2pt]\scriptsize Video Seal};
\node[box] (fp) at (7.7,0) {Fingerprint\\\scriptsize \val{fp:count/methods} methods};
\node[data] (idx) at (11.55,0) {Registry index\\\scriptsize assets + distractors};
\draw[arr] (reg) -- (wm);
\draw[arr] (wm) -- (fp);
\draw[arr] (fp) -- (idx);
\node[lab, anchor=south] at (5.775,0.72) {registration};

\node[data] (neg) at (0,-1.9) {Never-registered\\\scriptsize media (negatives)};
\node[box] (att) at (3.85,-1.9) {Transform or edit\\\scriptsize \val{fp:count/attacks} attacks, 6 edits};
\node[box] (qfp) at (7.7,-1.9) {Fingerprint\\\scriptsize query};
\node[box] (srch) at (11.55,-1.9) {Exact search\\\scriptsize top-50, threshold};
\draw[arr] (reg.south east) -- node[lab, right, pos=0.45, inner sep=3pt] {positives} (att.north west);
\draw[arr] (neg) -- node[lab, below, inner sep=2pt] {negatives} (att);
\draw[arr] (att) -- (qfp);
\draw[arr] (qfp) -- (srch);
\draw[arr] (idx) -- (srch);
\node[lab, anchor=north] at (5.775,-2.55) {queries};

\node[metric] (m1) at (0,-3.75) {\textbf{Robustness}\\\scriptsize R@1, $\mu$AP, TPR at calibrated FPR};
\node[metric] (m2) at (3.85,-3.75) {\textbf{Localization}\\\scriptsize edited region; time offset};
\node[metric] (m3) at (7.7,-3.75) {\textbf{Security}\\\scriptsize evasion, forgery, inversion};
\node[metric] (m4) at (11.55,-3.75) {\textbf{Complementarity}\\\scriptsize watermark key vs.\ fingerprint};
\draw[densely dashed, black!50, line width=0.5pt] ($(m1.north west)+(-0.1,0.18)$) -- ($(m4.north east)+(0.1,0.18)$);
\end{tikzpicture}
\caption{Benchmark protocol. \val{fp:count/methods} fingerprinting methods are applied to registered assets and distractors (registry) and to queries derived from registered assets (positives) or from never-registered media (negatives) under \val{fp:count/attacks} transformations and six kinds of partial edit. Thresholds are fixed on half of the negatives and evaluated on the other half. The same queries also drive the localization, security and watermark-complementarity measurements.}
\label{fig:fp-protocol}
\end{figure*}
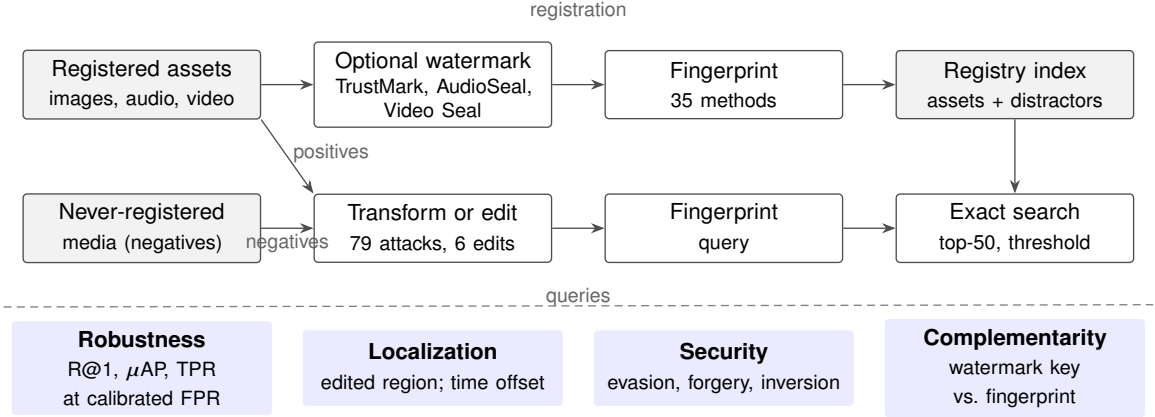

Fig.~\ref{fig:fp-protocol} summarises the protocol. Each method fingerprints a set of registered assets and a larger set of distractors, which together form the registry. Registered assets are then transformed or partially edited and used as queries; media that were never registered go through the same transformations and serve as negative queries. Every query is searched exhaustively against the registry, and a threshold fixed on one half of the negatives decides whether the best match is reported. All randomness is seeded, so every method sees byte-identical queries.

\subsubsection{Methods Under Test}
We included every fingerprinting method we could find that is distributed with an implementation and, where learned, with weights, and that targets natural images, music and speech, or natural video (Table~\ref{tab:fp-methods}). We recorded the licence of the code and, separately, of the weights and of the data they were trained on, because the three differ in several projects. A method is \emph{product-eligible} (tier~P) when code and weights carry a permissive licence (MIT, BSD, Apache-2.0 or CC0) and the weights were not trained on data restricted to non-commercial use. The strongest learned copy detectors are not: SSCD and the ISC21 winner are MIT-licensed code, but their weights carry no licence of their own and were trained on DISC21, whose repository is CC BY-NC 4.0; DINOHash has no licence file; and the neural music fingerprint NMFP is GPL-3.0. We measured them as reference points (tier~R, marked \reftier), as the watermarking track does for its unlicensed checkpoints (Section~\ref{sec:wm-method}). We excluded Apple NeuralHash (weights extracted from a device, no licence), PhotoDNA (licensed only for child-safety use), Adobe's image comparator network (no public code), SuperPoint and TruFor (non-commercial weights), and Olaf and Panako, which we did not run: they are AGPL-3.0, whose network-use clause rules them out for a hosted provenance service, and Chromaprint and audfprint already represent their landmark and sub-band designs.

\paragraph{Images.} Twenty-one descriptors: PDQ \cite{meta2019pdq}, with and without matching the query's eight dihedral variants; aHash, dHash, pHash (64 and 256 bits) and wHash from imagehash \cite{buchner_imagehash}; BlockMean, Marr--Hildreth and ColorMoment from OpenCV \cite{opencv_imghash}; Blockhash \cite{commonsmachinery_blockhash}; the ISCC Image-Code at 64 bits (the ISO default) and 256 bits \cite{iso24138}; DINOv2-S and DINOv2-B class tokens \cite{oquab2024dinov2}; a keyed 256-bit sign projection of DINOv2-S (below); OpenCLIP ViT-B/32 \cite{cherti2023openclip}; SSCD with ResNet-50 and ResNeXt-101 backbones \cite{pizzi2022sscd}; the ISC21 descriptor-track winner \cite{yokoo2021isc}; and DINOHash \cite{singhi2025dinohash}. ISCC's experimental semantic image code \cite{iscc_sci} ran in our pilot but not at full scale: it binarises an ISC21-derived descriptor that we evaluate directly, and its reference model runs one image at a time at about 3\,s per image on a CPU core. Learned descriptors see the whole image resized to their square input without cropping, so that a crop attack cannot remove content the descriptor was never shown. Blockhash's reference implementation loops over pixels in Python; we vectorised it and verified bit-identical output on five image sizes including odd ones.

\paragraph{Keyed projection.} DINOv2-S LSH-256 multiplies the L2-normalised DINOv2-S descriptor by a $384 \times 256$ Gaussian matrix drawn from a secret seed and keeps the signs. It is a standard random-hyperplane hash, included because it is compact, product-eligible and, unlike a raw descriptor, meaningless to anyone without the key; the security experiments test that claim.

\paragraph{Audio.} Chromaprint \cite{lalinsky_chromaprint}, with candidate generation on the 20 most significant bits of each sub-fingerprint and verification by bit error rate at the voted offset (the AcoustID index design); the ISCC Audio-Code at 64 and 256 bits, computed from the whole-file Chromaprint vector; audfprint's spectral-peak landmarks \cite{ellis_audfprint}, scored by the number of hashes that agree on one time offset; LAION-CLAP \cite{wu2023clap} as a semantic embedding, mean-pooled over 10\,s windows; and NMFP \cite{araz2025nmfp}, a neural music fingerprint with 1\,s segment embeddings, run in its authors' TensorFlow environment and matched by segment voting.

\paragraph{Video.} vPDQ \cite{meta_vpdq} at one hash per second, scoring a candidate by the share of query frames with a reference frame within Hamming distance 31; TMK+PDQF \cite{meta2019pdq, poullot2015tmk}, scored by its level-2 time-kernel similarity using Meta's command-line tools; the ISCC Video-Code at 64 and 256 bits from MPEG-7 frame signatures \cite{paschalakis2012mpeg7}; and DINOv2-S and SSCD on frames sampled at 1\,fps, either mean-pooled into one clip descriptor or matched frame by frame with temporal voting, the baseline recipe of the 2023 Video Similarity Challenge \cite{pizzi2024vsc}.

\begin{table*}[!t]
\centering
\scriptsize
\caption{Methods, licences, stored size and extraction time. Bits: stored bits per registered asset (float descriptors count 32 bits per dimension; time-based systems report the mean over the registered items). Time: per image (1024 px), per second of audio or per 5\,s clip.}
\label{tab:fp-methods}
\setlength{\tabcolsep}{3pt}
\begin{tabular}{lllp{3.3cm}p{3.6cm}rr}
\toprule
Modality & Method & Class & Code licence & Weights / data & Bits & Time (ms) \\
\midrule
Image & PDQ & Perceptual hash & BSD (ThreatExchange); pdqhash binding  & n/a & 256 & 31 CPU \\
 & PDQ (dihedral) & Perceptual hash & BSD (ThreatExchange); pdqhash binding  & n/a & 256 & 31 CPU \\
 & aHash & Perceptual hash & BSD-2-Clause (imagehash) & n/a & 64.0 & 4.3 CPU \\
 & dHash & Perceptual hash & BSD-2-Clause (imagehash) & n/a & 64.0 & 4.4 CPU \\
 & pHash & Perceptual hash & BSD-2-Clause (imagehash) & n/a & 64.0 & 5.2 CPU \\
 & pHash-256 & Perceptual hash & BSD-2-Clause (imagehash) & n/a & 256 & 5.6 CPU \\
 & wHash & Perceptual hash & BSD-2-Clause (imagehash) & n/a & 64.0 & 63 CPU \\
 & BlockMean & Perceptual hash & Apache-2.0 (opencv\_contrib) & n/a & 256 & 1.0 CPU \\
 & Marr–Hildreth & Perceptual hash & Apache-2.0 (opencv\_contrib) & n/a & 576 & 5.5 CPU \\
 & ColorMoment & Perceptual hash & Apache-2.0 (opencv\_contrib) & n/a & 1344 & 3.6 CPU \\
 & Blockhash & Perceptual hash & MIT & n/a & 256 & 19 CPU \\
 & ISCC Image-64 & ISCC (ISO 24138) & Apache-2.0 & n/a & 64.0 & 11 CPU \\
 & ISCC Image-256 & ISCC (ISO 24138) & Apache-2.0 & n/a & 256 & 11 CPU \\
 & DINOv2-S & General embedding & Apache-2.0 & Apache-2.0 (HF model card) & 12288 & 2.0 GPU \\
 & DINOv2-B & General embedding & Apache-2.0 & Apache-2.0 (HF model card) & 24576 & 3.9 GPU \\
 & DINOv2-S LSH-256 & General embedding & Apache-2.0 & Apache-2.0 (HF model card) & 256 & 1.7 GPU \\
 & OpenCLIP B/32 & General embedding & MIT (open\_clip) & MIT (laion/CLIP-ViT-B-32-laion2B-s34B-b79K) & 16384 & 1.1 GPU \\
 & SSCD R50\reftier & Copy-detection trained & MIT & none stated; trained on DISC21 (CC BY-NC 4.0 & 16384 & 2.7 GPU \\
 & SSCD RX101\reftier & Copy-detection trained & MIT & none stated; trained on DISC21 (CC BY-NC 4.0 & 32768 & 4.6 GPU \\
 & ISC21 1st\reftier & Copy-detection trained & MIT & MIT repo release; trained on DISC21 (CC BY-N & 8192 & 8.3 GPU \\
 & DINOHash-96\reftier & Copy-detection trained & none (no LICENSE file) & none (no LICENSE file) & 96.0 & 3.6 GPU \\
\addlinespace
Audio & Chromaprint & Perceptual hash & MIT (chromaprint core; fpcalc binary l & n/a & 7072 & 2.0 CPU \\
 & audfprint & Perceptual hash & MIT & n/a & 40397 & 1.4 CPU \\
 & ISCC Audio-64 & ISCC (ISO 24138) & Apache-2.0 & n/a & 64.0 & 2.0 CPU \\
 & ISCC Audio-256 & ISCC (ISO 24138) & Apache-2.0 & n/a & 256 & 2.0 CPU \\
 & CLAP & General embedding & CC0-1.0 & CC0-1.0 (LAION-AI/CLAP release) & 16384 & 1.1 GPU \\
 & NMFP\reftier & Copy-detection trained & GPL-3.0 & Zenodo 15719945 (licence as repo, GPL-3.0);  & 239108 & -- \\
\addlinespace
Video & vPDQ & Perceptual hash & BSD (ThreatExchange vpdq) & n/a & 1258 & 132 CPU \\
 & TMK+PDQF & Perceptual hash & BSD (ThreatExchange tmk) & n/a & 2106752 & 51 CPU \\
 & ISCC Video-64 & ISCC (ISO 24138) & Apache-2.0 & n/a & 64.0 & 54 CPU \\
 & ISCC Video-256 & ISCC (ISO 24138) & Apache-2.0 & n/a & 256 & 54 CPU \\
 & DINOv2-S frames & General embedding & Apache-2.0 & Apache-2.0 (HF model card) & 61440 & 136 GPU \\
 & DINOv2-S mean & General embedding & Apache-2.0 & Apache-2.0 (HF model card) & 12288 & 136 GPU \\
 & SSCD frames\reftier & Copy-detection trained & MIT & none stated; trained on DISC21 (CC BY-NC 4.0 & 81920 & 137 GPU \\
 & SSCD mean\reftier & Copy-detection trained & MIT & none stated; trained on DISC21 (CC BY-NC 4.0 & 16384 & 137 GPU \\
\bottomrule
\end{tabular}
\par\vspace{2pt}\parbox{\linewidth}{\raggedright\scriptsize \reftier~Reference tier: weights without a licence, trained on non-commercial data, or copyleft code; measured for comparison, not product-eligible.}
\end{table*}

\subsubsection{Corpora}
\emph{Images.} Amazon Berkeley Objects (ABO, CC BY 4.0) \cite{collins2022abo} supplies product photographs from real marketplace listings, the setting in which copied and relabelled product images circulate. We sampled \val{fp:corpus/abo/reg} registered originals (one main image per product, shorter side at least 800\,px), stored at a longer side of 1024\,px as the asset a provenance service would register, and \val{fp:corpus/abo/dist} distractors (one image per other product). Queries come from \val{fp:corpus/abo/pos} registered originals; negative queries from \val{fp:corpus/abo/neg} images of products that are neither registered nor distractors; and hard negatives from \val{fp:corpus/abo/hard} \emph{other} photographs of registered products. ABO reuses photographs across listings and colour variants, so before scoring we flagged pairs that are copies of each other under a rule fixed in advance on unattacked originals (PDQ distance at most 16 bits, SSCD cosine at least 0.80, or DINOv2-B cosine at least 0.95). The rule removed \val{fp:dedup/n_dist_flagged} distractors and \val{fp:dedup/n_neg_flagged} negative sources, and marked \val{fp:dedup/n_hard_near_duplicate} of \val{fp:dedup/n_hard} hard negatives as recoloured versions of the registered photograph, which we report separately. As a second, literature-comparable set we fetched a subset of the DISC21 development split \cite{douze2021isc}: \val{fp:corpus/disc/gt} ground-truth query--reference pairs, \val{fp:corpus/disc/negq} queries without a reference, and \val{fp:corpus/disc/refs} references in total. DISC21 is licensed for non-commercial research; we use it for measurement only.

\emph{Audio.} \val{fp:corpus/audio/music} music tracks from the FMA large set \cite{defferrard2017fma}, restricted to tracks released under CC BY, CC BY-SA or CC0 and excluding the FMA-medium subset, and \val{fp:corpus/audio/speech} LibriSpeech utterances of at least 8\,s \cite{panayotov2015librispeech}, all at 22.05\,kHz mono. Music clips are 30\,s. Of these, \val{fp:corpus/audio/reg} are registered, \val{fp:corpus/audio/dist} are distractors and \val{fp:corpus/audio/neg} are negative sources.

\emph{Video.} Five-second clips cut every 5\,s from four Blender open films \cite{blender2010sintel, blender2008bbb, blender2012tos, blender2006ed}, skipping titles and end credits, and the first 5\,s of \val{fp:corpus/video/xiph} Xiph test sequences \cite{xiphderf}, scaled to 1280\,px wide at 25\,fps and stored as H.264 at CRF~12: \val{fp:corpus/video/reg} registered, \val{fp:corpus/video/dist} distractor and \val{fp:corpus/video/neg} negative clips. Clips from the same film share style and characters, which makes the distractors deliberately hard.

\subsubsection{Transformations}
\emph{Images} (\val{fp:count/image/attacks}, grouped into families in Fig.~\ref{fig:fp-image-heatmap}): JPEG at quality 75, 50 and 30 and WebP at 50; downscaling to 0.5 and 0.25 and a 0.75 aspect change; centred crops keeping 90, 70, 50 and 30\% of each side and an off-centre 50\% crop; rotation by 5, 15 and 90 degrees, horizontal flip, perspective, skew and 25\% padding; brightness, contrast, saturation, gamma, greyscale, blur, Gaussian noise and pixelation; text, emoji (30\% of the image), meme caption, screenshot and overlay onto another photograph at half size, following AugLy \cite{papakipos2022augly}; and two chains, a social-media chain (downscale, JPEG~70, 90\% crop) and a re-capture simulation (perspective, defocus, sensor noise, tone curve, JPEG~80). \emph{Audio} (\val{fp:count/audio/attacks}): MP3 and AAC at 64\,kb/s, Opus at 24\,kb/s, resampling through 8\,kHz; white noise at 20 and 10\,dB SNR and babble from other music at 10 and 0\,dB; telephone band, 2\,kHz low-pass, dynamic compression and reverberation; pitch $+2$ semitones, tempo 0.9 and 1.1 and speed 1.05; excerpts of 10 and 5\,s at a random offset, an excerpt with MP3 and noise, and a re-recording simulation. Unlike the watermarking track, codec outputs are not re-aligned: a fingerprint must cope with codec delay itself. \emph{Video} (\val{fp:count/video/attacks}): H.264 at CRF~28 and 35, H.265 at CRF~32; downscaling to 0.5 and 0.25; 75\% and 50\% crops; flip, 90 and 5 degree rotation, letterboxing; greyscale, brightness, contrast, blur and noise; text caption, logo, and picture-in-picture at half size on another clip; 15\,fps, 1.25$\times$ speed, a 2.5\,s excerpt, the same 2.5\,s inserted into 5\,s of another clip, and a screen-recording simulation.

\subsubsection{Retrieval Metrics and Operating Points}
Search is exact (flat index, top~50): Hamming similarity for binary codes, cosine for L2-normalised descriptors, and each audio or video system's own score. We report Recall@1 and the micro-average precision ($\mu$AP) of the ISC21 protocol \cite{douze2021isc}, both threshold-free. A registry, however, must decide, and a false decision binds content to someone else's manifest. We therefore fix thresholds on the \emph{calibration} half of the negative sources (all their transformed copies) and report on the other half. Two kinds of target are used. The query-level target fixes the rate at which a never-registered query's best match passes at the benchmark's registry size (1\%). The pair-level target fixes the rate at which a (negative query, reference) pair passes. For a registry of $N$ assets and an average pair-level false-match probability $p$, the expected number of false matches per negative query is $Np$; the probability of at least one false match is at most $\min(1,Np)$, without requiring independent reference comparisons. Our headline target is a pair-level rate of $10^{-7}$: at the benchmark's \val{fp:count/image/nref} references, this corresponds to about one expected false match per hundred negative queries in aggregate. This calculation is not a measured query-level false-binding rate. Table~\ref{tab:fp-ci} reports held-out query-level rates at the separately calibrated query-level threshold. We also report $10^{-8}$, the smallest rate our \val{fp:count/image/neg_cal} calibration queries resolve. A positive query counts as detected only if its best match is the true asset and passes the threshold. A target is reported only if the calibration set resolves it, that is, if it implies at least five false pairs (or queries) among the calibration negatives; otherwise the threshold would merely sit above every negative observed. This is a reporting-resolution rule, not evidence that the comparisons are independent or that the tail probability is estimated precisely. The queries are not independent: every registered or negative source contributes one query per transformation, and the transformed copies of one source tend to succeed or fail together. Intervals for these retrieval metrics therefore come from a bootstrap over sources (2{,}000 replicates): registered sources are drawn with replacement together with all their queries, calibration negative sources likewise, with the threshold recomputed from the drawn calibration set in every replicate, and held-out negative sources for the out-of-sample false-match rate. Calibration and held-out negatives come from disjoint sources. Planned paired comparisons use a sign-flip test on each source's difference in detections between the two methods, with Holm adjustment \cite{holm1979}, and report a bootstrap interval of the difference in detection rate; both methods are resampled with the same sources. The sign-flip test and the verification experiment (Table~\ref{tab:fp-verify}) hold the thresholds at their full-sample values, so their uncertainty excludes threshold estimation; the bootstrap interval of a paired difference recomputes the thresholds in every replicate.

\subsubsection{Partial Edits and Localization}
Partial edits are generated from \val{fp:edit/n_sources} registered originals, with the edited region known exactly: a smooth blob of 1, 3, 10 or 25\% of the image, placed where the image has texture. The edit is a splice from another photograph, a copy-move within the image, classical (Telea) or learned (LaMa \cite{suvorov2022lama}) inpainting of the region, a hue rotation, or replacement by a flat label with new text, followed by no post-processing, JPEG~75, downscaling with JPEG~80, or a 90\% crop. A balanced design gives \val{fp:edit/n_edits} edited images. We first ask whether each fingerprint still retrieves the original at its calibrated threshold. We then localize the edit against the registered original: the query is registered onto the original with SIFT and RANSAC \cite{lowe2004sift, fischler1981ransac} or with DISK and LightGlue \cite{tyszkiewicz2020disk, lindenberger2023lightglue}, and compared through six difference maps: blurred pixel difference, SSIM \cite{wang2004ssim}, LPIPS \cite{zhang2018lpips}, DINOv2-S patch distance, PDQ over an $8 \times 8$ grid of tiles, and a fusion of the five. Map thresholds are fixed at a pixel false-positive rate of 1\% on unedited originals put through the same post-processing, on half of the sources; the other half is evaluated. For audio and video we measure temporal localization: how accurately each system recovers the offset of an excerpt within its source, and which interval of an inserted clip matches.

\subsubsection{Security}
We consider an adversary who knows the fingerprinting method and wants either to \emph{evade} it (republish a registered asset so that it no longer matches) or to \emph{forge} a match (make an unrelated image bind to a registered asset, for example to borrow its credentials). Evasion uses projected gradient descent \cite{madry2018pgd} under an $L_\infty$ budget of 2, 4, 8 or 16 grey levels: white-box on the learned descriptors, through differentiable re-implementations of PDQ, pHash, aHash, dHash and the ISCC Image-Code for those hashes, and by transfer from an ensemble of DINOv2-S, SSCD and OpenCLIP for all methods. Success is always judged by the real extractor on the 8-bit result. Image and audio attacks were judged at the method's pair-level $10^{-6}$ threshold, and video attacks at its query-level 1\% threshold, because the video calibration set does not resolve $10^{-6}$ (Section~\ref{sec:fp-res-video}). For images, $10^{-6}$ is a looser operating point than the headline: an evasion must push the score below it, which makes the reported evasion rates conservative, and we re-score white-box evasions at the $10^{-7}$ threshold from their final scores. Forgery success at $10^{-6}$ is correspondingly an upper bound on success at $10^{-7}$. Forgery pushes a negative image towards a registered target and succeeds only if the target becomes the top match in the full registry above threshold. Because a soft-binding value is written into the manifest, we also test \emph{inversion}: a decoder trained on distractor descriptors reconstructs a 64$\times$64 thumbnail from a registered asset's descriptor, and we measure whether the reconstruction identifies the asset among the others. For audio and video we apply the same attacks to the learned systems and measure, for every method, the smallest pitch, tempo, speed or crop change that breaks it.

\subsubsection{Watermark Complementarity}
The registered assets of the three modalities were also watermarked with MIT-licensed models from the watermarking track: TrustMark~Q at its default strength for images \cite{bui2025trustmark}, AudioSeal \cite{sanroman2024audioseal} and Video Seal \cite{fernandez2024videoseal}. The marked copies went through the same transformations. For every query we record whether the watermark satisfies its method-specific success criterion, and whether each fingerprint identifies the asset. The three watermarks decide this differently, and the decisions relate differently to recovering a registry identifier. TrustMark transmits a 100-bit codeword; in the BCH\_5 mode we use it carries a 61-bit data payload, 35 parity bits and 4 version bits. Its BCH decoder either corrects the codeword or reports failure; the key counts as recovered when the decoded 61-bit payload equals the asset's key exactly, which is identification: the decoded payload itself names the registry entry. AudioSeal (the base model) carries only 16 bits and has no error correction. Its detection score is the fraction of audio frames classified as watermarked; we require at least half of the frames, a more permissive rule than the 0.8 of the watermarking track's blind detector (Section~\ref{sec:wm-fp-method}), because here detection counts only together with an exact match of all 16 bits to the asset's key. No AudioSeal decode of the watermarking track's unmarked audio exceeded \val{wm:audioseal/unmarked_max_score}, so its blind false-positive count is the same under either rule. A recovered key identifies an asset only if no other asset holds it. The space contains only $2^{16}$ keys, so a registry larger than that cannot give every asset a distinct AudioSeal key; our keys were drawn pseudo-randomly per asset without a uniqueness check, and the \val{fp:wm/audio/n_assets} registered assets received \val{fp:wm/audio/nkeys} distinct keys, so \val{fp:wm/audio/n_colliding} assets share one key. Video Seal carries 256 bits without error correction, and exact recovery of all 256 bits is rare after compression; we count the query as passing expected-key verification when at least $k$ of the 256 decoded bits agree with the asset's key, with $k$ the smallest value for which a random 256-bit word agrees with a given key with probability at most $10^{-6}$. For positives this is verification against the expected key, not a test of unique identification among all registered keys. A registry reader without that expectation would compare the decoded bits with every registered key; on the never-registered negatives we did exactly that, so a false Video Seal key there means agreement with \emph{any} of the registered keys. Under the independent-fair-bit null used to set $k$, the union bound gives a per-query false-match probability of at most $\min(1,N \times 10^{-6})$ for $N$ keys. This analytic bound is conditional on that null and does not replace the measured false-match rate. Testing all keys on negative queries does not establish unique identification on positive queries. For AudioSeal negatives we report both the detector's false detections and detections whose message equals a registered key. We also measure how far the watermark itself moves each fingerprint, and whether the registry should hold the fingerprint of the original or of the marked asset.

\subsubsection{Second-Stage Verification}
To test whether checking a retrieved candidate against the stored original removes the false matches that a global descriptor lets through, we ran SIFT with RANSAC between every image query (positive and negative) and the registered original of its top-1 candidate under DINOv2-S and PDQ, at the localization track's working size, and recorded the number of geometric inliers. A match is accepted when the retrieval score passes a threshold and the inlier count reaches $k$; both are fixed on the calibration negatives and evaluated on the held-out ones.

\subsubsection{Hosts and Reproducibility}
Learned descriptors were extracted on one NVIDIA A10G (AWS g5.2xlarge, AMD EPYC 7R32); CPU descriptors single-threaded on the same processor model (an AWS c5a.16xlarge was added to spread the CPU work). Every source tree, checkpoint and corpus file is pinned by commit or SHA-256. Because the watermarking track found host-dependent decoder outputs (Section~\ref{sec:wm-res-stability}), we fingerprinted the same 200 PNG files on the GPU host and on an Apple M5 Pro laptop and compared the descriptors bit by bit.

The retrieval tracks and the watermark combination were run twice, on 2026-09-26 and again on 2026-09-28 on fresh hosts of the same types, because the first run did not keep the per-query scores that source-level intervals require. The corpus was rebuilt from its sources and matched the pinned SHA-256 of every image, audio and video item. The image track reproduced the first run exactly (every pooled rate to within $10^{-6}$), and the video track to within two of 3{,}000 queries. The audio track did not reproduce exactly, for two reasons that we found and fixed: the background music of the two babble transformations had been drawn from whichever surplus FMA downloads were present, which differed between runs (the pool is now pinned by name and hash), and CLAP was extracted on the CPU host in the second run instead of the GPU. All reported numbers come from the second run.

\section{Watermark Results}
\label{sec:wm-results}
This section reports the watermarking track: perceptual quality, robustness and payload recovery for each modality, then false positives, cost and cross-device determinism. Configurations are ranked only if they qualify (open licence, gate robustness of at least 0.95 pooled and on the worst subset, and false-positive evidence); reference-only methods are shown for comparison.

\begin{table*}[!t]
\centering
\footnotesize
\caption{Image watermarking on 85 images (24 Kodak, 25 DIV2K, 28 CLIC, 8 16-bit HDR). Means over images (FLIP: 60 images without DIV2K; CVVDP: 32 Kodak and HDR images); bit accuracy after JPEG at quality 75, pooled and on the worst content subset.}
\label{tab:wm-image}
\setlength{\tabcolsep}{3.4pt}
\begin{tabular}{lcrrrrrrrrrrr}
\toprule
Configuration & Q & PSNR & SSIM & LPIPS & FLIP & CVVDP & \multicolumn{2}{c}{JPEG 75 bit acc.} & Blind & GPU & Cost & Comb. \\
  &   & (dB)$\uparrow$ & $\uparrow$ & $\downarrow$ & $\downarrow$ & $\uparrow$ & pooled & worst set & FPR$\downarrow$ & (ms/MP) & (\$/k MP) & $\uparrow$ \\
\midrule
PixelSeal (0.15)\textsuperscript{\bestmark} & \checkmark & 48.3 & 0.9969 & \textbf{0.0012} & 0.024 & 9.93 & 0.988 & 0.980 & n/a & 9.13 & 0.0026 & \textbf{0.95} \\
PixelSeal (0.2) & \checkmark & 45.9 & 0.9952 & 0.0020 & 0.030 & 9.88 & 0.998 & 0.996 & n/a & 9.13 & 0.0026 & 0.90 \\
Video Seal (0.2)\textsuperscript{\costmark}\textsuperscript{\speedmark} & \checkmark & 44.8 & 0.9953 & 0.0021 & 0.034 & 9.82 & 0.997 & 0.996 & n/a & \textbf{8.79} & \textbf{0.0025} & 0.90 \\
PixelSeal (0.4) & \checkmark & 40.1 & 0.9863 & 0.0077 & 0.051 & 9.57 & 0.999 & 0.999 & n/a & 9.13 & 0.0026 & 0.79 \\
Video Seal (0.4) & \checkmark & 38.9 & 0.9856 & 0.0082 & 0.058 & 9.39 & 1.000 & 0.999 & n/a & \textbf{8.79} & \textbf{0.0025} & 0.79 \\
WAM (1) & \checkmark & 45.3 & 0.9948 & 0.0099 & 0.043 & 9.97 & 0.994 & 0.990 & n/a & 19.1 & 0.0053 & 0.57 \\
WAM (2) & \checkmark & 39.4 & 0.9852 & 0.0367 & 0.071 & 9.88 & \textbf{1.000} & \textbf{1.000} & n/a & 19.1 & 0.0053 & 0.51 \\
TrustMark-P (0.8) & \checkmark & \textbf{50.5} & \textbf{0.9985} & 0.0017 & \textbf{0.022} & \textbf{9.98} & 0.977 & 0.952 & 0.021 & 45.1 & 0.013 & 0.43 \\
TrustMark-P (1) & \checkmark & 49.4 & 0.9980 & 0.0024 & 0.024 & 9.97 & 0.990 & 0.977 & 0.021 & 45.1 & 0.013 & 0.43 \\
TrustMark-P (1.5) & \checkmark & 46.8 & 0.9964 & 0.0049 & 0.029 & 9.93 & 0.998 & 0.995 & 0.021 & 45.1 & 0.013 & 0.41 \\
TrustMark-B (1) & \checkmark & 41.0 & 0.9940 & 0.0029 & 0.043 & 9.45 & 0.999 & 0.997 & 0.038 & 55.7 & 0.013\textsuperscript{c} & 0.35 \\
TrustMark-Q (1) & \checkmark & 42.1 & 0.9943 & 0.0027 & 0.040 & 9.48 & 0.999 & 0.999 & 0.026 & 63.5 & 0.014\textsuperscript{c} & 0.34 \\
TrustMark-C (1) & \checkmark & 40.0 & 0.9926 & 0.0035 & 0.047 & 9.38 & 0.995 & 0.992 & 0.044 & 54.4 & 0.014\textsuperscript{c} & 0.34 \\
TrustMark-Q (1.2) & \checkmark & 40.7 & 0.9925 & 0.0037 & 0.045 & 9.33 & 1.000 & 1.000 & 0.026 & 63.5 & 0.014\textsuperscript{c} & 0.33 \\
TrustMark-Q (1.3) & \checkmark & 40.0 & 0.9914 & 0.0043 & 0.047 & 9.25 & 1.000 & 0.999 & 0.026 & 63.5 & 0.014\textsuperscript{c} & 0.33 \\
TrustMark-Q (1.4) & \checkmark & 39.4 & 0.9904 & 0.0049 & 0.049 & 9.18 & 1.000 & 0.999 & 0.026 & 63.5 & 0.014\textsuperscript{c} & 0.32 \\
TrustMark-Q (1.5) & \checkmark & 38.9 & 0.9893 & 0.0056 & 0.052 & 9.11 & 1.000 & 0.999 & 0.026 & 63.5 & 0.014\textsuperscript{c} & 0.32 \\
DWT-DCT-SVD & \checkmark & 39.1 & 0.9782 & 0.0323 & 0.051 & 9.90 & 0.974 & 0.961 & n/a & 409 & 0.014\textsuperscript{c} & 0.20 \\
PixelSeal (0.1) & -- & 51.6 & 0.9983 & 0.0006 & 0.016 & 9.97 & 0.907 & 0.863 & n/a & 9.13 & 0.0026 & -- \\
Video Seal (0.1) & -- & 50.5 & 0.9985 & 0.0006 & 0.018 & 9.95 & 0.887 & 0.869 & n/a & 8.79 & 0.0025 & -- \\
PixelSeal (0.125) & -- & 49.8 & 0.9976 & 0.0009 & 0.020 & 9.95 & 0.965 & 0.936 & n/a & 9.13 & 0.0026 & -- \\
InvisMark\textsuperscript{\dag} & -- & 50.1 & 0.9963 & 0.0017 & 0.022 & 9.99 & 0.919 & 0.873 & n/a & 8.73 & 0.0024 & -- \\
DWT-DCT & -- & 39.3 & 0.9673 & 0.0212 & 0.029 & 9.95 & 0.518 & 0.508 & n/a & 409 & 0.014\textsuperscript{c} & -- \\
ChunkySeal (0.2) & -- & 44.4 & 0.9925 & 0.0090 & 0.048 & 9.84 & 0.934 & 0.789 & n/a & 50.4 & 0.014 & -- \\
RivaGAN\textsuperscript{\dag} & -- & 40.5 & 0.9735 & 0.0643 & 0.077 & 9.85 & 0.996 & 0.988 & n/a & 5250 & 1.5 & -- \\
\bottomrule
\end{tabular}
\par\vspace{2pt}\parbox{\linewidth}{\raggedright\scriptsize \costmark\ lowest cost, \speedmark\ fastest GPU embedding, \bestmark\ highest combined score, each among qualifying configurations (Q). Ties between strengths of one network are broken by quality. Bold: best qualifying value per column. \dag\ Code or weights lack an open licence; reported for reference and excluded from ranking. Blind FPR: share of unmarked decodes (four per image) on which the model's own detector fires, per decode; Table~\ref{tab:wm-fpr} gives it per source (n/a: no blind detector). Cost: estimated USD per 1{,}000 megapixels on the cheaper of GPU and CPU (\textsuperscript{c}: CPU cheaper). Latency is the median over the network's strength settings (Section~\ref{sec:wm-cost}).}
\end{table*}

\begin{figure*}[!t]
\centering
\includegraphics[width=\textwidth]{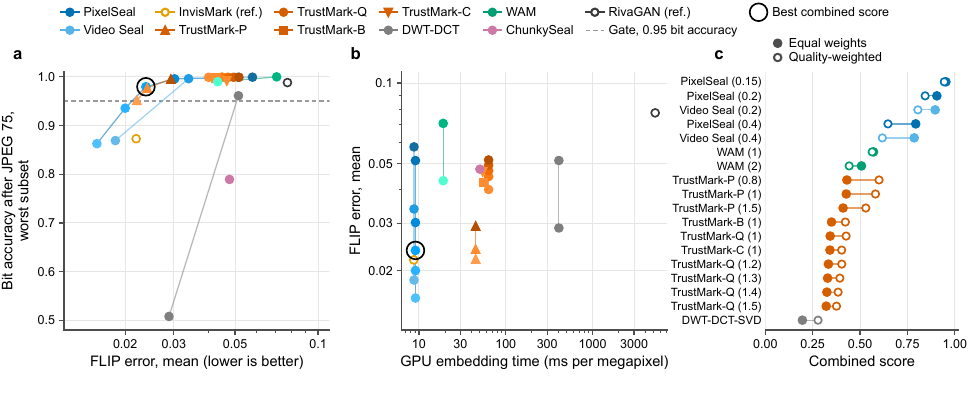}
\caption{Image watermarking trade-offs. (a) Perceptual error (FLIP, lower is better) against bit accuracy after JPEG 75 on the worst content subset; lines join strength settings of one network, darker shades are stronger settings, and open markers are reference-only methods. The dashed line is the 0.95 gate. (b) GPU embedding time against FLIP. (c) Combined score of the qualifying configurations with equal weights (filled) and quality-weighted (open). The ringed configuration has the highest combined score. Values are in Table~\ref{tab:wm-image}.}
\label{fig:wm-image}
\end{figure*}

\subsection{Images}
\label{sec:wm-res-image}
Table~\ref{tab:wm-image} and Fig.~\ref{fig:wm-image} summarise the image results. Of the \val{wm:image/n_configs} image configurations, \val{wm:image/n_qualify} qualify. PixelSeal at strength 0.15 has the highest combined score (\val{wm:image/PixelSeal@0.15/comb}). Its mean FLIP is \val{wm:image/PixelSeal@0.15/flip} (95\% CI \val{wm:image/PixelSeal@0.15/flip/lo} to \val{wm:image/PixelSeal@0.15/flip/hi}), PSNR \val{wm:image/PixelSeal@0.15/psnr}~dB and LPIPS \val{wm:image/PixelSeal@0.15/lpips}, and it keeps \val{wm:image/PixelSeal@0.15/gate} bit accuracy after JPEG 75 (\val{wm:image/PixelSeal@0.15/worst} on Kodak, its worst subset). Against TrustMark-Q at its default strength, it has \val{wm:ratio/ps_tq/flip} times the FLIP error (\val{wm:ps/flip} versus \val{wm:tq/flip}), an LPIPS of \val{wm:ps/lpips} versus \val{wm:tq/lpips}, and embeds \val{wm:ratio/ps_tq/gpu} times faster on the GPU.

PixelSeal's operating range is narrow. At 0.125 it passes the pooled gate (\val{wm:image/PixelSeal@0.125/gate}) but not on Kodak (\val{wm:image/PixelSeal@0.125/worst}); at 0.1 it fails both (\val{wm:image/PixelSeal@0.1/gate}). Kodak was the worst subset for most learned methods. Its images are small (768$\times$512), so a watermark trained at 256$\times$256 is spread over fewer pixels per bit than on the larger DIV2K and CLIC images.

The lowest FLIP among qualifying configurations belongs to TrustMark-P at strength 0.8 (\val{wm:image/trustmark-P@0.8/flip}). Its advantage over PixelSeal at 0.15 was not significant after correction (Table~\ref{tab:wm-tests}), and it passes the gate with little margin on Kodak (\val{wm:image/trustmark-P@0.8/worst}). It also embeds about five times more slowly. PixelSeal at 0.15 did have significantly lower FLIP than Video Seal at its default strength and than TrustMark-Q at 1.2 (both $p_{\mathrm{Holm}}<0.001$).

Bit accuracy and payload recovery tell different stories at low strength (Table~\ref{tab:wm-exact}). PixelSeal at 0.15 keeps \val{wm:image/PixelSeal@0.15/gate} of its bits after JPEG 75, and every image passes the expected-payload test (\val{wm:image/PixelSeal@0.15/keymatch_gate}), but all 256 bits are correct in only \val{wm:image/PixelSeal@0.15/exact_gate} of images. A verifier that knows the expected payload needs only the threshold test, which PixelSeal passes with a wide margin; a reader that must recover an unknown identifier bit for bit would need the default strength or an error-correcting code. TrustMark, whose BCH layer corrects the residual errors, recovers the exact payload for \val{wm:image/trustmark-Q@1/exact_gate} of images at its default strength.

\begin{table*}[!t]
\centering
\footnotesize
\caption{Exact payload recovery. Bit acc.: share of payload bits recovered at the gate transformation (JPEG 75, MP3 128~kb/s, H.264 CRF 23). Exact: share of items whose decoded payload equals the embedded one, at the gate and averaged over all transformations. For TrustMark exact recovery is after BCH decoding; no other method has an error-correcting layer, so exact recovery requires every bit. $\geq k$: share of items whose decoded bits match the embedded payload in at least $k$ of $n$ positions, the expected-payload decision of Table~\ref{tab:wm-fpr} (-- where no such $k$ exists).}
\label{tab:wm-exact}
\setlength{\tabcolsep}{3.4pt}
\begin{tabular}{llrlrrrr}
\toprule
Modality & Configuration & Bits & ECC & Bit acc. & $\geq k$ (gate) & Exact (gate) & Exact (mean) \\
\midrule
Image & PixelSeal (0.15) & 256 & none & 0.988 & 1.000 & 0.318 & 0.254 \\
Image & PixelSeal (0.2) & 256 & none & 0.998 & 1.000 & 0.765 & 0.590 \\
Image & Video Seal (0.2) & 256 & none & 0.997 & 1.000 & 0.659 & 0.506 \\
Image & PixelSeal (0.4) & 256 & none & 0.999 & 1.000 & 0.871 & 0.771 \\
Image & Video Seal (0.4) & 256 & none & 1.000 & 1.000 & 0.906 & 0.745 \\
Image & WAM (1) & 32 & none & 0.994 & 0.976 & 0.918 & 0.828 \\
Image & WAM (2) & 32 & none & 1.000 & 1.000 & 1.000 & 0.964 \\
Image & TrustMark-P (0.8) & 100 & BCH\_5 & 0.977 & 1.000 & 0.847 & 0.721 \\
Image & TrustMark-P (1) & 100 & BCH\_5 & 0.990 & 1.000 & 0.976 & 0.794 \\
Image & TrustMark-P (1.5) & 100 & BCH\_5 & 0.998 & 1.000 & 1.000 & 0.841 \\
Image & TrustMark-B (1) & 100 & BCH\_5 & 0.999 & 1.000 & 1.000 & 0.855 \\
Image & TrustMark-Q (1) & 100 & BCH\_5 & 0.999 & 1.000 & 1.000 & 0.852 \\
Image & TrustMark-C (1) & 100 & BCH\_5 & 0.995 & 1.000 & 1.000 & 0.848 \\
Image & TrustMark-Q (1.2) & 100 & BCH\_5 & 1.000 & 1.000 & 1.000 & 0.854 \\
Image & TrustMark-Q (1.3) & 100 & BCH\_5 & 1.000 & 1.000 & 1.000 & 0.854 \\
Image & TrustMark-Q (1.4) & 100 & BCH\_5 & 1.000 & 1.000 & 1.000 & 0.854 \\
Image & TrustMark-Q (1.5) & 100 & BCH\_5 & 1.000 & 1.000 & 1.000 & 0.855 \\
Image & DWT-DCT-SVD & 64 & none & 0.974 & 0.953 & 0.576 & 0.280 \\
Image & ChunkySeal (0.2) & 1024 & none & 0.934 & 0.941 & 0.035 & 0.134 \\
Image & DWT-DCT & 64 & none & 0.518 & 0.012 & 0.000 & 0.142 \\
Image & InvisMark\textsuperscript{\dag} & 100 & none & 0.919 & 1.000 & 0.000 & 0.007 \\
Image & PixelSeal (0.1) & 256 & none & 0.907 & 0.988 & 0.024 & 0.016 \\
Image & PixelSeal (0.125) & 256 & none & 0.965 & 0.988 & 0.106 & 0.095 \\
Image & RivaGAN\textsuperscript{\dag} & 32 & none & 0.996 & 1.000 & 0.900 & 0.788 \\
Image & Video Seal (0.1) & 256 & none & 0.887 & 1.000 & 0.000 & 0.000 \\
\midrule
Audio & AudioSeal base (1) & 16 & none & 0.997 & -- & 0.978 & 0.980 \\
Audio & AudioSeal streaming (0.5) & 16 & none & 0.996 & -- & 0.986 & 0.842 \\
Audio & AudioSeal streaming (1) & 16 & none & 0.997 & -- & 0.993 & 0.894 \\
Audio & AudioSeal streaming (1.1) & 16 & none & 0.996 & -- & 0.993 & 0.894 \\
Audio & WavMark & 16 & none & 1.000 & -- & 1.000 & 0.993 \\
Audio & SilentCipher 44.1k\textsuperscript{\dag} & 40 & none & 1.000 & 1.000 & 1.000 & 0.841 \\
\midrule
Video & PixelSeal (0.4) & 256 & none & 0.999 & 1.000 & 0.833 & 0.771 \\
Video & Video Seal (0.4) & 256 & none & 1.000 & 1.000 & 1.000 & 0.875 \\
Video & Video Seal (0.7) & 256 & none & 0.997 & 1.000 & 0.500 & 0.458 \\
Video & ChunkySeal (0.2) & 1024 & none & 0.891 & 1.000 & 0.000 & 0.021 \\
Video & PixelSeal (0.2) & 256 & none & 0.984 & 1.000 & 0.667 & 0.583 \\
Video & Video Seal (0.2) & 256 & none & 0.986 & 1.000 & 0.667 & 0.500 \\
Video & Video Seal (1.1) & 256 & none & 0.954 & 1.000 & 0.000 & 0.000 \\
\bottomrule
\end{tabular}
\par\vspace{2pt}\parbox{\linewidth}{\raggedright\scriptsize Bits: transmitted bits. TrustMark's 100-bit BCH\_5 codeword carries a 61-bit payload; bit accuracy and $\geq k$ use the 100 transmitted bits, and exact recovery compares the 61 decoded payload bits. Perth carries no payload and is omitted. \dag\ Not openly licensed.}
\end{table*}

\begin{figure*}[!t]
\centering
\includegraphics[width=\textwidth]{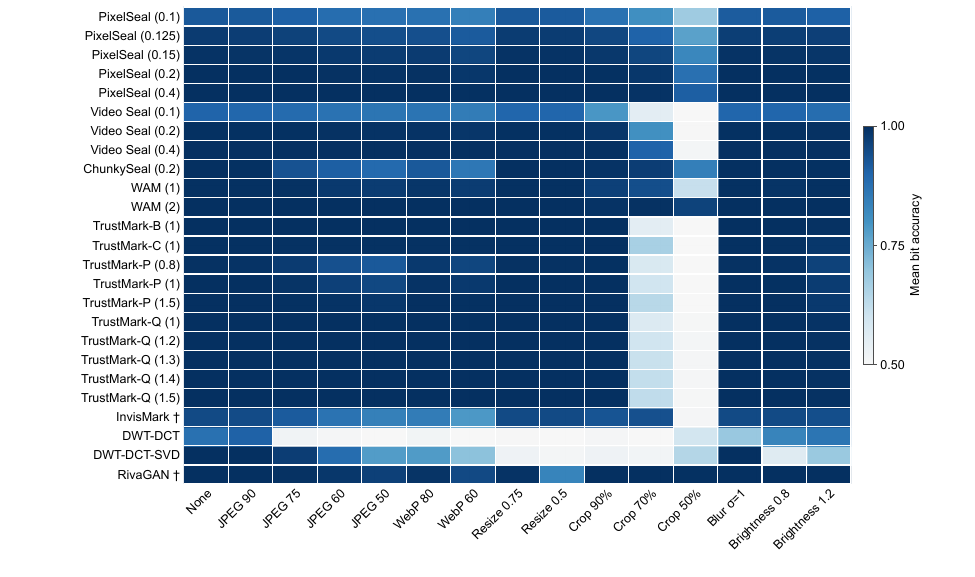}
\caption{Image robustness: mean bit accuracy of every configuration (rows) after every transformation (columns), from 0.5 (chance, white) to 1 (dark blue). \dag~Reference-only methods. The full values are in the aggregated results (Data and Code Availability).}
\label{fig:wm-image-rob}
\end{figure*}

Fig.~\ref{fig:wm-image-rob} shows where the methods differ most: under geometric change. Retaining 50\% of each side (25\% of the area) reduced every TrustMark variant and Video Seal to chance (for example \val{wm:image/trustmark-Q@1.2/rob/crop50} for TrustMark-Q at 1.2 and \val{wm:image/VideoSeal-1.0@default/rob/crop50} for Video Seal). PixelSeal at 0.15 kept \val{wm:image/PixelSeal@0.15/rob/crop50}, and WAM, which predicts a per-pixel watermark map, kept \val{wm:image/WAM-MIT@default/rob/crop50}. A 70\% crop already defeated TrustMark (\val{wm:image/trustmark-Q@1.2/rob/crop70}). Compression, resizing, blur and brightness changes were survived by all qualifying configurations.

Video Seal at 0.2 is the fastest and cheapest qualifying configuration, at \val{wm:image/VideoSeal-1.0@default/gpu}~ms per megapixel against \val{wm:image/PixelSeal@0.15/gpu} for PixelSeal. This 4\% difference is below the resolution of our timing (Section~\ref{sec:wm-cost}), so on speed and cost the two networks are equivalent. Among the other qualifying methods, WAM embeds at about twice the cost of these two and TrustMark at five to seven times. The classical DWT-DCT-SVD qualifies, but with lower quality (FLIP \val{wm:image/DWT-DCT-SVD/flip}). The plain DWT-DCT does not survive JPEG 75 (\val{wm:image/DWT-DCT/gate}).

Three methods fail for different reasons. ChunkySeal carries 1{,}024 bits but loses too many of them on Kodak after JPEG 75 (\val{wm:image/ChunkySeal@default/worst}); it is also about five times slower than PixelSeal on the GPU and needs \val{wm:image/ChunkySeal@default/cpu}~ms per megapixel on the CPU. InvisMark, a reference-only method, has the highest ColorVideoVDP score (\val{wm:image/InvisMark/cvvdp}) but fails the gate (\val{wm:image/InvisMark/gate}). RivaGAN survives JPEG but embeds about 600 times more slowly than PixelSeal.

\begin{figure*}[!t]
\centering
\includegraphics[width=\textwidth]{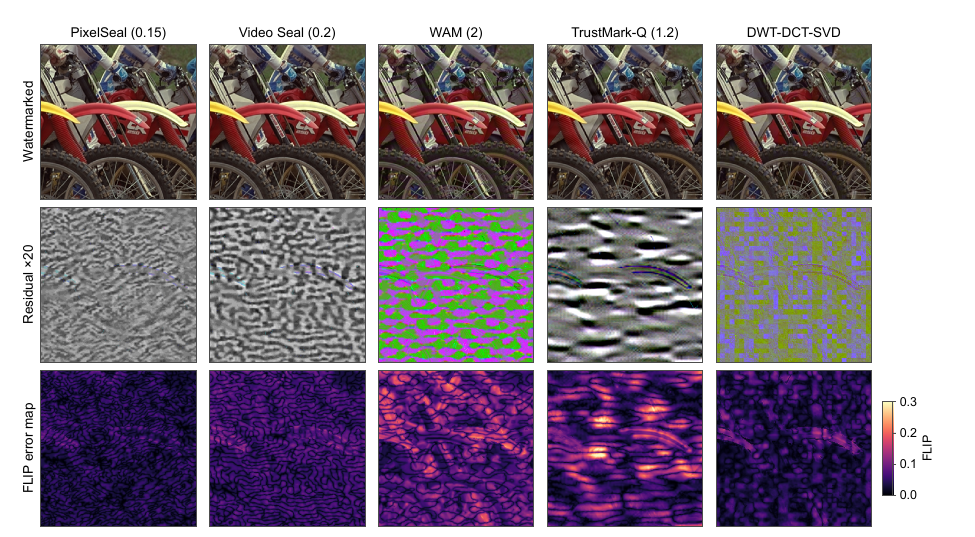}
\caption{Visual comparison on a 256$\times$256 crop of Kodak image 5. Top: watermarked crop. Middle: residual (marked minus original) amplified 20 times around mid-grey. Bottom: FLIP error map (0 to 0.3). Strengths are those of Table~\ref{tab:wm-image}.}
\label{fig:wm-gallery}
\end{figure*}

Fig.~\ref{fig:wm-gallery} shows the residuals behind these numbers. PixelSeal and Video Seal add fine, texture-like patterns whose FLIP error stays low across the image. TrustMark-Q's residual has larger blobs, which produce concentrated FLIP error in flat regions. WAM's residual is strongly coloured.

\begin{table*}[!t]
\centering
\scriptsize
\caption{Audio watermarking on 138 inputs: 46 source clips (40 LibriSpeech utterances, 6 music excerpts), each rendered at 16, 44.1 and 48~kHz. Means over inputs. PESQ is a speech model: the primary PESQ is the mean over the 120 speech inputs, and music PESQ (18 inputs) is shown for reference only. Bit accuracy after MP3 at 128~kb/s.}
\label{tab:wm-audio}
\setlength{\tabcolsep}{3.4pt}
\begin{tabular}{lcrrrrrrrrrrrr}
\toprule
Configuration & Q & SNR & SI-SNR & \multicolumn{2}{c}{PESQ} & STOI & $|\Delta$LUFS$|$ & \multicolumn{2}{c}{MP3 128k bit acc.} & Blind & GPU & Cost & Comb. \\
  &   & (dB)$\uparrow$ & (dB)$\uparrow$ & speech$\uparrow$ & music & $\uparrow$ & $\downarrow$ & pooled & worst set & FPR$\downarrow$ & (ms/s) & (\$/kh) & $\uparrow$ \\
\midrule
AudioSeal base (1)\textsuperscript{\costmark}\textsuperscript{\speedmark}\textsuperscript{\bestmark} & \checkmark & 27.3 & 27.0 & 4.44 & 4.23 & 0.9967 & 0.013 & 0.997 & 0.979 & 0.000 & \textbf{4.90} & \textbf{4.9} & \textbf{0.98} \\
AudioSeal streaming (0.5) & \checkmark & 32.9 & 32.7 & \textbf{4.55} & 4.56 & \textbf{0.9999} & \textbf{0.004} & 0.996 & 0.988 & 0.000 & 5.83 & 5.9 & 0.91 \\
AudioSeal streaming (1) & \checkmark & 26.9 & 26.6 & 4.39 & 4.45 & 0.9998 & 0.018 & 0.997 & 0.989 & 0.000 & 5.83 & 5.9 & 0.90 \\
AudioSeal streaming (1.1) & \checkmark & 26.0 & 25.8 & 4.36 & 4.43 & 0.9997 & 0.025 & 0.996 & 0.988 & 0.000 & 5.83 & 5.9 & 0.90 \\
WavMark & \checkmark & \textbf{36.3} & \textbf{35.9} & 4.15 & 4.36 & 0.9932 & 0.008 & \textbf{1.000} & \textbf{1.000} & 0.000 & 33.5 & 34 & 0.37 \\
SilentCipher 44.1k\textsuperscript{\dag} & -- & 44.5 & 44.1 & 4.59 & 4.59 & 0.9987 & 0.009 & 1.000 & 1.000 & 0.002 & 7.09 & 7.1 & -- \\
Perth & -- & 15.5 & 15.8 & 4.38 & 4.48 & 0.9931 & 0.245 & -- & -- & 0.043 & 1.59 & 1.3\textsuperscript{c} & -- \\
\bottomrule
\end{tabular}
\par\vspace{2pt}\parbox{\linewidth}{\raggedright\scriptsize \costmark\ lowest cost, \speedmark\ fastest GPU embedding, \bestmark\ highest combined score, each among qualifying configurations (Q). Ties between strengths of one network are broken by quality. Bold: best qualifying value per column. \dag\ Code or weights lack an open licence; reported for reference and excluded from ranking. Perth carries no bit payload, so it has no bit accuracy and cannot qualify; its detection rate is reported in the text. Cost: estimated USD per 1{,}000 hours of audio. GPU: ms of compute per second of audio. Blind FPR: share of unmarked decodes (nine per clip) on which the model's own detector fires; Table~\ref{tab:wm-fpr} gives it per source.}
\end{table*}

\begin{figure*}[!t]
\centering
\includegraphics[width=\textwidth]{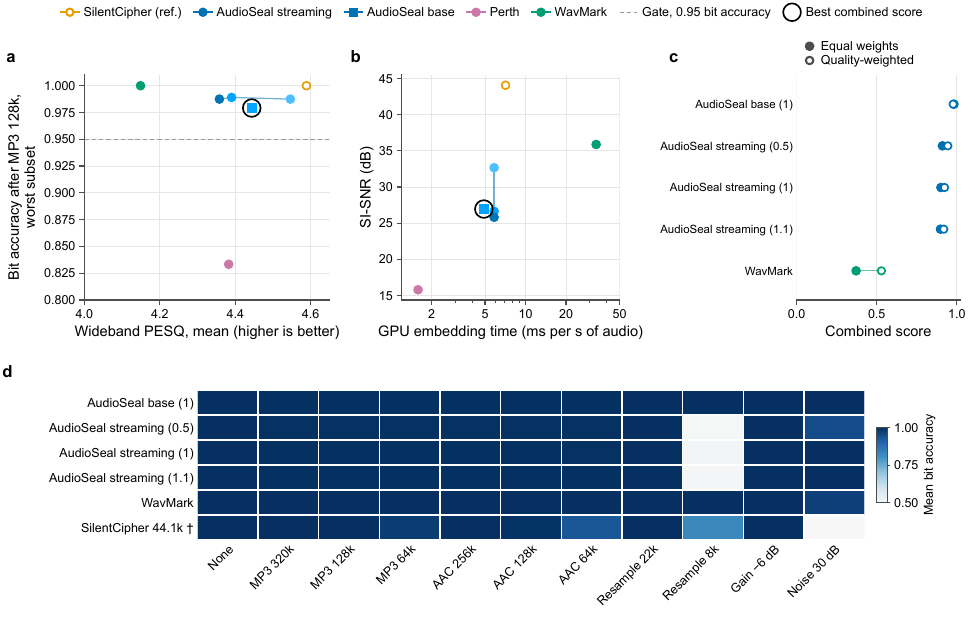}
\caption{Audio watermarking. (a) Wideband PESQ on the speech inputs against bit accuracy after MP3 at 128~kb/s on the worst subset (source and sample rate). (b) GPU embedding time against SI-SNR. (c) Combined score of qualifying configurations. (d) Mean bit accuracy after each transformation; Perth is omitted because it carries no payload. \dag~Reference only. Values are in Table~\ref{tab:wm-audio}.}
\label{fig:wm-audio}
\end{figure*}

\subsection{Audio}
\label{sec:wm-res-audio}
AudioSeal leads the audio comparison (Table~\ref{tab:wm-audio}, Fig.~\ref{fig:wm-audio}). The base model has the highest combined score (\val{wm:audio/AudioSeal-base@1/comb}). It is also the fastest and cheapest qualifying model, at \val{wm:audio/AudioSeal-base@1/gpu}~ms of GPU time per second of audio. The streaming model at strength 0.5 has the best quality of any qualifying configuration, with PESQ on speech \val{wm:audio/AudioSeal-streaming@0.5/pesq} and SI-SNR \val{wm:audio/AudioSeal-streaming@0.5/si_snr}~dB, against \val{wm:audio/AudioSeal-base@1/pesq} and \val{wm:audio/AudioSeal-base@1/si_snr}~dB for the base model. The PESQ difference on speech is significant (Table~\ref{tab:wm-tests}). On the \val{wm:audio/n_music} music inputs, where PESQ is outside its design scope, the two score \val{wm:audio/AudioSeal-streaming@0.5/pesq_music} and \val{wm:audio/AudioSeal-base@1/pesq_music}. The quality-weighted score still favours the base model, because the streaming model was about 20\% slower in our measurements.

Raising the streaming model's strength from 0.5 to 1.0 cost \val{wm:audio/AudioSeal-streaming@0.5/pesq}$\to$\val{wm:audio/AudioSeal-streaming@1/pesq} in PESQ and about 6~dB of SI-SNR, and brought no measurable gain in gate robustness (\val{wm:audio/AudioSeal-streaming@0.5/gate} against \val{wm:audio/AudioSeal-streaming@1/gate}). The two AudioSeal models differ in one respect that the gate does not capture: after a round trip through 8~kHz, the streaming model falls to chance (\val{wm:audio/AudioSeal-streaming@0.5/rob/resample_8k}), while the base model keeps \val{wm:audio/AudioSeal-base@1/rob/resample_8k}. Content that may pass through narrowband telephony therefore needs the base model.

The codec results above use the benchmark's alignment of each round trip to the original, which a real verifier cannot perform. We therefore repeated the six MP3 and AAC round trips for the openly licensed payload models on a second host (\val{wm:noalign/host}) and decoded each output twice, aligned and unaligned (Table~\ref{tab:wm-noalign}). The aligned re-run reproduced the published bit accuracy to within 0.001. Without alignment, AudioSeal and WavMark decoded as well as with it: pooled over the six codecs, AudioSeal base kept \val{wm:noalign/AudioSeal-base@1/all/unaligned} bit accuracy unaligned against \val{wm:noalign/AudioSeal-base@1/all/aligned} aligned, and WavMark \val{wm:noalign/WavMark/all/unaligned} against \val{wm:noalign/WavMark/all/aligned}. Both models locate the mark themselves (AudioSeal decodes frame by frame, WavMark searches for its synchronisation pattern), so codec delay does not hurt them, and the alignment step does not inflate their reported robustness.

\begin{table*}[!t]
\centering
\footnotesize
\caption{Audio decoding after codec round trips with and without alignment to the original. Pub.: the published bit accuracy (Table~\ref{tab:wm-audio}, aligned, benchmark host). Al./Un.: re-run on one host, decoded after cross-correlation alignment to the original (the benchmark protocol) or from the codec output cut to the original length with no offset search. All: the six MP3 and AAC settings pooled; exact: all payload bits correct.}
\label{tab:wm-noalign}
\setlength{\tabcolsep}{3.4pt}
\begin{tabular}{lrrrrrrrrrr}
\toprule
 & \multicolumn{3}{c}{MP3 128k bit acc.} & \multicolumn{3}{c}{AAC 64k bit acc.} & \multicolumn{2}{c}{All codecs bit acc.} & \multicolumn{2}{c}{All codecs exact} \\
Configuration & Pub. & Al. & Un. & Pub. & Al. & Un. & Al. & Un. & Al. & Un. \\
\midrule
AudioSeal base (1) & 0.997 & 0.997 & 0.999 & 0.998 & 0.998 & 0.999 & 0.997 & 0.999 & 0.978 & 0.986 \\
AudioSeal streaming (0.5) & 0.996 & 0.996 & 1.000 & 0.996 & 0.995 & 0.998 & 0.996 & 0.999 & 0.983 & 0.990 \\
AudioSeal streaming (1.1) & 0.996 & 0.997 & 1.000 & 0.997 & 0.996 & 1.000 & 0.997 & 1.000 & 0.993 & 1.000 \\
AudioSeal streaming (1) & 0.997 & 0.997 & 1.000 & 0.995 & 0.995 & 1.000 & 0.997 & 1.000 & 0.993 & 1.000 \\
WavMark & 1.000 & 1.000 & 1.000 & 1.000 & 1.000 & 1.000 & 1.000 & 1.000 & 1.000 & 1.000 \\
\bottomrule
\end{tabular}
\par\vspace{2pt}\parbox{\linewidth}{\raggedright\scriptsize Re-run on Apple M5 Pro with the speech inputs byte-identical to the benchmark corpus; the music inputs were re-decoded from the same source files by a different FFmpeg build (same content, different bytes). Perth carries no payload and is omitted.}
\end{table*}

WavMark is the most robust model we tested, with bit accuracy \val{wm:audio/WavMark/worst} after MP3 on every subset. It has lower PESQ (\val{wm:audio/WavMark/pesq}) and embeds six to seven times more slowly than AudioSeal. SilentCipher, a reference-only model, has the highest quality of all (SI-SNR \val{wm:audio/SilentCipher-44.1k/si_snr}~dB). Perth carries no payload; its detector fires on \val{wm:audio/Perth/blindpct}\% of unmarked clips, and after MP3 it detects the mark in only \val{wm:audio/Perth/worst} of the clips in its worst subset.

\begin{figure*}[!t]
\centering
\includegraphics[width=\textwidth]{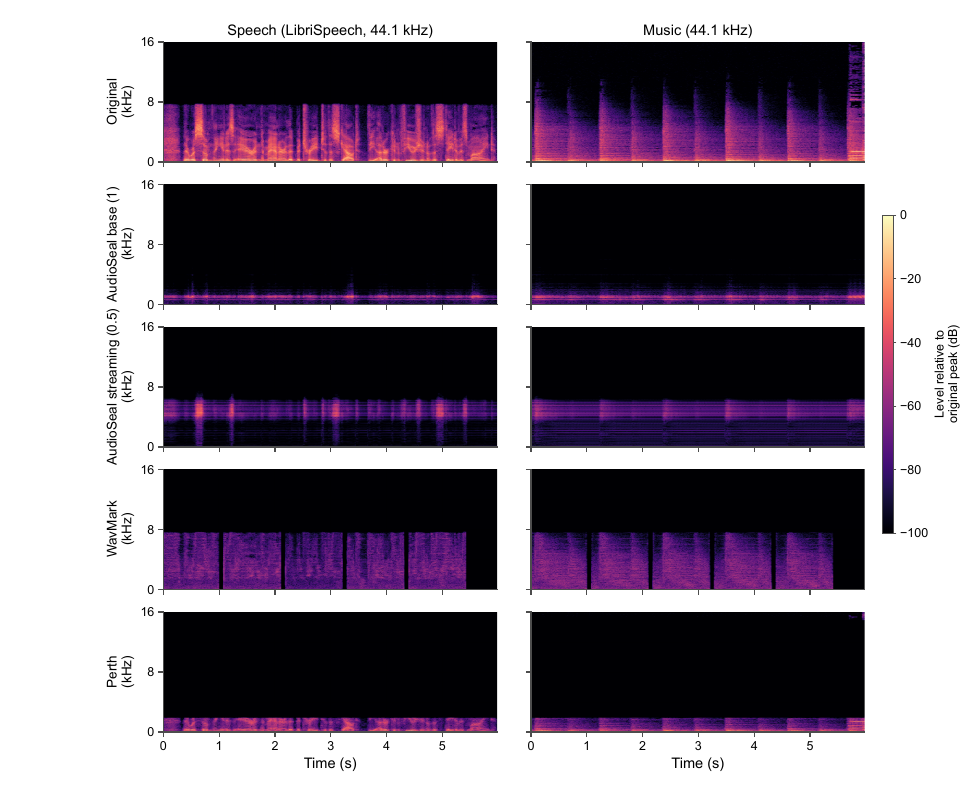}
\caption{Where the audio watermarks put their energy. Top row: spectrograms of the first 6~s of one speech clip (LibriSpeech, band-limited to 8~kHz by its source) and one music clip, both at 44.1~kHz. Other rows: spectrogram of the residual (marked minus original) for each model at the strength of Table~\ref{tab:wm-audio}, on the same decibel scale relative to the original's peak. SilentCipher is omitted because its reference environment was not available when these clips were re-embedded.}
\label{fig:wm-audio-res}
\end{figure*}

Fig.~\ref{fig:wm-audio-res} shows why the models respond differently to the same transformations. The AudioSeal base model concentrates its residual below about 2~kHz, which a round trip through 8~kHz preserves. The streaming model places most of its energy in a narrow band between about 4 and 6~kHz, above the 4~kHz Nyquist limit of that round trip, which is consistent with its failure there. WavMark spreads its residual across the band in one-second blocks, and Perth adds the strongest low-frequency residual of the four, in line with its low SI-SNR.

\begin{table*}[!t]
\centering
\footnotesize
\caption{Video watermarking on six 2.5-s 1080p clips. Bit accuracy after H.264 at CRF 23, pooled and on the worst clip.}
\label{tab:wm-video}
\setlength{\tabcolsep}{3.4pt}
\begin{tabular}{lcrrrrrrrrrrr}
\toprule
Configuration & Q & PSNR & SSIM & LPIPS & FLIP & VMAF & \multicolumn{2}{c}{H.264 CRF 23 bit acc.} & GPU & fps & Cost & Comb. \\
  &   & (dB)$\uparrow$ & $\uparrow$ & $\downarrow$ & $\downarrow$ & $\uparrow$ & pooled & worst clip & (ms/MP) & 1080p & (\$/k MP) & $\uparrow$ \\
\midrule
PixelSeal (0.4)\textsuperscript{\costmark}\textsuperscript{\speedmark}\textsuperscript{\bestmark} & \checkmark & 40.1 & 0.9674 & \textbf{0.0130} & \textbf{0.054} & \textbf{86.0} & 0.999 & 0.992 & \textbf{15.4} & 31.3 & \textbf{0.0043} & \textbf{1.00} \\
Video Seal (0.4) & \checkmark & \textbf{40.3} & \textbf{0.9886} & 0.0131 & 0.055 & 82.1 & \textbf{1.000} & \textbf{1.000} & 16.3 & 29.7 & 0.0045 & 0.96 \\
Video Seal (0.7) & \checkmark & 35.5 & 0.9710 & 0.0361 & 0.083 & 66.6 & 0.997 & 0.992 & 16.3 & 29.7 & 0.0045 & 0.91 \\
ChunkySeal (0.2) & -- & 44.5 & 0.9734 & 0.0151 & 0.048 & 96.4 & 0.891 & 0.652 & 23.7 & 20.4 & 0.0066 & -- \\
PixelSeal (0.2) & -- & 46.0 & 0.9869 & 0.0035 & 0.031 & 95.5 & 0.984 & 0.949 & 15.4 & 31.3 & 0.0043 & -- \\
Video Seal (0.2) & -- & 46.2 & 0.9964 & 0.0035 & 0.032 & 94.0 & 0.986 & 0.922 & 16.3 & 29.7 & 0.0045 & -- \\
Video Seal (1.1) & -- & 31.6 & 0.9403 & 0.0781 & 0.116 & 54.6 & 0.954 & 0.902 & 16.3 & 29.7 & 0.0045 & -- \\
\bottomrule
\end{tabular}
\par\vspace{2pt}\parbox{\linewidth}{\raggedright\scriptsize \costmark\ lowest cost, \speedmark\ fastest GPU embedding, \bestmark\ highest combined score, each among qualifying configurations (Q). Ties between strengths of one network are broken by quality. Bold: best qualifying value per column. \dag\ Code or weights lack an open licence; reported for reference and excluded from ranking. fps: embedding throughput for 1080p frames on one A10G. Cost: estimated USD per 1{,}000 megapixels of 1080p video (GPU only; no CPU run).}
\end{table*}

\begin{figure*}[!t]
\centering
\includegraphics[width=\textwidth]{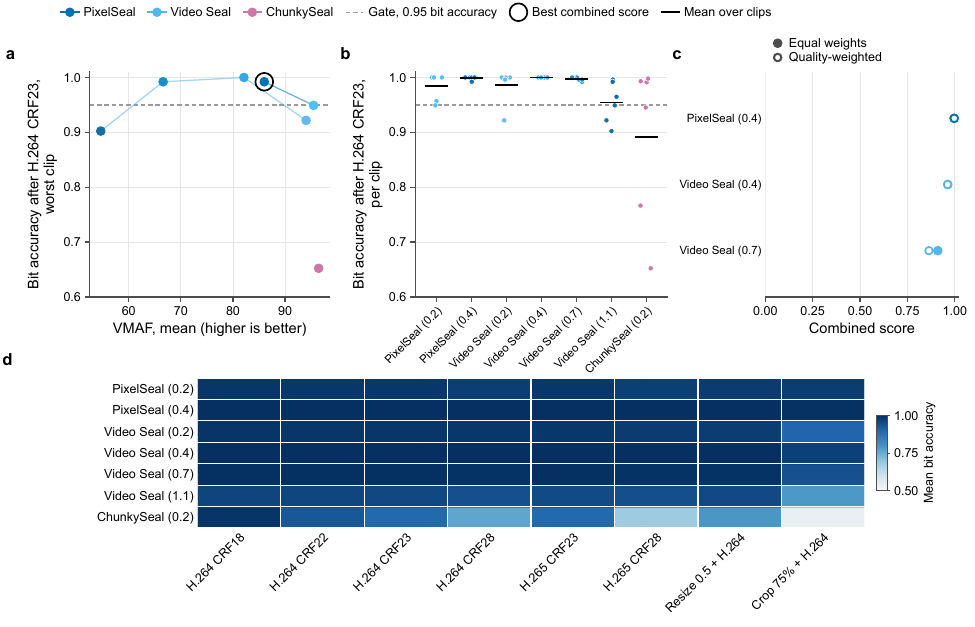}
\caption{Video watermarking. (a) VMAF against bit accuracy after H.264 at CRF 23 on the worst clip. (b) Bit accuracy after H.264 at CRF 23 for each clip (points) and the mean over clips (bars); defaults fail on one clip each. (c) Combined score of qualifying configurations. (d) Mean bit accuracy after each transformation. Values are in Table~\ref{tab:wm-video}.}
\label{fig:wm-video}
\end{figure*}

\subsection{Video}
\label{sec:wm-res-video}
At their default strength of 0.2, PixelSeal and Video Seal both pass the pooled gate but each fails on one clip (Fig.~\ref{fig:wm-video}b): PixelSeal on \emph{crowd run} (\val{wm:video/PixelSeal@default/worst}) and Video Seal on \emph{aspen} (\val{wm:video/VideoSeal-1.0@default/worst}). Only strengths of 0.4 and above pass on every clip. PixelSeal at 0.4 has the highest combined score (\val{wm:video/PixelSeal@0.4/comb}), with VMAF \val{wm:video/PixelSeal@0.4/vmaf} against \val{wm:video/VideoSeal-1.0@0.4/vmaf} for Video Seal at the same strength. With six clips this difference is descriptive only (Table~\ref{tab:wm-tests}). It is also the fastest and cheapest qualifying configuration, at \val{wm:video/PixelSeal@0.4/fps} frames per second for 1080p on one A10G. That corresponds to about \$\val{wm:video/PixelSeal@0.4/costhour} of on-demand GPU time per hour of 1080p video at 30 frames per second.

Strength does not increase robustness monotonically. Video Seal at 1.1 is less robust than at 0.4 after every transformation, with pooled bit accuracy after CRF 23 of \val{wm:video/VideoSeal-1.0@1.1/gate} against \val{wm:video/VideoSeal-1.0@0.4/gate}. Its quality is also much worse, with PSNR \val{wm:video/VideoSeal-1.0@1.1/psnr}~dB and VMAF \val{wm:video/VideoSeal-1.0@1.1/vmaf}. A strong residual is itself degraded by the codec, and the detector was not trained at that strength. ChunkySeal fails in video as it did in images (\val{wm:video/ChunkySeal@default/worst} on its worst clip).

\begin{figure*}[!t]
\centering
\includegraphics[width=\textwidth]{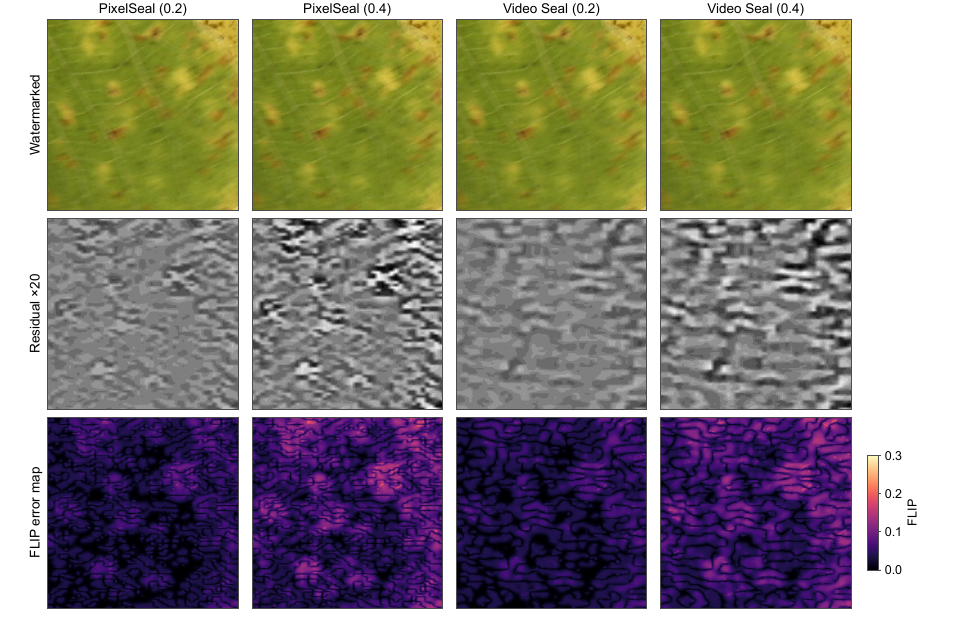}
\caption{Video residuals on one 1080p frame of \emph{aspen} after video-mode embedding (frame 12 of 24). Top: 320$\times$320 crop of the watermarked frame. Middle: residual amplified 20 times around mid-grey. Bottom: FLIP error map (0 to 0.3).}
\label{fig:wm-video-res}
\end{figure*}

Fig.~\ref{fig:wm-video-res} shows the cost of the stronger setting on a single frame. Doubling the strength from 0.2 to 0.4 makes the residual of both models visibly coarser and raises the FLIP error across the frame. PixelSeal's residual is finer-grained than Video Seal's.

\begin{table*}[!t]
\centering
\scriptsize
\caption{False positives on unmarked content. Theoretical: the per-key false-match probability $P_{\mathrm{FM}}$ of an expected-payload test that requires $k$ of $n$ bits to match, if unmarked content decodes to independent fair bits. Own payload: false matches against the payload recorded for the item, the check a verifier makes. Random payloads: the same decodes compared with unrelated payloads, which tests agreement with payloads the content was never marked with, conditional on these decodes. Blind detector: false detections of the model's own detector.}
\label{tab:wm-fpr}
\setlength{\tabcolsep}{3.4pt}
\begin{tabular}{lrrrrrrrrrrl}
\toprule
Network & \multicolumn{2}{c}{Theoretical} & \multicolumn{3}{c}{Own payload} & \multicolumn{2}{c}{Random payloads} & \multicolumn{3}{c}{Blind detector} & Evidence \\
\cmidrule(lr){2-3}\cmidrule(lr){4-6}\cmidrule(lr){7-8}\cmidrule(lr){9-11}
  & $k/n$ & $P_{\mathrm{FM}}$ & trials & sources & 95\% upper & trials & max & trials & sources & 95\% upper &   \\
\midrule
\multicolumn{12}{l}{\emph{Image}} \\
ChunkySeal & 589/1024 & $8.3\times10^{-7}$ & 0/340 & 0/85 & 0.042 & 0/17000 & 566 & \multicolumn{3}{c}{none} & analytic \\
InvisMark\textsuperscript{\dag} & 74/100 & $8.3\times10^{-7}$ & 0/340 & 0/85 & 0.042 & 0/17000 & 70 & \multicolumn{3}{c}{none} & analytic \\
PixelSeal & 167/256 & $6.2\times10^{-7}$ & 0/340 & 0/85 & 0.042 & 0/17000 & 158 & \multicolumn{3}{c}{none} & analytic \\
RivaGAN\textsuperscript{\dag} & 30/32 & $1.2\times10^{-7}$ & 0/240 & 0/60 & 0.060 & 0/12000 & 26 & \multicolumn{3}{c}{none} & analytic \\
TrustMark-B & 74/100 & $8.3\times10^{-7}$ & 0/340 & 0/85 & 0.042 & 0/17000 & 71 & 13/340 & 10/85 & 0.21 & analytic \\
TrustMark-C & 74/100 & $8.3\times10^{-7}$ & 0/340 & 0/85 & 0.042 & 0/17000 & 69 & 15/340 & 13/85 & 0.25 & analytic \\
TrustMark-P & 74/100 & $8.3\times10^{-7}$ & 0/340 & 0/85 & 0.042 & 0/17000 & 69 & 7/340 & 5/85 & 0.13 & analytic \\
TrustMark-Q & 74/100 & $8.3\times10^{-7}$ & 0/340 & 0/85 & 0.042 & 0/17000 & 69 & 9/340 & 8/85 & 0.18 & analytic \\
Video Seal & 167/256 & $6.2\times10^{-7}$ & 0/340 & 0/85 & 0.042 & 0/17000 & 161 & \multicolumn{3}{c}{none} & analytic \\
WAM & 30/32 & $1.2\times10^{-7}$ & 0/340 & 0/85 & 0.042 & 0/17000 & 26 & \multicolumn{3}{c}{none} & analytic \\
DWT-DCT & 51/64 & $9.4\times10^{-7}$ & 0/340 & 0/85 & 0.042 & 0/17000 & 46 & \multicolumn{3}{c}{none} & analytic \\
\midrule
\multicolumn{12}{l}{\emph{Audio}} \\
AudioSeal base & \multicolumn{2}{c}{none (16 bits)} & \multicolumn{5}{c}{--} & 0/414 & 0/46 & 0.077 & observed \\
AudioSeal streaming & \multicolumn{2}{c}{none (16 bits)} & \multicolumn{5}{c}{--} & 0/414 & 0/46 & 0.077 & observed \\
Perth & \multicolumn{2}{c}{no payload} & \multicolumn{5}{c}{--} & 18/414 & 2/46 & 0.15 & fails \\
SilentCipher\textsuperscript{\dag} & 35/40 & $6.9\times10^{-7}$ & 0/414 & 0/46 & 0.077 & 0/20700 & 31 & 1/414 & 1/46 & 0.12 & analytic \\
WavMark & \multicolumn{2}{c}{none (16 bits)} & \multicolumn{5}{c}{--} & 0/414 & 0/46 & 0.077 & observed \\
\midrule
\multicolumn{12}{l}{\emph{Video}} \\
ChunkySeal & 589/1024 & $8.3\times10^{-7}$ & 0/6 & 0/6 & 0.46 & 0/1200 & 566 & \multicolumn{3}{c}{none} & analytic \\
PixelSeal & 167/256 & $6.2\times10^{-7}$ & 0/6 & 0/6 & 0.46 & 0/1200 & 150 & \multicolumn{3}{c}{none} & analytic \\
Video Seal & 167/256 & $6.2\times10^{-7}$ & 0/6 & 0/6 & 0.46 & 0/1200 & 154 & \multicolumn{3}{c}{none} & analytic \\
\bottomrule
\end{tabular}
\par\vspace{2pt}\parbox{\linewidth}{\raggedright\scriptsize Trials are unmarked items with no transform and under three benign ones (images and audio), or the H.264 CRF 18 encode of each unmarked clip (video); strength settings of one network share the detector, so each network appears once. Sources: unmarked source items with at least one false match or detection, over all sources (an audio source is one clip rendered at three sample rates). 95\% upper: upper endpoint of the two-sided exact Clopper-Pearson 95\% interval with the source as the trial unit. Random payloads: false matches over comparisons, using one fixed list of 50 payloads (200 for video) for every item; max: the largest number of matching bits in any own- or random-payload comparison, to be read against $k$. Analytic: an expected-payload test with $P_{\mathrm{FM}}\leq10^{-6}$ exists and no unmarked trial reached $k$. Observed: no such test exists at this payload size, and the evidence is the blind detector's. None ($n$ bits): even an exact match of all $n$ bits has probability $2^{-n}>10^{-6}$.}
\end{table*}

\begin{figure*}[!t]
\centering
\includegraphics[width=\textwidth]{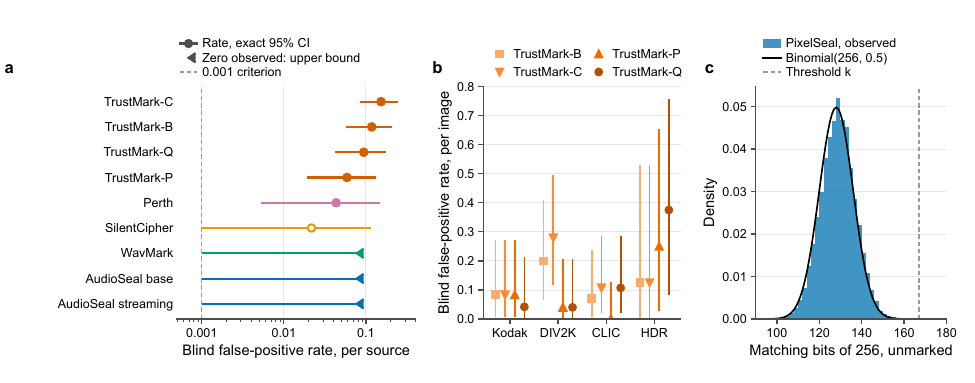}
\caption{False positives on unmarked content. (a) Blind false-positive rate per network with the unmarked source as the unit (a source counts if the detector fired on any of its transforms or renderings) and exact two-sided 95\% intervals; networks with no false detection are shown by their upper bound. The dashed line is the $10^{-3}$ target. (b) TrustMark's blind false-positive rate per image subset, per image. (c) Number of bits in unmarked PixelSeal decodes that match an expected 256-bit payload, against the binomial distribution for independent fair bits and the decision threshold $k$.}
\label{fig:wm-fpr}
\end{figure*}

\subsection{False Positives}
\label{sec:wm-res-fpr}
The blind detectors of all four TrustMark variants report a valid codeword on at least one of the four decodes of \val{wm:tm/blind/min} to \val{wm:tm/blind/max}\% of unmarked images (Table~\ref{tab:wm-fpr}, Fig.~\ref{fig:wm-fpr}a). The rate depends on content: on the 16-bit HDR subset it reached \val{wm:tm/blind/hdrk} of \val{wm:tm/blind/hdrn} images (Fig.~\ref{fig:wm-fpr}b), although the subset is small and the intervals are wide. BCH decoding corrects up to a fixed number of bit errors, so a random 100-bit word lies within the correction radius of some valid codeword with non-negligible probability. A decoder that reports ``found'' whenever correction succeeds cannot have a low false-positive rate.

The same TrustMark decodes, tested against the expected payload with $k=74$ of the 100 transmitted bits, produced no false match on any unmarked image. For PixelSeal, the number of matching bits in unmarked decodes follows the binomial null closely (Fig.~\ref{fig:wm-fpr}c), and no trial came near the threshold. We saw the same for every network with a reachable threshold: the largest number of matching bits in any comparison stayed below $k$ (Table~\ref{tab:wm-fpr}), which is consistent with the independence assumption behind the analytic bound. Table~\ref{tab:wm-fpr} separates these observations from the theoretical rate. For every image network with a reachable threshold, no own-payload trial produced a false match on any of the \val{wm:fpr/image/PixelSeal/own_src_n} unmarked images (\val{wm:fpr/image/RivaGAN/own_src_n} for RivaGAN), and no random-payload comparison reached $k$ either. With the image as the unit, the observed upper bound is \val{wm:fpr/image/PixelSeal/own_src_upper}\%, so the observations show only that the rate is not large; the $10^{-6}$ per-key rate rests on the theoretical value. For the 16-bit audio models the expected-payload test is unavailable. Their blind detectors fired on none of the \val{wm:fpr/audio/WavMark/blind_src_n} unmarked clips (414 decodes), but the resulting upper bound, \val{wm:fpr/audio/WavMark/blind_src_upper}\%, is far above $10^{-3}$, so the target is not established for them; showing it with zero detections would take \val{wm:fpr/sources_for_1e-3} unmarked sources. For video, the six clips give an observed bound of \val{wm:fpr/video/PixelSeal/own_src_upper}\%; there the evidence is the theoretical rate alone.

\begin{table*}[!t]
\centering
\footnotesize
\caption{Embedding cost and speed per network. Unit: one megapixel (image, video) or one second of audio. Cost in USD per 1{,}000 units for image and video, and per 1{,}000 hours for audio, at on-demand prices (g5.xlarge \$1.006/h, m6a.xlarge \$0.1728/h, us-east-1, effective 2026-09-01).}
\label{tab:wm-cost}
\setlength{\tabcolsep}{3.4pt}
\begin{tabular}{llrlrrrrl}
\toprule
Modality & Network & Configs & Any Q & GPU (ms/unit) & CPU (ms/unit) & GPU \$ & CPU \$ & Cheaper \\
\midrule
Image & InvisMark\textsuperscript{\dag} & 1 & no & 8.73 & -- & 0.0024 & -- & GPU \\
Image & Video Seal\textsuperscript{\costmark}\textsuperscript{\speedmark} & 3 & yes & 8.79 & 544 & 0.0025 & 0.026 & GPU \\
Image & PixelSeal\textsuperscript{\bestmark} & 5 & yes & 9.13 & 1104 & 0.0026 & 0.053 & GPU \\
Image & WAM & 2 & yes & 19.1 & 361 & 0.0053 & 0.017 & GPU \\
Image & TrustMark-P & 3 & yes & 45.1 & 273 & 0.013 & 0.013 & GPU \\
Image & TrustMark-B & 1 & yes & 55.7 & 278 & 0.016 & 0.013 & CPU \\
Image & TrustMark-Q & 5 & yes & 63.5 & 283 & 0.018 & 0.014 & CPU \\
Image & TrustMark-C & 1 & yes & 54.4 & 283 & 0.015 & 0.014 & CPU \\
Image & ChunkySeal & 1 & no & 50.4 & 19069 & 0.014 & 0.92 & GPU \\
Image & DWT-DCT & 2 & yes & 409 & 296 & 0.11 & 0.014 & CPU \\
Image & RivaGAN\textsuperscript{\dag} & 1 & no & 5250 & -- & 1.5 & -- & GPU \\
\midrule
Audio & Perth & 1 & no & 1.59 & 7.72 & 1.6 & 1.3 & CPU \\
Audio & AudioSeal base\textsuperscript{\costmark}\textsuperscript{\speedmark}\textsuperscript{\bestmark} & 1 & yes & 4.90 & 60.8 & 4.9 & 10 & GPU \\
Audio & AudioSeal streaming & 3 & yes & 5.83 & 59.3 & 5.9 & 10 & GPU \\
Audio & SilentCipher\textsuperscript{\dag} & 1 & no & 7.09 & 437 & 7.1 & 75 & GPU \\
Audio & WavMark & 1 & yes & 33.5 & -- & 34 & -- & GPU \\
\midrule
Video & PixelSeal\textsuperscript{\costmark}\textsuperscript{\speedmark}\textsuperscript{\bestmark} & 2 & yes & 15.4 & -- & 0.0043 & -- & GPU \\
Video & Video Seal & 4 & yes & 16.3 & -- & 0.0045 & -- & GPU \\
Video & ChunkySeal & 1 & no & 23.7 & -- & 0.0066 & -- & GPU \\
\bottomrule
\end{tabular}
\par\vspace{2pt}\parbox{\linewidth}{\raggedright\scriptsize Latency is the median over the network's strength settings. \textsuperscript{i}~CPU latency measured on another strength of the same network. Costs exclude I/O, storage, model loading and idle capacity. \costmark\ lowest cost, \speedmark\ fastest GPU embedding, \bestmark\ highest combined score, each among qualifying configurations (Q). Ties between strengths of one network are broken by quality..}
\end{table*}

\begin{figure*}[!t]
\centering
\includegraphics[width=\textwidth]{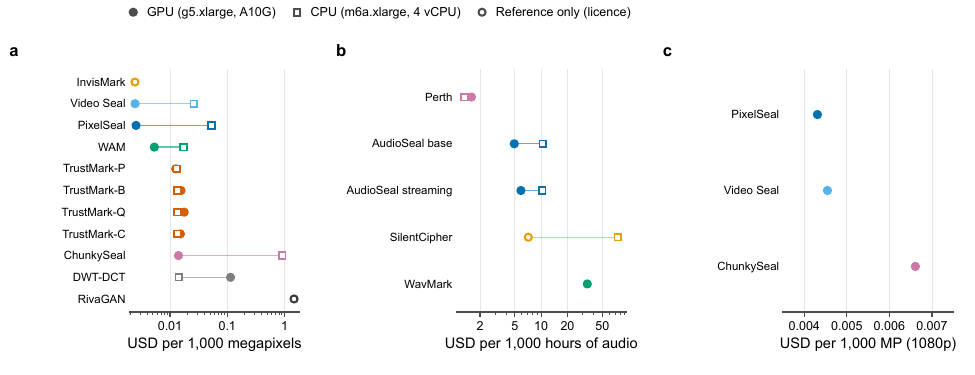}
\caption{Estimated embedding cost per network on GPU (filled circles) and CPU (open squares), at on-demand prices. (a) Images, USD per 1{,}000 megapixels. (b) Audio, USD per 1{,}000 hours. (c) Video, USD per 1{,}000 megapixels of 1080p frames (GPU only). Networks are sorted by their cheaper option. Values are in Table~\ref{tab:wm-cost}.}
\label{fig:wm-cost}
\end{figure*}

\subsection{Cost and Speed}
\label{sec:wm-res-cost}
Table~\ref{tab:wm-cost} and Fig.~\ref{fig:wm-cost} compare networks rather than strength settings. For the neural image methods, the GPU is between one and two orders of magnitude cheaper than the CPU. The exception is TrustMark, whose small network is marginally cheaper on the CPU instance. The classical DWT-DCT methods run only on the CPU. For audio, the GPU is cheaper for every learned model. The cheapest qualifying image network costs about \$\val{wm:image/VideoSeal-1.0@default/cost} per 1{,}000 megapixels, and AudioSeal base about \$\val{wm:audio/AudioSeal-base@1/cost} per 1{,}000 hours of audio. At these prices, compute cost is rarely what decides between methods. The deciding factors are the gate and the false-positive behaviour.

\begin{table*}[!t]
\centering
\footnotesize
\caption{Cross-device determinism of watermark decoding. $\Delta\ell$: difference in detector logits between decodes of the same file.}
\label{tab:wm-stability}
\setlength{\tabcolsep}{3.4pt}
\begin{tabular}{lrrrrcc}
\toprule
Decoded file & Same host & x86-64 vs arm64 & Bit flips & Bit strings & YUV equal & RGB equal \\
  & max $|\Delta\ell|$ & max $|\Delta\ell|$ &   & distinct &   &   \\
\midrule
aspen, strength 0.4, CRF 18 & 0.00096 & 0.459 & 0 & 1 & yes & no \\
aspen, strength 0.4, CRF 22 & 0.00090 & 0.889 & 0 & 1 & yes & no \\
aspen, strength 1.1, CRF 18 & 0.00068 & 0.346 & 2 & 2 & yes & no \\
aspen, strength 1.1, CRF 22 & 0.00076 & 0.440 & 0 & 1 & yes & no \\
controlled burn, strength 0.4, CRF 18 & 0.0016 & 0.306 & 0 & 1 & yes & no \\
controlled burn, strength 0.4, CRF 22 & 0.0017 & 0.284 & 0 & 1 & yes & no \\
controlled burn, strength 1.1, CRF 18 & 0.0014 & 0.206 & 1 & 2 & yes & no \\
controlled burn, strength 1.1, CRF 22 & 0.0015 & 0.289 & 0 & 1 & yes & no \\
red kayak, strength 0.4, CRF 18 & 0.00034 & 0.069 & 0 & 1 & yes & no \\
red kayak, strength 0.4, CRF 22 & 0.00035 & 0.085 & 0 & 1 & yes & no \\
red kayak, strength 1.1, CRF 18 & 0.00036 & 0.067 & 0 & 1 & yes & no \\
red kayak, strength 1.1, CRF 22 & 0.00034 & 0.064 & 0 & 1 & yes & no \\
TrustMark-Q, image after JPEG 75 & 0.070 & 0.070 & 0 & 1 & -- & -- \\
\bottomrule
\end{tabular}
\par\vspace{2pt}\parbox{\linewidth}{\raggedright\scriptsize Same host: CPU (1 and 4 threads), CUDA or MPS on one machine. Across hosts: AMD EPYC (x86-64, Linux) vs Apple M-series (arm64, macOS). YUV/RGB equal: whether decoded frames are byte-identical across hosts before and after the YUV to RGB conversion.}
\end{table*}

\begin{figure}[!t]
\centering
\includegraphics{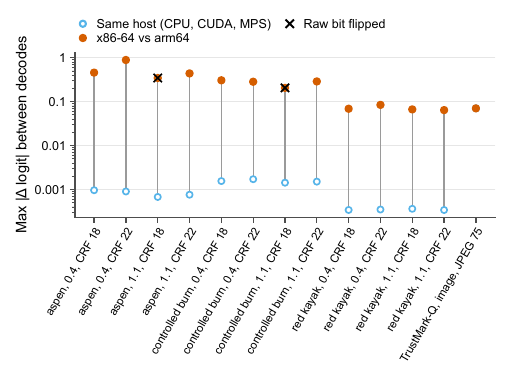}
\caption{Largest difference in detector logits between decodes of the same file on one host (open circles; CPU with one and four threads, and CUDA or MPS) and across an x86-64 and an arm64 host (filled circles). Crosses mark files for which at least one raw bit changed across hosts. Labels give the clip, Video Seal strength and H.264 CRF.}
\label{fig:wm-stability}
\end{figure}

\subsection{Cross-Device Determinism}
\label{sec:wm-res-stability}
Decoding the same video file on one machine gave logits that agreed to within \val{wm:stab/maxsame} across CPU thread counts and the accelerator (Table~\ref{tab:wm-stability}, Fig.~\ref{fig:wm-stability}). The TrustMark-Q image behaves differently: its CPU decodes agreed to within \val{wm:stab/tm/cpu_dlogit} on both hosts, and only the CUDA decode departed from them, by up to \val{wm:stab/tm/maxsame}, which changed no bit. Its same-host and cross-host differences in Table~\ref{tab:wm-stability} are that one CUDA difference. Across the x86-64 and arm64 hosts, the logits differed by up to \val{wm:stab/maxcross}. In \val{wm:stab/nflipfiles} of \val{wm:stab/files} files this flipped raw payload bits, up to \val{wm:stab/maxflips} of 256. Both of those files were marked at strength 1.1, where some bits already had logits close to zero.

The decoded YUV frames were byte-identical on the two hosts for every file. The RGB frames produced from them by the same library version were not. The 2$\times$2 follow-up isolates the two steps. Given the same saved RGB arrays, the two hosts produced logits that differed by at most \val{wm:rgb/model_dlogit}, and the backends of one host by at most \val{wm:rgb/backend_dlogit}. Given the other host's RGB conversion of the same files, one host's logits changed by up to \val{wm:rgb/conv_dlogit} and flipped up to \val{wm:rgb/maxflips} bits (in \val{wm:rgb/flippairs} of the 24 file--host combinations). The network is therefore reproducible across the two platforms to within floating-point noise, and the discrepancy comes from the platform-specific colour conversion. A verifier that must return the same bits everywhere should do the YUV-to-RGB conversion in a portable, fully specified implementation, or run the detector on YUV input.

\begin{table*}[!t]
\centering
\footnotesize
\caption{Planned paired comparisons (two-sided Wilcoxon signed-rank on per-source values, Holm-adjusted over the five tests). Audio inputs of one clip are averaged first, so $n$ counts clips; audio PESQ is on speech.}
\label{tab:wm-tests}
\setlength{\tabcolsep}{3.4pt}
\begin{tabular}{lllrrrr}
\toprule
Modality & A & B & Metric & $n$ & median(A$-$B) & $p_{\mathrm{Holm}}$ \\
\midrule
Image & PixelSeal (0.15) & TrustMark-P (0.8) & FLIP & 60 & 0.0019 & 0.062 \\
Image & PixelSeal (0.15) & Video Seal (0.2) & FLIP & 60 & -0.0099 & $<$0.001 \\
Image & PixelSeal (0.15) & TrustMark-Q (1.2) & FLIP & 60 & -0.0212 & $<$0.001 \\
Audio & AudioSeal base (1) & AudioSeal streaming (0.5) & PESQ (speech) & 40 & -0.1072 & $<$0.001 \\
Video & PixelSeal (0.4) & Video Seal (0.4) & VMAF & 6 & 3.6629 & 0.062 \\
\bottomrule
\end{tabular}
\par\vspace{2pt}\parbox{\linewidth}{\raggedright\scriptsize Negative FLIP differences favour A; positive PESQ and VMAF differences favour A.}
\end{table*}

\section{Fingerprint Results}
\label{sec:fp-results}
This section reports the fingerprinting track: image retrieval at calibrated operating points, false matches and registry scale, geometric verification, audio and video retrieval, partial edits and localization, and security. Detection is reported at pair-level false-match rate $10^{-7}$ for images and at the query-level 1\% threshold, the strictest the calibration sets resolve, for audio and video.

\subsection{Image Retrieval}
\label{sec:fp-res-image}
\begin{figure*}[!t]
\centering
\includegraphics[width=\textwidth]{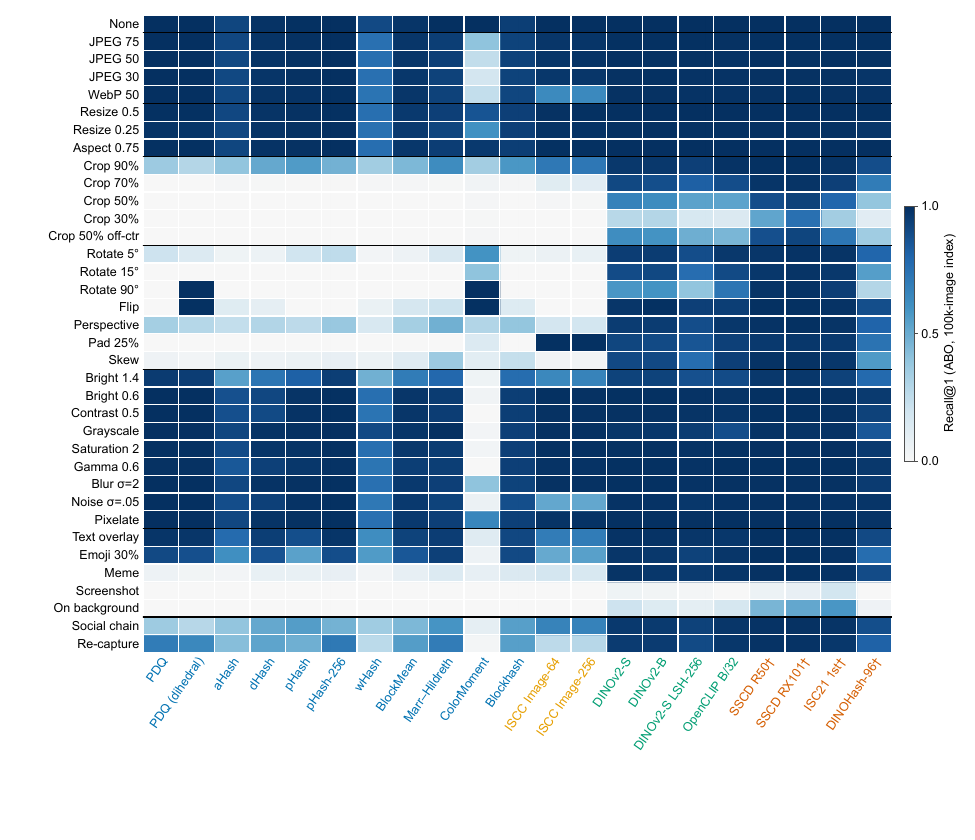}
\caption{Recall@1 of every image method under every transformation, on the ABO registry of \val{fp:count/image/nref} images. Rows are grouped by family (compression, scaling, cropping, geometry, photometric, overlay, chains). Method names are coloured by class; \dag~marks reference-tier methods.}
\label{fig:fp-image-heatmap}
\end{figure*}

Table~\ref{tab:fp-image} summarises the image track and Fig.~\ref{fig:fp-image-heatmap} breaks Recall@1 down by transformation. Three groups separate clearly. The copy-detection descriptors trained on DISC21 lead: at the headline operating point (pair-level FPR $10^{-7}$) the ISC21 winner detects \val{fp:image/ISC21-1st/TPR@pair@1e-07/pct}\% of attacked queries and SSCD-R50 \val{fp:image/SSCD-mixup/TPR@pair@1e-07/pct}\%. The best product-eligible methods are the general-purpose DINOv2 embeddings, at \val{fp:image/DINOv2-B/TPR@pair@1e-07/pct}\% (DINOv2-B) and \val{fp:image/DINOv2-S/TPR@pair@1e-07/pct}\% (DINOv2-S), and the perceptual hashes follow, led by PDQ with dihedral matching at \val{fp:image/PDQ-dihedral/TPR@pair@1e-07/pct}\% and plain PDQ at \val{fp:image/PDQ/TPR@pair@1e-07/pct}\%. The ISCC Image-Code reaches \val{fp:image/ISCC-Image-64/TPR@pair@1e-07/pct}\% at its default 64 bits and \val{fp:image/ISCC-Image-256/TPR@pair@1e-07/pct}\% at 256 bits. All paired differences in Table~\ref{tab:fp-paired} are significant after Holm correction.

\begin{table*}[!t]
\centering
\footnotesize
\caption{Image retrieval over all attacks (ABO; registry of registered originals and distractors). TPR counts a query only if its top-1 is the true asset and the score passes a threshold fixed on held-out calibration negatives: query-level FPR 1\% at this registry size, or pair-level FPR $10^{-7}$ (95\% interval from a bootstrap over sources, which redraws the calibration negatives and so the threshold). Hard: rate at which another photograph of a registered product binds to that product at the 1\% threshold.}
\label{tab:fp-image}
\setlength{\tabcolsep}{3pt}
\begin{tabular}{lcccccc}
\toprule
Method & R@1 & $\mu$AP & TPR$_{q,1\%}$ & TPR$_{p,10^{-7}}$ & Hard FB & DISC21 $\mu$AP \\
\midrule
PDQ & 0.579 & 0.520 & 0.495 & 0.495 \tiny[0.48, 0.50] & 0.000 & 0.295 \\
PDQ (dihedral) & 0.623 & 0.569 & 0.549 & 0.546 \tiny[0.54, 0.55] & 0.000 & 0.367 \\
aHash & 0.503 & 0.189 & 0.000 & 0.000 \tiny[0.00, 0.00] & 0.000 & 0.027 \\
dHash & 0.560 & 0.356 & 0.313 & 0.000 \tiny[0.00, 0.00] & 0.000 & 0.195 \\
pHash & 0.563 & 0.469 & 0.373 & 0.373 \tiny[0.00, 0.38] & 0.000 & 0.240 \\
pHash-256 & 0.589 & 0.520 & 0.494 & 0.492 \tiny[0.48, 0.50] & 0.001 & 0.302 \\
wHash & 0.425 & 0.134 & 0.000 & 0.000 \tiny[0.00, 0.00] & 0.000 & 0.024 \\
BlockMean & 0.562 & 0.338 & 0.337 & 0.000 \tiny[0.00, 0.00] & 0.001 & 0.069 \\
Marr–Hildreth & 0.586 & 0.455 & 0.362 & 0.314 \tiny[0.28, 0.35] & 0.001 & 0.292 \\
ColorMoment & 0.281 & 0.101 & 0.000 & 0.000 \tiny[0.00, 0.00] & 0.000 & 0.000 \\
Blockhash & 0.560 & 0.250 & 0.000 & 0.000 \tiny[0.00, 0.00] & 0.000 & 0.182 \\
ISCC Image-64 & 0.558 & 0.491 & 0.427 & 0.427 \tiny[0.31, 0.43] & 0.001 & 0.269 \\
ISCC Image-256 & 0.561 & 0.507 & 0.466 & 0.458 \tiny[0.45, 0.47] & 0.001 & 0.292 \\
DINOv2-S & 0.877 & 0.703 & 0.625 & 0.612 \tiny[0.60, 0.63] & 0.000 & 0.444 \\
DINOv2-B & 0.872 & 0.703 & 0.653 & 0.650 \tiny[0.64, 0.66] & 0.003 & 0.460 \\
DINOv2-S LSH-256 & 0.832 & 0.664 & 0.538 & 0.519 \tiny[0.50, 0.55] & 0.000 & 0.399 \\
OpenCLIP B/32 & 0.864 & 0.657 & 0.401 & 0.356 \tiny[0.33, 0.40] & 0.002 & 0.234 \\
SSCD R50\reftier & 0.929 & 0.763 & 0.725 & 0.712 \tiny[0.71, 0.73] & 0.007 & 0.795 \\
SSCD RX101\reftier & 0.942 & 0.787 & 0.763 & 0.703 \tiny[0.69, 0.77] & 0.007 & 0.823 \\
ISC21 1st\reftier & 0.919 & 0.815 & 0.762 & 0.755 \tiny[0.75, 0.77] & 0.004 & 0.917 \\
DINOHash-96\reftier & 0.769 & 0.533 & 0.237 & 0.237 \tiny[0.23, 0.24] & 0.000 & 0.212 \\
\bottomrule
\end{tabular}

\end{table*}

\begin{figure}[!t]
\centering
\includegraphics[width=\columnwidth]{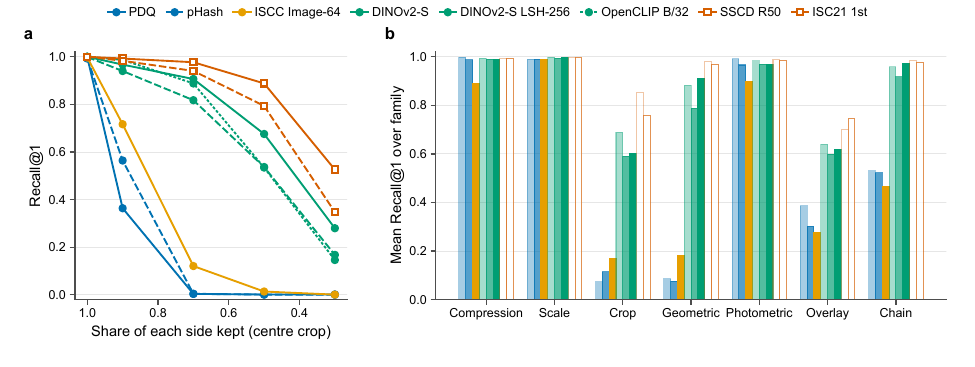}
\caption{(a) Recall@1 against the share of each side kept by a centred crop. (b) Mean Recall@1 over each transformation family for eight representative methods.}
\label{fig:fp-image-families}
\end{figure}

\paragraph{Where each class fails.} The heatmap explains the ranking. Every method, hash or learned, survives compression, rescaling and photometric changes almost perfectly. The hashes fail on geometry: a centred crop keeping 70\% of each side drops PDQ to \val{fp:image/PDQ/att/crop70/pct}\% Recall@1 and the ISCC Image-Code to \val{fp:image/ISCC-Image-64/att/crop70/pct}\%, while DINOv2-S keeps \val{fp:image/DINOv2-S/att/crop70/pct}\% and SSCD \val{fp:image/SSCD-mixup/att/crop70/pct}\%. A 15-degree rotation removes the hashes almost entirely (PDQ \val{fp:image/PDQ/att/rot15/pct}\%). PDQ's dihedral matching recovers flips and 90-degree rotations at no cost in precision (Table~\ref{tab:fp-paired}), which is why it leads the hashes. Two robustness properties are specific to ISCC: its uniform-border trim makes it the only hash that survives 25\% padding (\val{fp:image/ISCC-Image-64/att/pad25/pct}\%), while its $32\times32$ greyscale DCT is, apart from ColorMoment, the hash most affected by additive noise (\val{fp:image/ISCC-Image-64/att/noise0.05/pct}\% against \val{fp:image/PDQ/att/noise0.05/pct}\% for PDQ). Overlays are the hardest family for everyone: when the image is placed onto a screenshot template or onto another photograph at half size, the best method retrieves \val{fp:image/ISC21-1st/att/background0.5/pct}\% of the background overlays (ISC21) and \val{fp:image/ISC21-1st/att/screenshot/pct}\% of the screenshots, the best product-eligible one \val{fp:image/DINOv2-S/att/background0.5/pct}\% and \val{fp:image/DINOv2-S/att/screenshot/pct}\%, and every hash none. Such queries contain the registered image only as a region, and a single global descriptor cannot match it. Fig.~\ref{fig:fp-image-families}a shows the crop curve and Fig.~\ref{fig:fp-image-families}b the per-family means.

\begin{figure*}[!t]
\centering
\includegraphics[width=\textwidth]{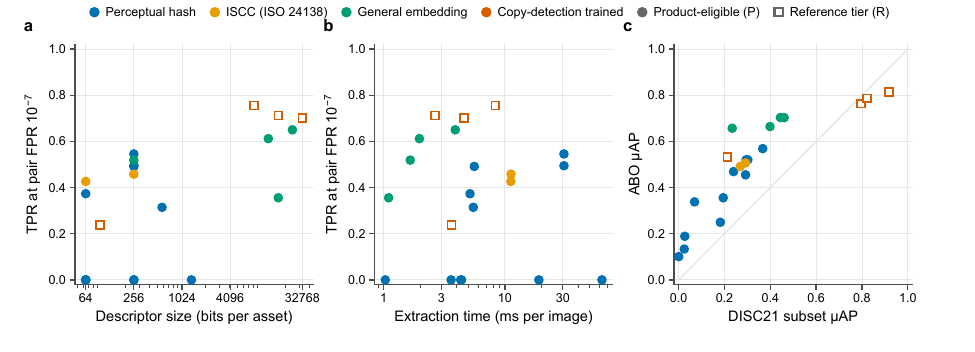}
\caption{(a) Detection at pair-level FPR $10^{-7}$ against stored size per asset. (b) Against extraction time per 1024-pixel image (learned methods on one A10G GPU in batches of 64, hashes on one CPU thread; measured on an idle host). (c) $\mu$AP on ABO against $\mu$AP on the DISC21 subset.}
\label{fig:fp-image-tradeoffs}
\end{figure*}

\paragraph{Size, speed and generalisation.} Fig.~\ref{fig:fp-image-tradeoffs} relates detection at the headline operating point to storage and extraction time. The keyed 256-bit projection of DINOv2-S keeps \val{fp:image/DINOv2-S-LSH256/TPR@pair@1e-07/pct}\% detection in 256 bits, against \val{fp:image/DINOv2-S/TPR@pair@1e-07/pct}\% for the 12\,288-bit float descriptor it is derived from. At PDQ's size it detects slightly fewer queries than dihedral PDQ at the strict threshold (\val{fp:image/PDQ-dihedral/TPR@pair@1e-07/pct}\%) but ranks the original first far more often (Recall@1 \val{fp:image/DINOv2-S-LSH256/R@1/pct}\% against \val{fp:image/PDQ-dihedral/R@1/pct}\%), because it survives crops and rotations. Panel~c shows that the ranking on ABO agrees with the ranking on our DISC21 subset, but that DISC21 separates the DISC21-trained descriptors far more sharply: SSCD and ISC21 reach $\mu$AP \val{fp:image/SSCD-mixup/disc} and \val{fp:image/ISC21-1st/disc} there, against \val{fp:image/DINOv2-S/disc} for DINOv2-S. DISC21's queries use the same augmentation library these models were trained with, and its references are photographs rather than product shots, so literature numbers on DISC21 overstate the gap on other content.

\begin{table}[!t]
\centering
\footnotesize
\caption{Planned paired comparisons on the same image queries at pair-level FPR $10^{-7}$. The unit is the registered source with all its transformed queries: $p_{\mathrm{Holm}}$ from a sign-flip test on per-source differences in detections (Holm-adjusted over the eight comparisons); the interval of the TPR difference is a bootstrap over sources that also redraws the calibration negatives.}
\label{tab:fp-paired}
\setlength{\tabcolsep}{3pt}
\begin{tabular}{llcccrr}
\toprule
A & B & TPR$_A$ & TPR$_B$ & A$-$B [95\% CI] & sources & $p_{\mathrm{Holm}}$ \\
\midrule
SSCD R50 & DINOv2-S & 0.712 & 0.612 & 0.100 \tiny[0.084, 0.115] & 2{,}000 & $<10^{-3}$ \\
DINOv2-S & DINOv2-S LSH-256 & 0.612 & 0.519 & 0.093 \tiny[0.066, 0.106] & 2{,}000 & $<10^{-3}$ \\
PDQ & ISCC Image-64 & 0.495 & 0.427 & 0.068 \tiny[0.058, 0.175] & 2{,}000 & $<10^{-3}$ \\
PDQ & PDQ (dihedral) & 0.495 & 0.546 & -0.051 \tiny[-0.055, -0.050] & 2{,}000 & $<10^{-3}$ \\
ISCC Image-64 & ISCC Image-256 & 0.427 & 0.458 & -0.032 \tiny[-0.136, -0.020] & 2{,}000 & $<10^{-3}$ \\
SSCD R50 & ISC21 1st & 0.712 & 0.755 & -0.043 \tiny[-0.056, -0.029] & 2{,}000 & $<10^{-3}$ \\
DINOv2-S & OpenCLIP B/32 & 0.612 & 0.356 & 0.256 \tiny[0.220, 0.286] & 2{,}000 & $<10^{-3}$ \\
pHash & PDQ & 0.373 & 0.495 & -0.121 \tiny[-0.497, -0.112] & 2{,}000 & $<10^{-3}$ \\
\bottomrule
\end{tabular}

\end{table}

\subsection{False Matches, Variants and Registry Scale}
\label{sec:fp-res-fpr}
\begin{figure*}[!t]
\centering
\includegraphics[width=\textwidth]{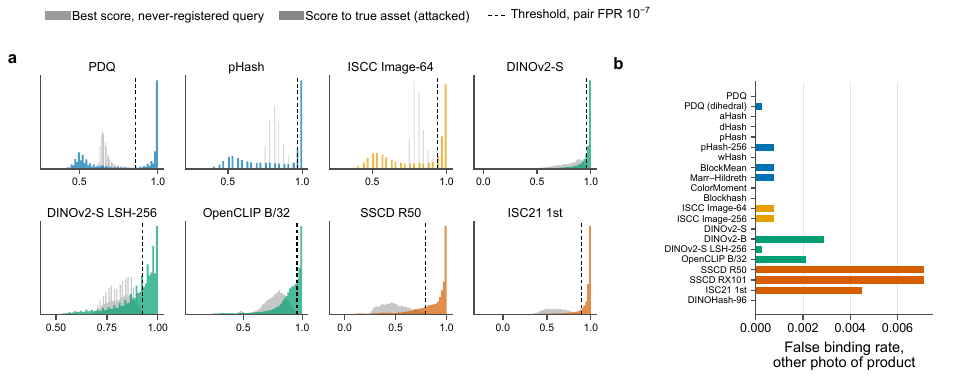}
\caption{(a) Score distributions: best score of a never-registered query (grey) and score of an attacked query to its true asset (colour), pooled over all transformations; dashed line: threshold at pair-level FPR $10^{-7}$. (b) Rate at which another photograph of a registered product binds to that product at the 1\% threshold.}
\label{fig:fp-scores}
\end{figure*}

Thresholds fixed on the calibration negatives held out of sample: at the query-level 1\% threshold the held-out never-registered queries matched at rates between \val{fp:image/fprq/min}\% and \val{fp:image/fprq/max}\% across methods (Table~\ref{tab:fp-ci}). Fig.~\ref{fig:fp-scores}a shows why the learned descriptors can be operated at a strict threshold: their score to the true asset is well separated from the best score of a never-registered query, whereas for the hashes the two distributions meet near the threshold. Short codes also run out of resolution: at pair-level FPR $10^{-8}$ the 64-bit pHash and ISCC codes would need a Hamming distance below zero, so no threshold exists and their detection is \val{fp:image/pHash-64/TPR@pair@1e-08} and \val{fp:image/ISCC-Image-64/TPR@pair@1e-08}.

\paragraph{Different photographs of the same product.} A provenance registry must not bind a second photograph of a product to the manifest of the first. On \val{fp:dedup/n_hard} hard negatives the hashes almost never did, and the learned descriptors rarely did at the 1\% threshold: the highest rate was SSCD's \val{fp:image/SSCD-mixup/hard/pct}\% (Fig.~\ref{fig:fp-scores}b). Recoloured or re-mounted versions of the \emph{same} photograph are another matter. The \val{fp:dedup/n_hard_near_duplicate} hard negatives that our duplicate rule marked as variants bound to the registered image in most cases for the learned descriptors (e.g. 139 of 216 for SSCD-R50 and 96 for DINOv2-S) and rarely for the hashes (31 for PDQ, 3 for pHash).

\paragraph{Catalogue variants dominate the false-match budget.} ABO, like any real catalogue, reuses a design across products: the same artwork on phone cases of different models, the same ring with another stone. Table~\ref{tab:fp-sens} repeats the image evaluation with two other treatments of such variants. R@1 barely moves, but the calibrated operating points move a great deal: with no cleanup detection at the 1\% threshold falls to between \val{fp:sens/none/DINOv2-S-LSH256/TPR@query@0.01} and \val{fp:sens/none/SSCD-mixup/TPR@query@0.01} for every method (and to zero for pHash-256, which can then no longer set a threshold), because the threshold must then exclude near-copies of registered designs that are, in this benchmark, labelled as different assets. The choice of method matters less than whether the registry groups variants of one design under one asset.

\begin{table}[!t]
\centering
\footnotesize
\caption{Sensitivity of the image operating points to the treatment of catalogue duplicates. \emph{Any}: the rule fixed in advance (a pair counts as a duplicate if PDQ, SSCD or DINOv2-B says so); \emph{two}: two of the three must agree; \emph{none}: no cleanup, so recoloured or re-mounted versions of a registered design count as false matches. R@1 changes by at most a few points; the calibrated TPR does not.}
\label{tab:fp-sens}
\setlength{\tabcolsep}{3pt}
\begin{tabular}{lcccccc}
\toprule
 & \multicolumn{2}{c}{Any (primary)} & \multicolumn{2}{c}{Two of three} & \multicolumn{2}{c}{None} \\
Method & TPR$_{q,1\%}$ & TPR$_{p,10^{-7}}$ & TPR$_{q,1\%}$ & TPR$_{p,10^{-7}}$ & TPR$_{q,1\%}$ & TPR$_{p,10^{-7}}$ \\
\midrule
PDQ & 0.495 & 0.495 & 0.476 & 0.471 & 0.214 & 0.214 \\
PDQ (dihedral) & 0.549 & 0.546 & 0.523 & 0.523 & 0.223 & 0.223 \\
pHash-256 & 0.494 & 0.492 & 0.476 & 0.472 & 0.000 & 0.000 \\
ISCC Image-256 & 0.466 & 0.458 & 0.448 & 0.448 & 0.211 & 0.000 \\
DINOv2-S & 0.625 & 0.612 & 0.546 & 0.544 & 0.194 & 0.192 \\
DINOv2-S LSH-256 & 0.538 & 0.519 & 0.462 & 0.462 & 0.174 & 0.174 \\
DINOv2-B & 0.653 & 0.650 & 0.542 & 0.537 & 0.201 & 0.199 \\
OpenCLIP B/32 & 0.401 & 0.356 & 0.365 & 0.344 & 0.185 & 0.170 \\
SSCD R50\reftier & 0.725 & 0.712 & 0.601 & 0.512 & 0.236 & 0.217 \\
ISC21 1st\reftier & 0.762 & 0.755 & 0.704 & 0.665 & 0.213 & 0.212 \\
\bottomrule
\end{tabular}

\end{table}

\begin{figure*}[!t]
\centering
\includegraphics[width=\textwidth]{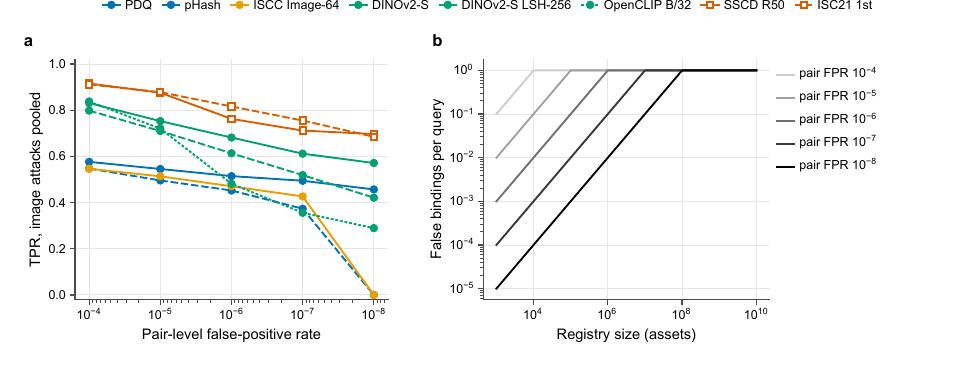}
\caption{(a) Detection of attacked image queries against the pair-level false-positive rate at which the threshold is set. (b) Expected number of false matches per query as a function of registry size for fixed pair-level rates.}
\label{fig:fp-det}
\end{figure*}

\paragraph{Registry scale.} For $N$ registered assets and average pair-level false-match probability $p$, the expected number of false matches per negative query is $Np$ (Fig.~\ref{fig:fp-det}b). This count differs from the probability that a query has at least one false match, which is bounded above by $\min(1,Np)$. Detection falls as the target tightens (Fig.~\ref{fig:fp-det}a): from $10^{-7}$ to $10^{-8}$ SSCD-R50 loses \val{fp:image/SSCD-mixup/TPR@pair@1e-07/pct}\,$\to$\,\val{fp:image/SSCD-mixup/TPR@pair@1e-08/pct}\% and DINOv2-S \val{fp:image/DINOv2-S/TPR@pair@1e-07/pct}\,$\to$\,\val{fp:image/DINOv2-S/TPR@pair@1e-08/pct}\%. For a registry of ten million assets, an actual average per-reference false-match probability at most $10^{-9}$ for the relevant query population would bound the query-level false-match probability by 1\%. That operating point is below the empirically resolved range of this benchmark.

\paragraph{Does geometric verification close the gap?} A natural remedy is a second stage that checks the retrieved candidate against the stored original. We tested it on every positive and negative image query: the top-1 candidate of DINOv2-S or PDQ is accepted only if SIFT with RANSAC finds at least $k$ geometric inliers between the query and the registered image, with both the retrieval threshold and $k$ fixed on the calibration negatives (Table~\ref{tab:fp-verify}). For the evaluated DINOv2-S and PDQ pipelines, no detection improvement from verification was observed under a matched calibration false-binding constraint. With $k$ chosen so that the calibration false bindings do not exceed those of DINOv2-S alone at pair-level $10^{-7}$, the two-stage rule detected \val{fp:verify/DINOv2-S/query@0.01/k_match/tpr} of attacked queries against \val{fp:verify/DINOv2-S/one/tpr} for retrieval alone, and starting from the looser pair-level $10^{-4}$ threshold it detected only \val{fp:verify/DINOv2-S/pair@0.0001/k_match/tpr}. With the localization track's fixed $k=12$ it removed about four in ten held-out false bindings (\val{fp:verify/DINOv2-S/one/fp}$\,\to\,$\val{fp:verify/DINOv2-S/query@0.01/k12/fp} of \val{fp:verify/n_test}) at a cost of under a point of detection (\val{fp:verify/DINOv2-S/query@0.01/k12/tpr}), and PDQ behaved alike (\val{fp:verify/PDQ/one/fp}$\,\to\,$\val{fp:verify/PDQ/query@0.01/k12/fp}). Eliminating the remaining false bindings on the calibration set required hundreds of inliers (\val{fp:verify/DINOv2-S/query@0.01/k_zero/k} for DINOv2-S) and left \val{fp:verify/DINOv2-S/query@0.01/k_zero/tpr} detection. The false bindings that survive retrieval are, in large part, near-copies of registered designs, the catalogue variants of the sensitivity analysis, and they agree with the registered image geometrically as well as globally: a local-feature check confirms that two images share content, not that they are the same asset.

\begin{table}[!t]
\centering
\footnotesize
\caption{Second-stage verification of image matches. The top-1 candidate is accepted only if its retrieval score passes a threshold and SIFT+RANSAC finds at least $k$ geometric inliers between the query and the stored original. Both thresholds are fixed on the calibration negatives: matched, the smallest $k$ whose calibration false bindings do not exceed those of retrieval alone at pair-level $10^{-7}$; zero cal., the smallest $k$ no calibration negative reaches; and the localization track's $k=12$. FB: false bindings among the 26{,}820 held-out never-registered queries. TPR interval: bootstrap over sources.}
\label{tab:fp-verify}
\setlength{\tabcolsep}{3pt}
\begin{tabular}{llrccc}
\toprule
Method & Retrieval threshold & $k$ & TPR & FB & FB rate \\
\midrule
DINOv2-S & pair $10^{-7}$ & -- & 0.612 & 180 & 0.0067 \\
 & query 1\% & 9 (matched) & 0.609 \tiny[0.60, 0.61] & 124 & 0.0046 \\
 & query 1\% & 12 ($k=12$) & 0.604 \tiny[0.60, 0.61] & 104 & 0.0039 \\
 & query 1\% & 278 (zero cal.) & 0.206 \tiny[0.20, 0.22] & 0 & 0.0000 \\
 & pair $10^{-4}$ & 266 (matched) & 0.268 \tiny[0.25, 0.28] & 146 & 0.0054 \\
 & pair $10^{-4}$ & 12 ($k=12$) & 0.793 \tiny[0.79, 0.80] & 4{,}390 & 0.1637 \\
 & pair $10^{-4}$ & 1119 (zero cal.) & 0.023 \tiny[0.02, 0.03] & 0 & 0.0000 \\
\addlinespace
PDQ & pair $10^{-7}$ & -- & 0.495 & 267 & 0.0100 \\
 & query 1\% & 0 (matched) & 0.495 \tiny[0.49, 0.50] & 267 & 0.0100 \\
 & query 1\% & 12 ($k=12$) & 0.481 \tiny[0.48, 0.48] & 134 & 0.0050 \\
 & query 1\% & 644 (zero cal.) & 0.065 \tiny[0.06, 0.07] & 0 & 0.0000 \\
 & pair $10^{-4}$ & 167 (matched) & 0.299 \tiny[0.29, 0.31] & 212 & 0.0079 \\
 & pair $10^{-4}$ & 12 ($k=12$) & 0.559 \tiny[0.56, 0.56] & 2{,}565 & 0.0956 \\
 & pair $10^{-4}$ & 684 (zero cal.) & 0.066 \tiny[0.06, 0.07] & 0 & 0.0000 \\
\bottomrule
\end{tabular}

\end{table}

\subsection{Audio}
\label{sec:fp-res-audio}
\begin{figure*}[!t]
\centering
\includegraphics[width=\textwidth]{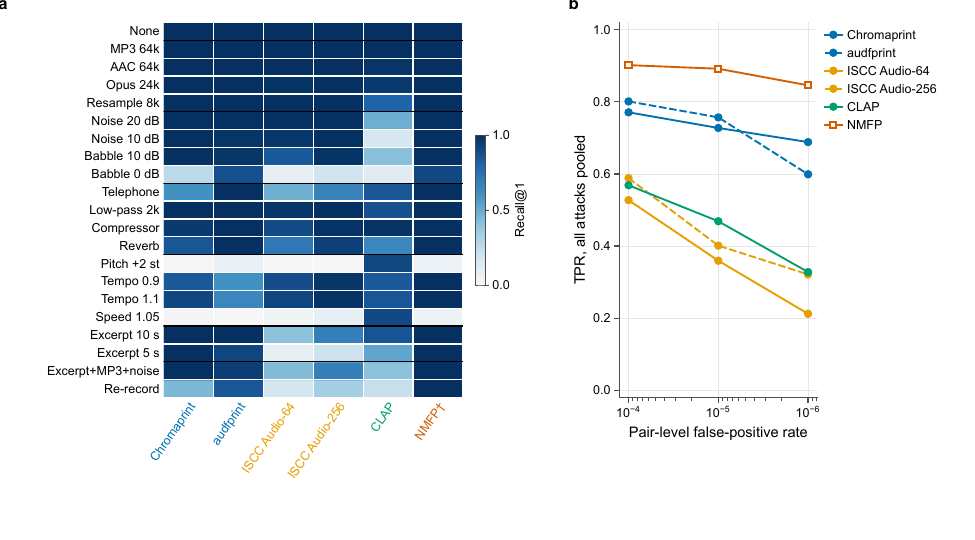}
\caption{(a) Recall@1 of each audio method under each transformation (\val{fp:corpus/audio/reg} registered items, \val{fp:count/audio/nref} in the index). (b) Detection against the pair-level false-positive rate; unresolved targets are omitted.}
\label{fig:fp-audio}
\end{figure*}

Fig.~\ref{fig:fp-audio} and Table~\ref{tab:fp-audio} report the audio track. As in the image track, the corpus contained duplicates: FMA carries some recordings under two track identifiers, which all three recording-level systems matched perfectly. We removed the \val{fp:audio/dedup/neg} negative sources that were copies of indexed tracks and counted a duplicate of a registered track as the same asset (\val{fp:audio/dedup/items} registered tracks had one), using the same any-of-three rule as for images. Without this step every method's calibrated detection falls by half or more (e.g. Chromaprint \val{fp:audio/Chromaprint/nodedup/tpr} instead of \val{fp:audio/Chromaprint/TPR@query@0.01}); we report the corrected values. With \val{fp:count/audio/neg_cal} calibration queries the pair-level targets below $10^{-6}$ are not resolved, so we compare at the query-level 1\% threshold.

The neural music fingerprint NMFP detects the most transformed queries (\val{fp:audio/NMFP-triplet/TPR@query@0.01/pct}\%), followed by Chromaprint (\val{fp:audio/Chromaprint/TPR@query@0.01/pct}\%) and audfprint (\val{fp:audio/audfprint/TPR@query@0.01/pct}\%); the ISCC Audio-Code reaches \val{fp:audio/ISCC-Audio-256/TPR@query@0.01/pct}\% at 256 bits and \val{fp:audio/ISCC-Audio-64/TPR@query@0.01/pct}\% at its default 64 bits, and the semantic CLAP embedding \val{fp:audio/CLAP/TPR@query@0.01/pct}\%. The families fail on different transformations (Fig.~\ref{fig:fp-audio}a). Codecs, resampling and moderate noise are harmless to all of them. Excerpts separate the systems that align in time from those that summarise a whole file: Chromaprint, audfprint and NMFP find a 5\,s excerpt of a 30\,s track in \val{fp:audio/Chromaprint/att/excerpt5/pct}\%, \val{fp:audio/audfprint/att/excerpt5/pct}\% and \val{fp:audio/NMFP-triplet/att/excerpt5/pct}\% of cases, while the ISCC Audio-Code, a single SimHash over the whole Chromaprint vector, finds \val{fp:audio/ISCC-Audio-64/att/excerpt5/pct}\%. Loud babble from other music (0\,dB) defeats the whole-file codes and CLAP, and most Chromaprint matches (\val{fp:audio/Chromaprint/att/babble_0db/pct}\%), while the landmark system audfprint (\val{fp:audio/audfprint/att/babble_0db/pct}\%), whose sparse spectral peaks survive mixing, and NMFP (\val{fp:audio/NMFP-triplet/att/babble_0db/pct}\%) keep most of them. Babble results depend on which background tracks are mixed in: the two runs of this benchmark drew different backgrounds (Section~\ref{sec:fp-method}), and NMFP's rate at 0\,dB was about half as high with the first set. Pitch and speed changes of a few percent defeat every recording-level fingerprint (Chromaprint \val{fp:audio/Chromaprint/att/pitch+2/pct}\% at $+2$ semitones, NMFP \val{fp:audio/NMFP-triplet/att/pitch+2/pct}\%), and only CLAP, which describes content rather than the recording, survives them (\val{fp:audio/CLAP/att/pitch+2/pct}\%) at the cost of failing under noise (\val{fp:audio/CLAP/att/noise_10db/pct}\% at 10\,dB SNR). The simulated re-recording is handled well by NMFP (\val{fp:audio/NMFP-triplet/att/rerecord/pct}\%) and audfprint (\val{fp:audio/audfprint/att/rerecord/pct}\%) and poorly by the others.

Speech behaves like music for the recording-level systems (Recall@1 on 10\,s excerpts: Chromaprint \val{fp:audio/Chromaprint/speech/att/excerpt10/pct}\% on speech and \val{fp:audio/Chromaprint/music/att/excerpt10/pct}\% on music), and ISCC does better on speech than on music excerpts (\val{fp:audio/ISCC-Audio-256/speech/att/excerpt10/pct}\% against \val{fp:audio/ISCC-Audio-256/music/att/excerpt10/pct}\% at 256 bits). NMFP is the only reference-tier audio method; it is GPL-3.0 and was trained on FMA, whose training tracks we could not exclude from our FMA-large corpus by identifier, so its lead may be partly in-distribution.

\begin{table*}[!t]
\centering
\scriptsize
\caption{Audio retrieval over all attacks, and Recall@1 under selected attacks. TPR$_{q,1\%}$: detection at the threshold set for a 1\% query-level false-match rate on the calibration negatives; FPR$_q$ (test): the rate that threshold actually gave on the held-out negatives. TPR$_{p,10^{-6}}$: the strictest pair-level target that 4053 calibration queries against 6350 references resolve; stricter targets are not reported. Intervals: 95\% bootstrap over sources.}
\label{tab:fp-audio}
\setlength{\tabcolsep}{3pt}
\begin{tabular}{lccccccccc}
\toprule
Method & R@1 & $\mu$AP & TPR$_{q,1\%}$ & FPR$_{q}$ (test) & TPR$_{p,10^{-6}}$ & Music 10\,s exc. & Speech 10\,s exc. & Pitch +2 & Re-record \\
\midrule
Chromaprint & 0.805 & 0.817 & 0.704 \tiny[0.68, 0.72] & 0.016 \tiny[0.00, 0.04] & 0.688 \tiny[0.66, 0.71] & 1.000 & 1.000 & 0.006 & 0.451 \\
audfprint & 0.851 & 0.851 & 0.660 \tiny[0.53, 0.74] & 0.008 \tiny[0.00, 0.03] & 0.599 \tiny[0.46, 0.72] & 0.982 & 1.000 & 0.071 & 0.852 \\
ISCC Audio-64 & 0.662 & 0.581 & 0.409 \tiny[0.21, 0.41] & 0.017 \tiny[0.00, 0.03] & 0.212 \tiny[0.00, 0.41] & 0.247 & 0.657 & 0.004 & 0.183 \\
ISCC Audio-256 & 0.749 & 0.658 & 0.454 \tiny[0.31, 0.51] & 0.019 \tiny[0.00, 0.04] & 0.321 \tiny[0.24, 0.47] & 0.555 & 0.882 & 0.006 & 0.342 \\
CLAP & 0.707 & 0.446 & 0.386 \tiny[0.33, 0.45] & 0.017 \tiny[0.01, 0.04] & 0.328 \tiny[0.27, 0.41] & 0.957 & 0.725 & 0.910 & 0.228 \\
NMFP\reftier & 0.905 & 0.937 & 0.862 \tiny[0.82, 0.88] & 0.018 \tiny[0.00, 0.04] & 0.846 \tiny[0.79, 0.88] & 1.000 & 1.000 & 0.057 & 1.000 \\
\bottomrule
\end{tabular}

\end{table*}

\subsection{Video}
\label{sec:fp-res-video}
\begin{figure*}[!t]
\centering
\includegraphics[width=\textwidth]{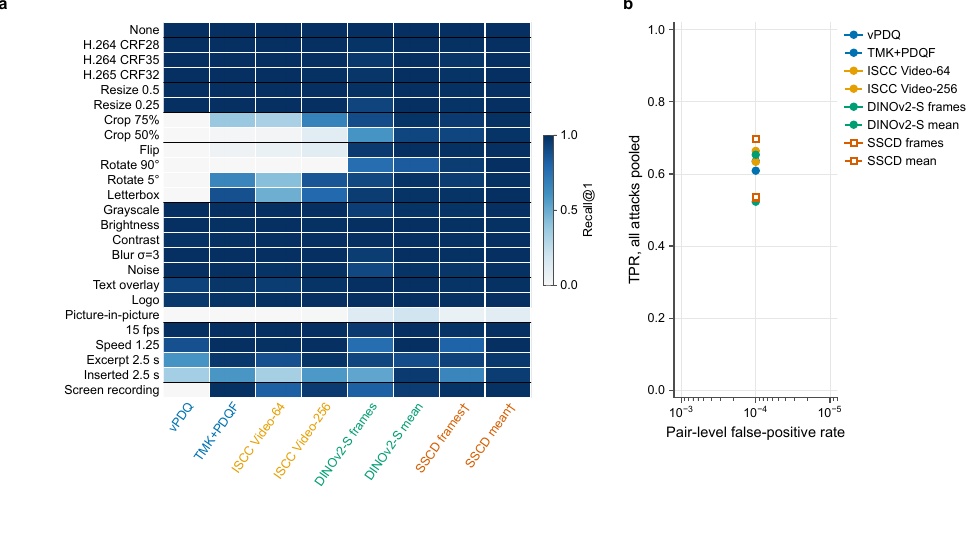}
\caption{(a) Recall@1 of each video method under each transformation (\val{fp:corpus/video/reg} registered clips, \val{fp:count/video/nref} in the index). (b) Detection against the pair-level false-positive rate; unresolved targets are omitted.}
\label{fig:fp-video}
\end{figure*}

Fig.~\ref{fig:fp-video} and Table~\ref{tab:fp-video} report the video track. The \val{fp:count/video/neg_cal} calibration queries come from \val{fp:video/cal_sources} negative clips against \val{fp:count/video/nref} references, which resolves pair-level targets down to $10^{-4}$ but not below; Table~\ref{tab:fp-video} therefore reports the query-level 1\% threshold and the pair-level $10^{-4}$ target, and no stricter one. The query-level threshold did not deliver 1\% out of sample: on the held-out negatives it let through between \val{fp:video/fprq/min}\% and \val{fp:video/fprq/max}\% of never-registered queries, depending on the method. The threshold sits just above the sixth-highest of 500 calibration scores, so that five pass, and those queries are \val{fp:video/q_per_source} versions (the original and its transformations) of each of \val{fp:video/cal_sources} clips from a few films, whose clips resemble each other. The effective sample is the number of source clips, so the 1\% point is estimated imprecisely: the source-level 95\% intervals of the held-out rate reach up to \val{fp:video/fprq/cihi}\%. We report every video detection rate next to the false-match rate its threshold actually produced, and treat the video operating points as indicative. Frame-level learned descriptors are the most robust: SSCD and DINOv2-S reach Recall@1 of \val{fp:video/SSCD-mixup-mean/R@1/pct}\% and \val{fp:video/DINOv2-S-mean/R@1/pct}\% when their per-frame descriptors are averaged over the clip, and they are the only methods that survive 50\% crops, rotations and flips. Averaging, however, is also why their calibrated detection is lower than their Recall@1 suggests (\val{fp:video/SSCD-mixup-mean/TPR@query@0.01/pct}\% and \val{fp:video/DINOv2-S-mean/TPR@query@0.01/pct}\%): clips from the same film average to similar vectors, which raises the threshold. The ISCC Video-Code and TMK+PDQF are less robust but detect more at the 1\% threshold (\val{fp:video/ISCC-Video-256/TPR@query@0.01/pct}\% and \val{fp:video/TMK+PDQF/TPR@query@0.01/pct}\%), at the price of more held-out false matches (\val{fp:video/ISCC-Video-256/FPRq@query@0.01/pct}\% and \val{fp:video/TMK+PDQF/FPRq@query@0.01/pct}\%, against \val{fp:video/SSCD-mixup-mean/FPRq@query@0.01/pct}\% and \val{fp:video/DINOv2-S-mean/FPRq@query@0.01/pct}\% for the averaged learned descriptors); with 20 calibration clips these differences in false-match rate are within the intervals of Table~\ref{tab:fp-video}. Picture-in-picture defeats every method, as the overlay attacks did for images.

\begin{table*}[!t]
\centering
\scriptsize
\caption{Video retrieval over all attacks, and Recall@1 under selected attacks. TPR$_{q,1\%}$: detection at the threshold set for a 1\% query-level false-match rate on the calibration negatives; FPR$_q$ (test): the rate that threshold actually gave on the held-out negatives. TPR$_{p,10^{-4}}$: the strictest pair-level target that 500 calibration queries against 443 references resolve; stricter targets are not reported. Intervals: 95\% bootstrap over sources.}
\label{tab:fp-video}
\setlength{\tabcolsep}{3pt}
\begin{tabular}{lccccccccc}
\toprule
Method & R@1 & $\mu$AP & TPR$_{q,1\%}$ & FPR$_{q}$ (test) & TPR$_{p,10^{-4}}$ & Crop 50\% & PiP & Inserted & Screen rec. \\
\midrule
vPDQ & 0.628 & 0.617 & 0.577 \tiny[0.56, 0.59] & 0.032 \tiny[0.00, 0.08] & 0.609 \tiny[0.57, 0.62] & 0.000 & 0.000 & 0.342 & 0.000 \\
TMK+PDQF & 0.778 & 0.723 & 0.631 \tiny[0.61, 0.66] & 0.066 \tiny[0.00, 0.16] & 0.635 \tiny[0.63, 0.70] & 0.017 & 0.000 & 0.583 & 0.992 \\
ISCC Video-64 & 0.731 & 0.700 & 0.620 \tiny[0.61, 0.65] & 0.062 \tiny[0.00, 0.16] & 0.634 \tiny[0.62, 0.66] & 0.025 & 0.000 & 0.333 & 0.817 \\
ISCC Video-256 & 0.801 & 0.749 & 0.652 \tiny[0.63, 0.67] & 0.068 \tiny[0.00, 0.17] & 0.664 \tiny[0.66, 0.69] & 0.108 & 0.008 & 0.575 & 0.958 \\
DINOv2-S frames & 0.872 & 0.691 & 0.437 \tiny[0.41, 0.58] & 0.062 \tiny[0.01, 0.15] & 0.523 \tiny[0.45, 0.63] & 0.592 & 0.125 & 0.525 & 0.817 \\
DINOv2-S mean & 0.947 & 0.831 & 0.547 \tiny[0.51, 0.73] & 0.032 \tiny[0.00, 0.16] & 0.653 \tiny[0.56, 0.77] & 0.917 & 0.200 & 0.958 & 0.950 \\
SSCD frames\reftier & 0.921 & 0.806 & 0.502 \tiny[0.48, 0.56] & 0.032 \tiny[0.00, 0.10] & 0.535 \tiny[0.50, 0.69] & 0.917 & 0.067 & 0.658 & 0.983 \\
SSCD mean\reftier & 0.958 & 0.893 & 0.543 \tiny[0.52, 0.78] & 0.036 \tiny[0.00, 0.14] & 0.697 \tiny[0.55, 0.83] & 0.983 & 0.108 & 0.950 & 0.992 \\
\bottomrule
\end{tabular}

\end{table*}

\subsection{Partial Edits and Localization}
\label{sec:fp-res-edits}
\begin{figure*}[!t]
\centering
\includegraphics[width=\textwidth]{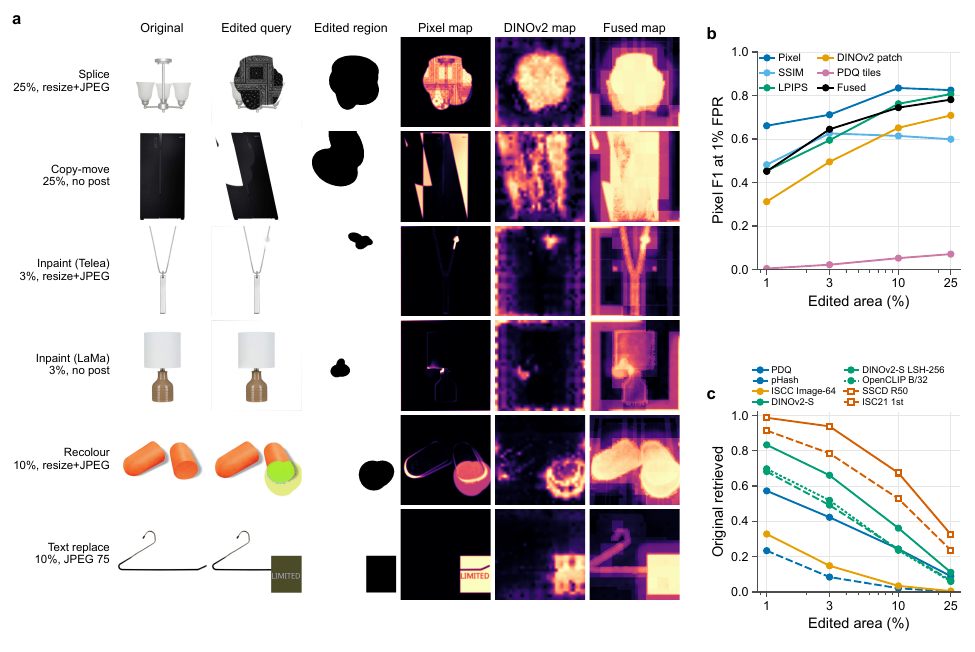}
\caption{(a) Examples of each edit type: registered original, edited query, ground-truth region, and three difference maps after SIFT alignment. (b) Pixel F1 at the calibrated threshold against the edited area, per map. (c) Share of edited queries whose original is retrieved above the query-level 1\% threshold, against edited area, for the eight representative methods of Fig.~\ref{fig:fp-image-families}.}
\label{fig:fp-loc}
\end{figure*}

Partial edits cost the global descriptors little in ranking but much in calibrated detection. SSCD-R50 still ranks the original first for \val{fp:editret/SSCD-mixup/R@1}\ of the \val{fp:edit/n_edits} edited images and DINOv2-S for \val{fp:editret/DINOv2-S/R@1}, while PDQ does for \val{fp:editret/PDQ/R@1} and the ISCC Image-Code for \val{fp:editret/ISCC-Image-64/R@1}. At the 1\% threshold, however, an edit covering a quarter of the image leaves SSCD's score above threshold in only \val{fp:editret/SSCD-mixup/query@0.01/area/0.25} of cases, against \val{fp:editret/SSCD-mixup/query@0.01/area/0.01} for 1\% edits (Fig.~\ref{fig:fp-loc}c). A fingerprint that tolerates an edit well enough to find the original is, by design, not a tamper detector.

Once the original is found, the edit can be localized against it (Table~\ref{tab:fp-loc}, Fig.~\ref{fig:fp-loc}). Table~\ref{tab:fp-loc} is an oracle measurement: each edited query is compared with its true registered original, whether or not a fingerprint would have retrieved it. SIFT alignment succeeded for \val{fp:loc/sift/aligned}\% of the evaluated edits and DISK+LightGlue for \val{fp:loc/lightglue/aligned}\%. The simplest map, a blurred pixel difference after alignment, gave the best F1 at the calibrated 1\% pixel false-positive rate (\val{fp:loc/sift/pixel/f1} with SIFT), and LPIPS the highest AUC (\val{fp:loc/sift/lpips/auc}); the fused map was the most robust to the 90\% crop, which LightGlue aligns better than SIFT (fused F1 \val{fp:loc/lightglue/fused/post/crop90/f1} against \val{fp:loc/sift/fused/post/crop90/f1}). Text replacement and recolouring were the easiest edits (pixel F1 \val{fp:loc/sift/pixel/type/textreplace/f1} and \val{fp:loc/sift/pixel/type/recolor/f1}) and learned inpainting the hardest (\val{fp:loc/sift/pixel/type/inpaint-lama/f1}), because a good inpainting of a textureless background changes few pixels. PDQ computed per tile could not localize at all at a 1\% false-positive rate (F1 \val{fp:loc/sift/pdq/f1}): JPEG re-encoding alone flips tile hashes. End to end, the original must first be retrieved. Counting an edit whose original a fingerprint does not rank first as a localization failure, the SIFT pixel-difference F1 over the \val{fp:loc/e2e/sift/n} evaluated edits falls from \val{fp:loc/sift/pixel/f1} (oracle) to \val{fp:loc/e2e/sift/SSCD-mixup/pixel/f1} with SSCD-R50 retrieval, \val{fp:loc/e2e/sift/DINOv2-S/pixel/f1} with DINOv2-S, \val{fp:loc/e2e/sift/PDQ/pixel/f1} with PDQ and \val{fp:loc/e2e/sift/ISCC-Image-64/pixel/f1} with the ISCC Image-Code, which ranked the original first for \val{fp:loc/e2e/sift/SSCD-mixup/retrieved}, \val{fp:loc/e2e/sift/DINOv2-S/retrieved}, \val{fp:loc/e2e/sift/PDQ/retrieved} and \val{fp:loc/e2e/sift/ISCC-Image-64/retrieved}\% of them. These figures use top-1 retrieval without a threshold; requiring the calibrated threshold as well lowers them further, by the gap between Recall@1 and TPR for edited queries reported above.

\begin{table}[!t]
\centering
\footnotesize
\caption{Localization of partial edits against the registered original (oracle retrieval; evaluation half of the sources). F1 at the threshold fixed for a 1\% pixel false-positive rate on unedited, post-processed originals; best F1 chooses the threshold per map with hindsight.}
\label{tab:fp-loc}
\setlength{\tabcolsep}{3pt}
\begin{tabular}{lcccccc}
\toprule
 & \multicolumn{3}{c}{SIFT + RANSAC} & \multicolumn{3}{c}{DISK + LightGlue} \\
Map & AUC & F1 & best F1 & AUC & F1 & best F1 \\
\midrule
Pixel difference & 0.944 & 0.75 & 0.83 & 0.942 & 0.72 & 0.82 \\
SSIM & 0.927 & 0.57 & 0.74 & 0.924 & 0.59 & 0.71 \\
LPIPS & 0.975 & 0.64 & 0.79 & 0.973 & 0.61 & 0.78 \\
DINOv2 patches & 0.958 & 0.53 & 0.70 & 0.952 & 0.52 & 0.69 \\
PDQ tiles & 0.866 & 0.04 & 0.50 & 0.866 & 0.02 & 0.49 \\
Fused & 0.967 & 0.65 & 0.81 & 0.965 & 0.69 & 0.79 \\
\bottomrule
\end{tabular}

\end{table}

For time-based media, localization means finding where a query sits in the registered item. Chromaprint recovered the offset of 10\,s and 5\,s excerpts and of re-recordings within 0.5\,s in \val{fp:tloc/audio/chromaprint/_all/within05}\% of cases (median error \val{fp:tloc/audio/chromaprint/_all/median_ms}\,ms) and audfprint in \val{fp:tloc/audio/audfprint/_all/within05}\% (median \val{fp:tloc/audio/audfprint/_all/median_ms}\,ms); its failures are mostly queries it did not match at all. For video, frame-level SSCD and DINOv2-S placed every 2.5\,s excerpt within 1\,s of its true position, and recovered the inserted interval of the insertion attack with a mean interval IoU of \val{fp:tloc/video/SSCD-mixup/insert/iou} and \val{fp:tloc/video/DINOv2-S/insert/iou}.

\subsection{Security}
\label{sec:fp-res-security}
\begin{figure*}[!t]
\centering
\includegraphics[width=\textwidth]{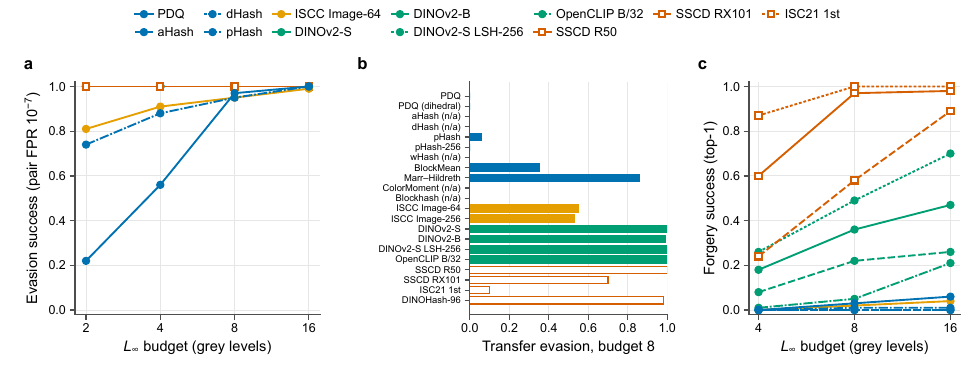}
\caption{Adversarial robustness on \val{fp:sec/n} registered ABO images per setting. (a) Evasion by white-box PGD (learned descriptors) and surrogate-gradient PGD (DCT hashes) against the $L_\infty$ budget, judged by the real extractor at pair-level FPR $10^{-7}$. (b) Evasion by perturbations crafted on a DINOv2-S, SSCD-R50 and OpenCLIP ensemble at a budget of 8 grey levels, judged at the pair-level $10^{-6}$ threshold the attack ran against; n/a: the method has no reachable operating point. (c) Targeted forgery: share of unrelated images that become their target's top match above the $10^{-6}$ threshold.}
\label{fig:fp-security}
\end{figure*}

\paragraph{Evasion.} Every learned descriptor could be evaded by a white-box attacker with an imperceptible perturbation (Fig.~\ref{fig:fp-security}a): at a budget of 2 grey levels (PSNR \val{fp:sec/whitebox/DINOv2-S/2/psnr}\,dB) projected gradient descent pushed \val{fp:sec/whitebox/DINOv2-S/2/pct7}\% of the registered images below the $10^{-7}$ threshold for DINOv2-S, and equally for SSCD, ISC21, OpenCLIP and the keyed projection, whose soft pre-sign values carry the gradient. The DCT hashes resisted longer but not for long: through differentiable surrogates the attack evaded PDQ for \val{fp:sec/surrogate/PDQ/2/pct7}\%, \val{fp:sec/surrogate/PDQ/4/pct7}\% and \val{fp:sec/surrogate/PDQ/8/pct7}\% of images at 2, 4 and 8 grey levels, pHash for \val{fp:sec/surrogate/pHash-64/2/pct7}\% already at 2, and the ISCC Image-Code for \val{fp:sec/surrogate/ISCC-Image-64/2/pct7}\% at 2. A black-box attacker who crafts the perturbation on an ensemble of DINOv2-S, SSCD and OpenCLIP (Fig.~\ref{fig:fp-security}b, budget 8) evades every DINOv2 variant and SSCD-R50 but transfers poorly to ISC21 (\val{fp:sec/transfer/8/ISC21-1st/pct}\%) and not at all to PDQ and pHash-256; the ISCC codes are evaded for about half of the images at every budget (\val{fp:sec/transfer/2/ISCC-Image-64/pct}\% at 2 grey levels), again through the border-trimming step, which small perturbations of a white background switch on or off.

\paragraph{Forgery.} Binding an unrelated image to a registered asset is harder but, for the learned copy detectors, practical (Fig.~\ref{fig:fp-security}c): at 4 grey levels the attack made \val{fp:sec/collision/ISC21-1st/4/pct}\% of source images the top match of their target in the full registry above threshold for ISC21 and \val{fp:sec/collision/SSCD-mixup/4/pct}\% for SSCD-R50, rising to \val{fp:sec/collision/ISC21-1st/8/pct}\% and \val{fp:sec/collision/SSCD-mixup/8/pct}\% at 8. DINOv2-S (\val{fp:sec/collision/DINOv2-S/8/pct}\% at 8) and OpenCLIP (\val{fp:sec/collision/OpenCLIP-B32/8/pct}\%) were harder to steer, and the hashes nearly impossible within these budgets (PDQ \val{fp:sec/collision/PDQ/16/pct}\% and ISCC \val{fp:sec/collision/ISCC-Image-64/16/pct}\% at 16 grey levels, where the PSNR is already \val{fp:sec/collision/PDQ/16/psnr}\,dB). The copy-detection descriptors, which generalise best across benign transformations, are thus also the easiest to forge: their similarity is smooth in the pixels, which is what both robustness and gradient steering exploit.

\begin{figure*}[!t]
\centering
\includegraphics[width=\textwidth]{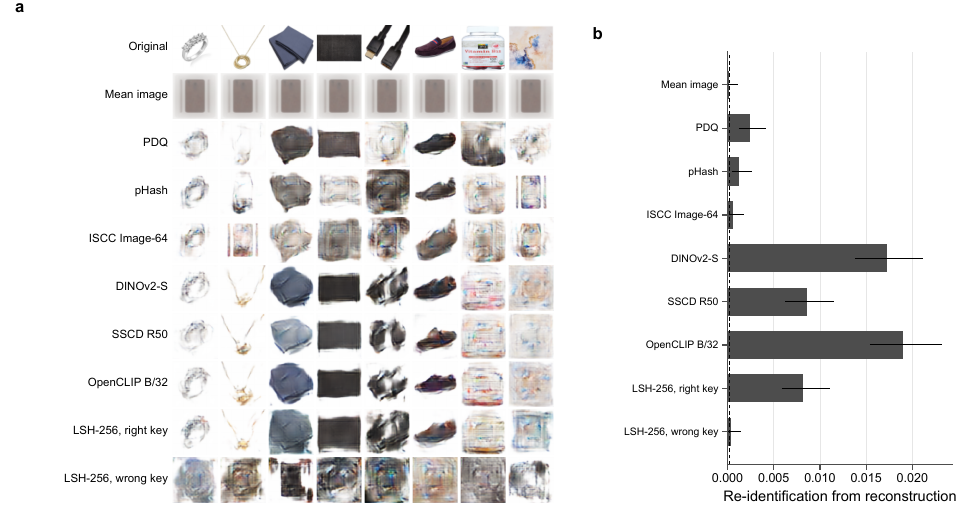}
\caption{Inverting stored fingerprints. (a) Registered images (top) and thumbnails reconstructed from their descriptors by a decoder trained on distractor descriptors; bottom rows: the keyed 256-bit projection decoded by an attacker with the right and with a wrong key. (b) Share of reconstructions that retrieve their own original among the \val{fp:sec/inv/ntest} registered images (95\% Clopper--Pearson intervals); dashed: chance.}
\label{fig:fp-inversion}
\end{figure*}

\paragraph{Inversion.} A soft-binding value is published inside the manifest, so we asked how much of the image it gives away. A decoder trained on the descriptors of the \val{fp:sec/inv/ntrain} distractors reconstructed a $64\times64$ thumbnail of each of \val{fp:sec/inv/ntest} registered images from its descriptor alone (Fig.~\ref{fig:fp-inversion}). Reconstructions from the float descriptors were recognisable in colour, layout and object class (LPIPS \val{fp:sec/inv/DINOv2-S/lpips} for DINOv2-S against \val{fp:sec/inv/mean-image/lpips} for the mean image), and using them as queries retrieved the true image among all \val{fp:sec/inv/ntest} in \val{fp:sec/inv/OpenCLIP-B32/reid/pct}\% (OpenCLIP), \val{fp:sec/inv/DINOv2-S/reid/pct}\% (DINOv2-S) and \val{fp:sec/inv/SSCD-mixup/reid/pct}\% (SSCD) of cases, 40 to 95 times chance. The binary hashes leaked less but still identified images above chance (PDQ \val{fp:sec/inv/PDQ/reid/pct}\%). The keyed projection behaved as intended: an attacker who knows the key still identifies images at about 40 times chance (\val{fp:sec/inv/DINOv2-S-LSH256/reid/pct}\%, half the rate from the float DINOv2-S descriptor), but a decoder trained with any other key reconstructs only the mean image (LPIPS \val{fp:sec/inv/DINOv2-S-LSH256-wrongkey/lpips}) and identifies images at chance (\val{fp:sec/inv/DINOv2-S-LSH256-wrongkey/reid/pct}\%).

\paragraph{Audio and video.} The same pattern holds in the other modalities. Audio attacks were judged, like the image attacks, at the pair-level $10^{-6}$ threshold of each method, taken from the first audio retrieval run; those thresholds differ from the re-run reported in Section~\ref{sec:fp-res-audio} by at most \val{fp:sec/audio/thdelta}, and re-scoring every CLAP attack at the re-run thresholds changes \val{fp:sec/audio/rescore_changed} outcomes. Video attacks were judged at a different threshold, described below. For audio, projected gradient descent on the CLAP embedding of \val{fp:sec/audio/n} registered tracks evaded CLAP for every track at every budget we tried, down to a target signal-to-noise ratio of 40\,dB (achieved median \val{fp:sec/audio/snr/40}\,dB). The perturbation did not transfer to the recording-level fingerprints: Chromaprint matched every attacked track at all budgets, and the ISCC Audio-Code and audfprint lost \val{fp:sec/audio/pgd/20/ISCC-Audio-64/pct}\% and \val{fp:sec/audio/pgd/20/audfprint/pct}\% only at the loudest budget (20\,dB target), where the added signal acts as noise rather than as a targeted attack (\val{fp:sec/audio/pgd/40/ISCC-Audio-64/pct}\% and \val{fp:sec/audio/pgd/40/audfprint/pct}\% at 40\,dB). An attacker does not need gradients against those systems, however: a search over pitch and tempo changes found the smallest change that broke each match, and the median was \val{fp:sec/audio/mag/Chromaprint/pitch} semitones or a tempo factor of \val{fp:sec/audio/mag/Chromaprint/tempo} for Chromaprint and \val{fp:sec/audio/mag/audfprint/pitch} semitones or \val{fp:sec/audio/mag/audfprint/tempo} for audfprint and the ISCC Audio-Code, where 0.25 semitones and 1.02 are the smallest steps of the grid. Only CLAP needed audible changes (\val{fp:sec/audio/mag/CLAP/pitch} semitones, tempo \val{fp:sec/audio/mag/CLAP/tempo}). Every audio fingerprint in this study is therefore evaded by a small change, either by gradients (CLAP) or by a pitch shift (all others).

For video, frame-wise projected gradient descent against DINOv2-S and SSCD on \val{fp:sec/video/n} registered clips evaded both frame-level neural descriptors, in sequence and mean-pooled form, for every clip at 4 grey levels, and transferred to none of the hashes: vPDQ, TMK+PDQF and both ISCC Video-Codes matched every attacked clip at 4 and 8 grey levels (judged at each method's query-level 1\% threshold). The video calibration set does not resolve a pair-level $10^{-6}$ target, so every video attack used that threshold, which let through \val{fp:video/fprq/min}\% to \val{fp:video/fprq/max}\% of held-out negative queries (Section~\ref{sec:fp-res-video}). The video results therefore do not establish attack performance at a pair-level false-match rate. Each hash's pair-level $10^{-4}$ threshold is lower than its query-level threshold, so the hashes' matches hold at that target too; the per-clip scores of the neural descriptors were not kept, so their evasion was not re-scored at it. Simple edits again did what gradients could not. A centred crop keeping 90\% of the width and height, the first step of the grid, broke the median match for vPDQ, TMK+PDQF, both ISCC Video-Codes and the sequence-matched DINOv2-S; only SSCD and mean-pooled DINOv2-S held at 90\% and broke at a median of \val{fp:sec/video/magnitude/SSCD-mixup-seq/crop_keep_median}. Playback speed-up was the reverse: TMK+PDQF and the 256-bit ISCC Video-Code still matched all \val{fp:sec/video/magnitude/TMK+PDQF/speed_never} clips at twice the speed (the 64-bit code \val{fp:sec/video/magnitude/ISCC-Video-64/speed_never}), because dropping frames leaves the remaining frames' hashes intact, while the sequence-matched neural descriptors broke at a median factor of \val{fp:sec/video/magnitude/DINOv2-S-seq/speed_median}. No video fingerprint survived both a small crop and a white-box attack.

\subsection{Determinism and Cost}
\label{sec:fp-res-cost}
\begin{figure}[!t]
\centering
\includegraphics[width=\columnwidth]{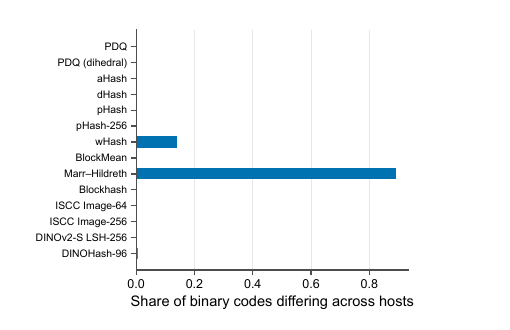}
\caption{Share of binary image codes that differ between an x86/A10G host and an arm64/M5 Pro laptop on byte-identical inputs. Float descriptors differ only in rounding ($1-\cos \le 2\times10^{-4}$).}
\label{fig:fp-stability}
\end{figure}

\paragraph{Determinism across hosts.} A registry compares a fingerprint computed at signing time with one computed at lookup time, possibly on different hardware. We fingerprinted the same \val{fp:stab/n} PNG files (100 registered originals and their JPEG-75 copies, byte-identical on both hosts) on the AMD EPYC and NVIDIA A10G host and on an Apple M5 Pro laptop (Fig.~\ref{fig:fp-stability}). PDQ, the imagehash DCT and difference hashes, BlockMean, Blockhash and the ISCC codes were bit-identical on every image, and the float descriptors agreed to $1-\cos \le$ \val{fp:stab/SSCD-large/cosmax}. Two classical hashes were not deterministic: the Marr--Hildreth hash differed on \val{fp:stab/MarrHildreth/differ/pct}\% of images by up to \val{fp:stab/MarrHildreth/bitsmax} of its 576 bits, enough to push \val{fp:stab/MarrHildreth/below} of the \val{fp:stab/n} self-comparisons below its own match threshold, and wHash differed on \val{fp:stab/wHash-64/differ/pct}\% by up to \val{fp:stab/wHash-64/bitsmax} of 64 bits. Both depend on floating-point filtering whose result differs between processor architectures; a service that fingerprints on one platform and verifies on another should avoid them or pin the implementation.

\paragraph{Extraction cost.} We timed extraction separately on the otherwise idle A10G host (AMD EPYC 7R32), because the timings recorded during the benchmark include contention from dozens of concurrent workers. Each image method fingerprinted the same 256 registered originals at 1024 pixels on the long side, CPU methods on a single thread and learned methods on the GPU in batches of 64 including the host-to-device copy. The hashes cost between \val{fp:lat/image/BlockMean} (BlockMean) and \val{fp:lat/image/wHash-64}\,ms (wHash) per image on one CPU thread, with PDQ at \val{fp:lat/image/PDQ}\,ms and the ISCC Image-Code at \val{fp:lat/image/ISCC-Image-64}\,ms; the learned descriptors cost \val{fp:lat/image/OpenCLIP-B32}--\val{fp:lat/image/ISC21-1st}\,ms on the GPU (DINOv2-S \val{fp:lat/image/DINOv2-S}, SSCD-R50 \val{fp:lat/image/SSCD-mixup}, ISC21 \val{fp:lat/image/ISC21-1st}). None of these costs matters for a signing service: on the same on-demand instance, fingerprinting a million images costs about US\$\val{fp:lat/image/PDQ/usd1m} with PDQ on its eight vCPUs, US\$\val{fp:lat/image/DINOv2-S/usd1m} with DINOv2-S and US\$\val{fp:lat/image/ISC21-1st/usd1m} with ISC21 on its GPU, before decoding. Storage differs more: PDQ, the ISCC codes and the keyed projection need 8--32 bytes per asset, a float DINOv2-S descriptor 1.5\,kB. Audio fingerprints ran at \val{fp:lat/audio/audfprint}--\val{fp:lat/audio/ISCC-Audio-256}\,ms per second of audio (CLAP \val{fp:lat/audio/CLAP}\,ms on the GPU), and video fingerprints at \val{fp:lat/video/TMK+PDQF}\,ms (TMK+PDQF) to \val{fp:lat/video/vPDQ}\,ms (vPDQ) per second of video on one CPU thread and \val{fp:lat/video/DINOv2-S-seq}\,ms for frame-level DINOv2-S, dominated in every case by decoding and frame sampling. We did not time registry search at scale; exact search over our \val{fp:count/image/nref}-item image index was not the bottleneck at any point.

\section{Watermarks and Fingerprints Together}
\label{sec:together}
The registered assets of the fingerprinting track were also watermarked with one openly licensed model per modality from the watermarking track, and the marked copies went through the fingerprinting track's transformations (Section~\ref{sec:fp-method}). This measures, on identical media and attacks, the coverage of each family's decision rule and their complementarity. The union of their successes describes combined coverage; a deployed watermark-first cascade would also need to handle incorrect and ambiguous keys.

\begin{figure*}[!t]
\centering
\includegraphics[width=\textwidth]{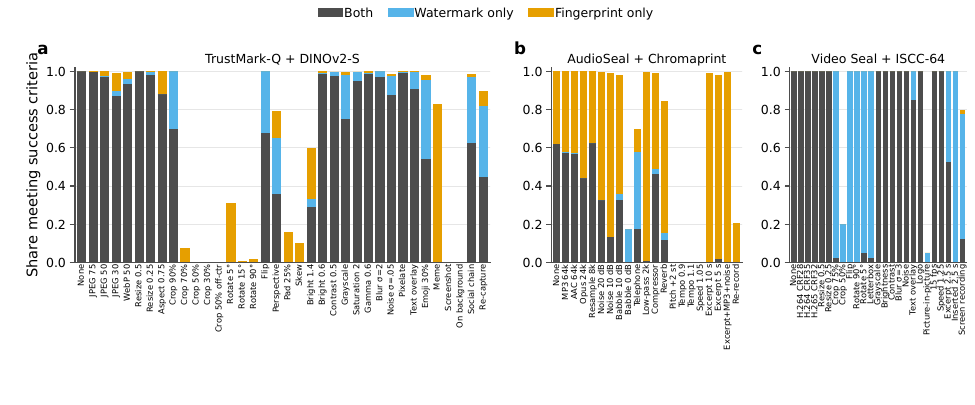}
\caption{Share of attacked copies satisfying only the watermark criterion (blue), only the fingerprint criterion (orange), or both (grey), per transformation. Watermark success means exact-key recovery for images and audio, and expected-key verification for video; the video panel does not measure unique registry identification. (a) Images: TrustMark-Q and DINOv2-S, with the fingerprint threshold calibrated to pair-level FPR $10^{-7}$. (b) Audio: AudioSeal and Chromaprint. (c) Video: Video Seal and the ISCC Video-Code. Audio and video fingerprint thresholds are calibrated to query-level FPR 1\%; achieved held-out rates are reported separately.}
\label{fig:fp-wm}
\end{figure*}

Table~\ref{tab:fp-wm} and Fig.~\ref{fig:fp-wm} combine each modality's MIT-licensed watermark with its fingerprints on the same attacked copies of watermarked assets. The two satisfy their respective success criteria on different queries. For images, TrustMark yields its key for \val{fp:wm/image/DINOv2-S/wm/pct}\% of attacked copies and DINOv2-S identifies \val{fp:wm/image/DINOv2-S/fp/pct}\%; the union of these successes covers \val{fp:wm/image/DINOv2-S/either/pct}\%, and \val{fp:wm/image/ISC21-1st/either/pct}\% with the ISC21 descriptor. The watermark alone survives additive noise, emoji and text overlays, perspective and the re-capture chain better than DINOv2-S; the fingerprint alone survives the meme layout, padding, a 5-degree rotation and skew, which move the watermark out of TrustMark's decoding grid. Neither survives large crops, rotations of 15 degrees or more, screenshots or overlays onto another photograph (Fig.~\ref{fig:fp-wm}a). For audio the fingerprint covers more queries: AudioSeal returned the exact 16-bit key for \val{fp:wm/audio/Chromaprint/wm/pct}\% of queries (\val{fp:wm/audio/keyok/unique/pct}\% if recoveries of the \val{fp:wm/audio/n_colliding} assets that share a key are not counted as identification) and Chromaprint identified \val{fp:wm/audio/Chromaprint/fp/pct}\%, and the watermark added a useful share only under telephone filtering and loud babble; nothing survived pitch, tempo or speed changes. For video expected-key watermark verification covers more queries: Video Seal passed that test for \val{fp:wm/video/ISCC-Video-64/wm/pct}\% of queries, including the crops, flips, rotations, letterboxing and insertions that break every video fingerprint, and the fingerprints added almost nothing (a few screen recordings); picture-in-picture defeated both. These union rates describe complementary coverage, not the measured accuracy of a deployed cascade; the video result does not establish unique registry-ID recovery.

\paragraph{Watermarking moves fingerprints.} Embedding the watermark is itself a transformation. For \val{fp:wm/image/ISCC-Image-64/drift/pct}\% of the watermarked images the ISCC Image-Code of the marked image no longer matched the unmarked original at the operating threshold, and for \val{fp:wm/image/SSCD-mixup/drift/pct}\% with SSCD, against \val{fp:wm/image/PDQ/drift/pct}\% for PDQ and \val{fp:wm/image/DINOv2-S/drift/pct}\% for DINOv2-S. We attribute the ISCC case to the same step that makes it survive padding: on product shots with white backgrounds, a faint residual in the border changes what the uniform-border trim removes, and the security experiments show the same sensitivity to imperceptible perturbations. A registry that stores the fingerprint of the unmarked original therefore loses matches: ISCC's detection of attacked marked copies fell from \val{fp:wm/image/ISCC-Image-64/fp/pct}\% to \val{fp:wm/image/ISCC-Image-64/unmarkedreg/fp/pct}\% when the registry held the unmarked fingerprint. The fingerprint must be computed on the published, watermarked asset.

\paragraph{False keys.} On never-registered, unmarked images TrustMark's error-correcting decoder returned a valid codeword for \val{fp:wm/image/falsekey/pct}\% of \val{fp:wm/image/falsekey/n} attacked queries (a detection; these decodes were not compared with the registered keys), in line with its blind false-positive rates in the watermarking track (Section~\ref{sec:wm-res-fpr}); AudioSeal's detector fired on \val{fp:wm/audio/falsekey/k} of \val{fp:wm/audio/falsekey/n} attacked queries from \val{fp:wm/audio/falsekey/nsrc} never-registered tracks, and that detection's message equalled none of the \val{fp:wm/audio/nkeys} distinct registered keys; and Video Seal, whose key needs agreement on far more bits, matched a registered key for \val{fp:wm/video/falsekey/pct}\%. A cascade must check a decoded key against the registry before trusting it, exactly as it thresholds a fingerprint match.

\begin{table}[!t]
\centering
\footnotesize
\caption{Watermark and fingerprint on the same attacked copies of watermarked assets, pooled over transformations. WM: exact-key recovery for TrustMark and AudioSeal, but expected-key verification for Video Seal; the latter does not establish unique registry identification. FP: correct fingerprint identification at a threshold calibrated to pair-level FPR $10^{-7}$ for images or query-level FPR 1\% for audio and video; achieved held-out rates are reported separately. Both and Either are the intersection and union of these success criteria, not measured cascade accuracy. Drift: share of watermarked assets whose own fingerprint falls below the threshold against the unmarked original.}
\label{tab:fp-wm}
\setlength{\tabcolsep}{3pt}
\begin{tabular}{llccccc}
\toprule
Watermark & Fingerprint & WM & FP & Both & Either & Drift \\
\midrule
TrustMark-Q & DINOv2-S & 0.62 & 0.61 & 0.55 & 0.69 & 0.06 \\
 & DINOv2-S LSH-256 & 0.62 & 0.51 & 0.47 & 0.66 & 0.21 \\
 & ISC21 1st\reftier & 0.62 & 0.76 & 0.60 & 0.78 & 0.09 \\
 & ISCC Image-64 & 0.62 & 0.47 & 0.43 & 0.66 & 0.62 \\
 & PDQ & 0.62 & 0.49 & 0.47 & 0.64 & 0.00 \\
 & SSCD R50\reftier & 0.62 & 0.68 & 0.58 & 0.72 & 0.18 \\
\addlinespace
AudioSeal & CLAP & 0.24 & 0.37 & 0.12 & 0.49 & -- \\
 & Chromaprint & 0.24 & 0.67 & 0.21 & 0.71 & -- \\
 & ISCC Audio-64 & 0.24 & 0.40 & 0.16 & 0.48 & -- \\
 & audfprint & 0.24 & 0.64 & 0.23 & 0.65 & -- \\
\addlinespace
Video Seal & ISCC Video-64 & 0.92 & 0.62 & 0.62 & 0.92 & -- \\
 & SSCD frames\reftier & 0.92 & 0.54 & 0.54 & 0.92 & -- \\
 & TMK+PDQF & 0.92 & 0.64 & 0.64 & 0.92 & -- \\
 & vPDQ & 0.92 & 0.58 & 0.58 & 0.92 & -- \\
\bottomrule
\end{tabular}

\end{table}

\section{Discussion}
\label{sec:discussion}

\paragraph{What a soft-binding service should deploy.} No single soft binding serves every purpose, and the results argue for a layered design with a different lead family per modality. For images, a watermark and a fingerprint recover about the same share of attacked copies but different ones, so the union of their successes covers \val{fp:wm/image/DINOv2-S/either/pct}\% of attacked copies with TrustMark and DINOv2-S, against \val{fp:wm/image/DINOv2-S/wm/pct}\% and \val{fp:wm/image/DINOv2-S/fp/pct}\% for either alone (Section~\ref{sec:together}). Among openly licensed watermarks, PixelSeal at a strength just above its robustness threshold is the best balance of perceptibility, robustness and speed; its margin at 0.15 is small on small images, so a deployment expecting heavier compression should move towards the default of 0.2 (Section~\ref{sec:wm-res-image}). Among openly licensed fingerprints, a DINOv2 embedding recovers the most copies and is the only one that survives large crops and rotations; its keyed 256-bit projection costs a few points of detection, and stores in the same 32 bytes as PDQ. In the evaluated inversion experiment, a decoder trained under a different key did not identify the source above chance (Section~\ref{sec:fp-res-security}); this does not establish that no attacker can invert it. The registered open standard, the ISCC Image-Code, is precise but loses roughly half of the transformed copies. For audio, fingerprints cover more queries: Chromaprint and audfprint recover far more attacked copies than the 16-bit AudioSeal key, whose base model is nevertheless the best watermark operating point. For video, expected-key watermark verification covers more queries: in the combination experiment Video Seal passed that test after the crops, flips and insertions that break every open video fingerprint, and in the watermarking track PixelSeal at 0.4 was the best-balanced video watermark. The strongest image copy detectors, SSCD and the ISC21 winner, would add a further 6--10 points but cannot be shipped as released, because their weights were trained on data licensed for non-commercial research.

\paragraph{The decision rule decides the false positives.} For both families the false-match rate belongs to the complete decision rule, not to the model. TrustMark's error-correcting decoder reports a valid codeword on several percent of unmarked images, yet none of the same decodes reached the expected-payload threshold on any of the \val{wm:fpr/image/TrustMark-Q/own_src_n} unmarked images (Section~\ref{sec:wm-res-fpr}); a provenance verifier resolves a soft binding against a manifest and can apply that test. The combination experiment shows the same for keys decoded from attacked, never-registered content (Section~\ref{sec:together}): a decoded key must be checked against the registry exactly as a fingerprint score is thresholded. Payload size limits what can be shown. With 16 bits, even an exact match leaves a chance of $1.5\times10^{-5}$ per key, so the 16-bit audio models rest on their blind detectors, whose observed bound over \val{wm:fpr/audio/WavMark/blind_src_n} clips does not establish $10^{-3}$. For fingerprints, $Np$ is the expected number of false matches per negative query for $N$ references and average pair-level false-match probability $p$, and bounds the probability of at least one false match from above. A pair-level rate at most $10^{-9}$ would therefore suffice for a 1\% query-level bound at $10^{7}$ assets; this operating point is below the empirically resolved range here. We expected a second, geometric verification stage to supply the missing precision, and measured it; under a matched calibration false-binding constraint it did not raise detection in the evaluated image pipelines, because the remaining false matches are near-copies of registered designs that also match geometrically (Section~\ref{sec:fp-res-fpr}). What alignment against the stored original does provide is localization of an edit, most of which survives when the original must first be retrieved.

\paragraph{Catalogues are full of near-copies.} The largest effect in the fingerprinting track is not the choice of method but the treatment of variants: when recoloured or re-mounted versions of one design count as distinct assets, every method's calibrated detection falls to 0.24 or less. Learned descriptors bind such variants far more readily than hashes, which is correct if the question is ``is this the same artwork'' and wrong if it is ``is this the asset this manifest was signed for''. A soft-binding service must decide which question it answers, group variants of a design under one registry entry if it answers the first, and treat a match as a pointer to candidate manifests rather than as an identification.

\paragraph{Neither family is a security boundary.} White-box attacks evaded every learned fingerprint with perturbations of a few grey levels, surrogate gradients evaded the DCT hashes, targeted forgeries bound unrelated images to registered assets, and pitch, tempo or small crops broke the audio and video fingerprints. Regeneration and adversarial attacks are known to remove or evade pixel-level watermarks \cite{zhao2024removable, jiang2023evading}; our watermarking track measured routine processing only. A soft-binding match should therefore be presented as a pointer to a candidate manifest that must be confirmed, never as evidence that content is authentic. The value written into a \texttt{c2pa.soft-binding} assertion is disclosed to everyone who reads the manifest, and our inversion decoder recovered recognisable thumbnails from unkeyed descriptors and hashes, so services should keep raw descriptors server-side and publish only keyed or opaque identifiers.

\paragraph{Fingerprint the published asset.} Embedding a watermark is itself a transformation. It moved the ISCC Image-Code of \val{fp:wm/image/ISCC-Image-64/drift/pct}\% of images past its own match threshold, and a registry holding the fingerprint of the unmarked original lost matches accordingly (Section~\ref{sec:together}). A service that does both must compute the fingerprint on the watermarked asset it publishes.

\paragraph{Reproducibility across hardware and runs.} Provenance claims should not depend on which machine checks them. A watermark decoded on an x86 host lost payload bits on an arm64 host; given identical RGB arrays both hosts agreed to within floating-point noise, so the cause is the platform-specific conversion from YUV to RGB that feeds the detector (Section~\ref{sec:wm-res-stability}). Converting colour in a fixed, portable implementation, or running detectors on YUV input, would remove the effect. Two classical image hashes, Marr--Hildreth and wHash, were likewise not bit-identical across architectures. Reproducing a benchmark also needs more pinned than its corpus: our two runs of the audio fingerprint track drew different background music for the babble transformations, because the pool came from incidental surplus downloads, and the results moved accordingly until the pool was pinned.

\paragraph{Limitations.} The watermark corpus is small by the standards of model training (85 images, 138 audio inputs from \val{wm:audio/n_sources} clips, six 2.5-s video clips); it separates the methods on the gate and on false positives, but the video comparisons are descriptive, FLIP and ColorVideoVDP were computed on subsets, and PESQ is reported on speech only. ABO, the fingerprint image corpus, is a product-photo catalogue whose white backgrounds and repeated designs make some things harder and others easier than natural photographs, as the comparison with DISC21 shows. The fingerprint audio negatives are music only. The video calibration set contains \val{fp:count/video/neg_cal} queries from \val{fp:video/cal_sources} negative clips and supports reporting the query-level 1\% and pair-level $10^{-4}$ targets under our resolution rule, with limited precision. The achieved held-out query-level false-match rates exceed the 1\% calibration target. Source-clip resampling accounts for transformations of the same clip, but does not by itself account for dependence between clips from the same film. The security experiments used small samples and simple attacks, so they bound attack difficulty from above. Latency was measured on one GPU type, and cost estimates use on-demand prices at one date and assume full utilisation. Strength settings were chosen around each method's default and do not trace a full rate--distortion curve. The licence status of several models changed during 2024--2026; we report the status recorded at the time of measurement.

\section{Conclusion}
\label{sec:conclusion}
We compared openly available watermarks and fingerprints for images, audio and video as soft bindings for content provenance, under one protocol, on licence-audited public media and at calibrated false-match rates. For images, PixelSeal and a DINOv2 embedding are the strongest shippable watermark and fingerprint, and they fail on different copies; fingerprints cover more attacked audio copies, while expected-key watermark verification covers more video copies. The strongest fingerprints cannot be shipped as released, and the one open fingerprint on the C2PA list, the ISCC code, is precise but loses half of the transformed copies. Five findings concern the design of a soft-binding service more than the choice of a model. False matches are a property of the decision rule, so decoded keys must be checked against the expected payload or the registry. Catalogue variants dominate the fingerprint false-match budget, and no detection improvement from geometric verification was observed under a matched calibration false-binding constraint in the evaluated image pipelines. Neither family resists an adversary who knows the method, and unkeyed fingerprints leak content when published. Watermarking moves fingerprints, so the published asset must be the one fingerprinted. And colour conversion that differs across platforms changes decoded watermark bits.

\section*{Data and Code Availability}
The benchmark scripts for both tracks, the corpus manifests with source URLs and SHA-256 hashes, the weight and source lockfiles, the per-configuration, per-method and per-query results, and the scripts that generate every table, figure and cited number in this paper are publicly available at \url{https://github.com/Original-Pictures/soft-bindings-benchmark} (release v1.0, which also contains the LaTeX source of this version). Rebuilding the paper from that release regenerates every number from the committed results. Corpus media are not redistributed; each item can be retrieved from its original source and verified against the manifest. DISC21 was used under its non-commercial research licence and ABO under CC BY 4.0.

\section*{Competing Interests}
The authors are employed by Original Pictures Technologies, Inc., which develops content-provenance software. The company did not develop any of the evaluated watermarking or fingerprinting methods, holds no licence or commercial relationship with the authors of those methods, and did not constrain the choice of methods or the reporting of results.

\section*{Use of Generative AI}
The authors used Claude (Anthropic), a large language model, in preparing this work. Claude drafted the manuscript text; wrote the benchmark orchestration, analysis and figure-generation code; searched for and verified the bibliographic references against publisher, Crossref, arXiv and repository records; and produced the figures from the recorded results. Claude also assisted with the revision after internal review, including the re-analysis of the recorded results described in the text. The authors designed the study, ran and supervised the measurements, reviewed the analysis code and its outputs, reviewed and edited all text, and take full responsibility for the content of this article. Every number, table and figure in the article is generated by scripts from the recorded results (Data and Code Availability). No AI system is listed as an author.

\bibliographystyle{unsrtnat}
\bibliography{refs}

\clearpage
\appendix
\section{Supplementary Tables}
\label{app:tables}
\begingroup
\let\supplementarytable\table
\renewcommand{\table}[1][]{\supplementarytable[H]}
\renewenvironment{table*}[1][]{\supplementarytable[H]}{\endtable}
\begin{table*}[!t]
\centering
\footnotesize
\caption{Quality means with bootstrap 95\% confidence intervals (10{,}000 resamples of sources). Columns: image PSNR (dB), LPIPS, FLIP; audio SI-SNR (dB), PESQ on speech, STOI; video PSNR (dB), LPIPS, VMAF.}
\label{tab:wm-ci}
\setlength{\tabcolsep}{3.4pt}
\begin{tabular}{llrlll}
\toprule
Modality & Configuration & $n$ & Metric 1 & Metric 2 & Metric 3 \\
\midrule
Image & PixelSeal (0.1) & 85 & 51.6 [51.1, 52.1] & 0.0006 [0.0005, 0.0007] & 0.0158 [0.0146, 0.0171] \tiny($n$=60) \\
Image & Video Seal (0.1) & 85 & 50.5 [50.1, 51.0] & 0.0006 [0.0006, 0.0007] & 0.0184 [0.0170, 0.0199] \tiny($n$=60) \\
Image & PixelSeal (0.125) & 85 & 49.8 [49.3, 50.2] & 0.0009 [0.0008, 0.0010] & 0.0200 [0.0186, 0.0214] \tiny($n$=60) \\
Image & InvisMark & 85 & 50.1 [49.6, 50.6] & 0.0017 [0.0015, 0.0019] & 0.0219 [0.0202, 0.0237] \tiny($n$=60) \\
Image & PixelSeal (0.15) & 85 & 48.3 [47.8, 48.7] & 0.0012 [0.0011, 0.0014] & 0.0238 [0.0223, 0.0253] \tiny($n$=60) \\
Image & TrustMark-P (0.8) & 85 & 50.5 [50.3, 50.8] & 0.0017 [0.0014, 0.0020] & 0.0220 [0.0215, 0.0226] \tiny($n$=60) \\
Image & TrustMark-P (1) & 85 & 49.4 [49.1, 49.7] & 0.0024 [0.0020, 0.0028] & 0.0240 [0.0233, 0.0247] \tiny($n$=60) \\
Image & PixelSeal (0.2) & 85 & 45.9 [45.5, 46.4] & 0.0020 [0.0018, 0.0022] & 0.0302 [0.0285, 0.0320] \tiny($n$=60) \\
Image & Video Seal (0.2) & 85 & 44.8 [44.4, 45.3] & 0.0021 [0.0019, 0.0024] & 0.0339 [0.0318, 0.0361] \tiny($n$=60) \\
Image & TrustMark-Q (1) & 85 & 42.1 [41.7, 42.5] & 0.0027 [0.0024, 0.0029] & 0.0401 [0.0387, 0.0415] \tiny($n$=60) \\
Image & TrustMark-P (1.5) & 85 & 46.8 [46.5, 47.2] & 0.0049 [0.0042, 0.0057] & 0.0293 [0.0283, 0.0303] \tiny($n$=60) \\
Image & TrustMark-B (1) & 85 & 41.0 [40.6, 41.4] & 0.0029 [0.0026, 0.0032] & 0.0426 [0.0413, 0.0440] \tiny($n$=60) \\
Image & TrustMark-Q (1.2) & 85 & 40.7 [40.3, 41.1] & 0.0037 [0.0034, 0.0040] & 0.0448 [0.0432, 0.0464] \tiny($n$=60) \\
Image & TrustMark-C (1) & 85 & 40.0 [39.6, 40.5] & 0.0035 [0.0032, 0.0038] & 0.0467 [0.0451, 0.0484] \tiny($n$=60) \\
Image & DWT-DCT & 85 & 39.3 [38.9, 39.6] & 0.0212 [0.0169, 0.0261] & 0.0288 [0.0276, 0.0300] \tiny($n$=60) \\
Image & TrustMark-Q (1.3) & 85 & 40.0 [39.6, 40.5] & 0.0043 [0.0039, 0.0046] & 0.0471 [0.0454, 0.0489] \tiny($n$=60) \\
Image & WAM (1) & 85 & 45.3 [44.9, 45.7] & 0.0099 [0.0087, 0.0113] & 0.0432 [0.0408, 0.0457] \tiny($n$=60) \\
Image & TrustMark-Q (1.4) & 85 & 39.4 [39.0, 39.9] & 0.0049 [0.0045, 0.0053] & 0.0494 [0.0476, 0.0513] \tiny($n$=60) \\
Image & ChunkySeal (0.2) & 85 & 44.4 [44.0, 44.8] & 0.0090 [0.0080, 0.0101] & 0.0477 [0.0451, 0.0504] \tiny($n$=60) \\
Image & PixelSeal (0.4) & 85 & 40.1 [39.6, 40.6] & 0.0077 [0.0068, 0.0087] & 0.0514 [0.0486, 0.0544] \tiny($n$=60) \\
Image & TrustMark-Q (1.5) & 85 & 38.9 [38.4, 39.3] & 0.0056 [0.0052, 0.0061] & 0.0517 [0.0497, 0.0537] \tiny($n$=60) \\
Image & Video Seal (0.4) & 85 & 38.9 [38.5, 39.4] & 0.0082 [0.0072, 0.0094] & 0.0578 [0.0544, 0.0614] \tiny($n$=60) \\
Image & DWT-DCT-SVD & 85 & 39.1 [38.8, 39.5] & 0.0323 [0.0279, 0.0371] & 0.0514 [0.0491, 0.0538] \tiny($n$=60) \\
Image & WAM (2) & 85 & 39.4 [39.1, 39.8] & 0.0367 [0.0326, 0.0408] & 0.0708 [0.0671, 0.0745] \tiny($n$=60) \\
Image & RivaGAN & 60 & 40.5 [40.5, 40.6] & 0.0643 [0.0530, 0.0764] & 0.0774 [0.0743, 0.0805] \\
\midrule
Audio & SilentCipher 44.1k & 138 (46) & 44.1 [43.0, 45.0] & 4.589 [4.579, 4.598] \tiny($n$=120) & 0.9987 [0.9970, 0.9997] \\
Audio & AudioSeal streaming (0.5) & 138 (46) & 32.7 [31.0, 34.4] & 4.546 [4.518, 4.569] \tiny($n$=120) & 0.9999 [0.9999, 1.0000] \\
Audio & AudioSeal base (1) & 138 (46) & 27.0 [26.1, 27.8] & 4.444 [4.423, 4.463] \tiny($n$=120) & 0.9967 [0.9952, 0.9978] \\
Audio & AudioSeal streaming (1) & 138 (46) & 26.6 [25.0, 28.3] & 4.390 [4.332, 4.437] \tiny($n$=120) & 0.9998 [0.9997, 0.9998] \\
Audio & Perth & 138 (46) & 15.8 [15.5, 16.1] & 4.383 [4.351, 4.413] \tiny($n$=120) & 0.9931 [0.9922, 0.9936] \\
Audio & AudioSeal streaming (1.1) & 138 (46) & 25.8 [24.2, 27.5] & 4.358 [4.295, 4.409] \tiny($n$=120) & 0.9997 [0.9996, 0.9998] \\
Audio & WavMark & 138 (46) & 35.9 [35.0, 36.5] & 4.149 [4.084, 4.211] \tiny($n$=120) & 0.9932 [0.9873, 0.9973] \\
\midrule
Video & PixelSeal (0.2) & 6 & 46.0 [44.3, 47.5] & 0.0035 [0.0021, 0.0053] & 95.5 [92.7, 97.9] \\
Video & Video Seal (0.2) & 6 & 46.2 [43.4, 49.2] & 0.0035 [0.0021, 0.0057] & 94.0 [91.0, 97.1] \\
Video & PixelSeal (0.4) & 6 & 40.1 [38.4, 41.7] & 0.0130 [0.0079, 0.0196] & 86.0 [79.9, 91.7] \\
Video & ChunkySeal (0.2) & 6 & 44.5 [43.3, 45.6] & 0.0151 [0.0079, 0.0270] & 96.4 [94.1, 98.8] \\
Video & Video Seal (0.4) & 6 & 40.3 [37.5, 43.3] & 0.0131 [0.0080, 0.0210] & 82.1 [75.4, 88.3] \\
Video & Video Seal (0.7) & 6 & 35.5 [32.7, 38.5] & 0.0361 [0.0245, 0.0508] & 66.6 [57.7, 75.8] \\
Video & Video Seal (1.1) & 6 & 31.6 [28.8, 34.7] & 0.0781 [0.0599, 0.0977] & 54.6 [45.1, 66.0] \\
\bottomrule
\end{tabular}
\par\vspace{2pt}\parbox{\linewidth}{\raggedright\scriptsize Percentile intervals. $n$: items (independent sources in parentheses). Audio sources are the 46 clips, each rendered at three sample rates; the three renderings of a clip are resampled together. A metric computed on a subset shows its own $n$ (FLIP on 60 images, PESQ on 120 speech inputs).}
\end{table*}

\begin{table*}[!t]
\centering
\scriptsize
\caption{Pooled results with 95\% intervals from a bootstrap over sources: registered sources are drawn with all their transformed queries, and calibration and held-out negative sources likewise, with the threshold recomputed in every replicate. FPR$_q$: share of held-out never-registered queries whose best score passes the query-level 1\% threshold (out-of-sample check of the calibration). TPR$_{p,10^{-7}}$ is not resolved by the audio and video calibration sets (--).}
\label{tab:fp-ci}
\setlength{\tabcolsep}{3pt}
\begin{tabular}{llccccrr}
\toprule
Mod. & Method & R@1 & TPR$_{q,1\%}$ & FPR$_{q,1\%}$ (test) & TPR$_{p,10^{-7}}$ & $n_+$ (sources) & $n_-$ (sources) \\
\midrule
image & PDQ & 0.579 \tiny[0.58, 0.58] & 0.495 \tiny[0.49, 0.50] & 0.010 \tiny[0.01, 0.02] & 0.495 \tiny[0.48, 0.50] & 72{,}000 (2{,}000) & 26{,}820 (745) \\
image & PDQ (dihedral) & 0.623 \tiny[0.62, 0.63] & 0.549 \tiny[0.54, 0.55] & 0.011 \tiny[0.00, 0.02] & 0.546 \tiny[0.54, 0.55] & 72{,}000 (2{,}000) & 26{,}820 (745) \\
image & aHash & 0.503 \tiny[0.50, 0.51] & 0.000 \tiny[0.00, 0.00] & 0.000 \tiny[0.00, 0.00] & 0.000 \tiny[0.00, 0.00] & 72{,}000 (2{,}000) & 26{,}820 (745) \\
image & dHash & 0.560 \tiny[0.56, 0.56] & 0.313 \tiny[0.00, 0.32] & 0.011 \tiny[0.00, 0.01] & 0.000 \tiny[0.00, 0.00] & 72{,}000 (2{,}000) & 26{,}820 (745) \\
image & pHash & 0.563 \tiny[0.56, 0.57] & 0.373 \tiny[0.37, 0.38] & 0.003 \tiny[0.00, 0.01] & 0.373 \tiny[0.00, 0.38] & 72{,}000 (2{,}000) & 26{,}820 (745) \\
image & pHash-256 & 0.589 \tiny[0.59, 0.59] & 0.494 \tiny[0.49, 0.50] & 0.013 \tiny[0.01, 0.02] & 0.492 \tiny[0.48, 0.50] & 72{,}000 (2{,}000) & 26{,}820 (745) \\
image & wHash & 0.425 \tiny[0.42, 0.43] & 0.000 \tiny[0.00, 0.00] & 0.000 \tiny[0.00, 0.00] & 0.000 \tiny[0.00, 0.00] & 72{,}000 (2{,}000) & 26{,}820 (745) \\
image & BlockMean & 0.562 \tiny[0.56, 0.57] & 0.337 \tiny[0.00, 0.34] & 0.009 \tiny[0.00, 0.01] & 0.000 \tiny[0.00, 0.00] & 72{,}000 (2{,}000) & 26{,}820 (745) \\
image & Marr–Hildreth & 0.586 \tiny[0.58, 0.59] & 0.362 \tiny[0.34, 0.38] & 0.008 \tiny[0.00, 0.01] & 0.314 \tiny[0.28, 0.35] & 72{,}000 (2{,}000) & 26{,}820 (745) \\
image & ColorMoment & 0.281 \tiny[0.28, 0.29] & 0.000 \tiny[0.00, 0.00] & 0.000 \tiny[0.00, 0.00] & 0.000 \tiny[0.00, 0.00] & 72{,}000 (2{,}000) & 26{,}820 (745) \\
image & Blockhash & 0.560 \tiny[0.55, 0.57] & 0.000 \tiny[0.00, 0.26] & 0.000 \tiny[0.00, 0.01] & 0.000 \tiny[0.00, 0.00] & 72{,}000 (2{,}000) & 26{,}820 (745) \\
image & ISCC Image-64 & 0.558 \tiny[0.55, 0.56] & 0.427 \tiny[0.42, 0.43] & 0.004 \tiny[0.00, 0.01] & 0.427 \tiny[0.31, 0.43] & 72{,}000 (2{,}000) & 26{,}820 (745) \\
image & ISCC Image-256 & 0.561 \tiny[0.56, 0.56] & 0.466 \tiny[0.46, 0.47] & 0.010 \tiny[0.00, 0.02] & 0.458 \tiny[0.45, 0.47] & 72{,}000 (2{,}000) & 26{,}820 (745) \\
image & DINOv2-S & 0.877 \tiny[0.87, 0.88] & 0.625 \tiny[0.61, 0.64] & 0.011 \tiny[0.01, 0.02] & 0.612 \tiny[0.60, 0.63] & 72{,}000 (2{,}000) & 26{,}820 (745) \\
image & DINOv2-B & 0.872 \tiny[0.87, 0.88] & 0.653 \tiny[0.65, 0.66] & 0.012 \tiny[0.01, 0.02] & 0.650 \tiny[0.64, 0.66] & 72{,}000 (2{,}000) & 26{,}820 (745) \\
image & DINOv2-S LSH-256 & 0.832 \tiny[0.83, 0.84] & 0.538 \tiny[0.52, 0.56] & 0.008 \tiny[0.00, 0.02] & 0.519 \tiny[0.50, 0.55] & 72{,}000 (2{,}000) & 26{,}820 (745) \\
image & OpenCLIP B/32 & 0.864 \tiny[0.86, 0.87] & 0.401 \tiny[0.36, 0.43] & 0.006 \tiny[0.00, 0.01] & 0.356 \tiny[0.33, 0.40] & 72{,}000 (2{,}000) & 26{,}820 (745) \\
image & SSCD R50\reftier & 0.929 \tiny[0.93, 0.93] & 0.725 \tiny[0.72, 0.73] & 0.010 \tiny[0.01, 0.02] & 0.712 \tiny[0.71, 0.73] & 72{,}000 (2{,}000) & 26{,}820 (745) \\
image & SSCD RX101\reftier & 0.942 \tiny[0.94, 0.94] & 0.763 \tiny[0.75, 0.78] & 0.010 \tiny[0.01, 0.02] & 0.703 \tiny[0.69, 0.77] & 72{,}000 (2{,}000) & 26{,}820 (745) \\
image & ISC21 1st\reftier & 0.919 \tiny[0.92, 0.92] & 0.762 \tiny[0.75, 0.78] & 0.008 \tiny[0.00, 0.01] & 0.755 \tiny[0.75, 0.77] & 72{,}000 (2{,}000) & 26{,}820 (745) \\
image & DINOHash-96\reftier & 0.769 \tiny[0.76, 0.77] & 0.237 \tiny[0.23, 0.32] & 0.002 \tiny[0.00, 0.01] & 0.237 \tiny[0.23, 0.24] & 72{,}000 (2{,}000) & 26{,}820 (745) \\
audio & Chromaprint & 0.805 \tiny[0.80, 0.81] & 0.704 \tiny[0.68, 0.72] & 0.016 \tiny[0.00, 0.04] & -- & 21{,}000 (1{,}000) & 3{,}990 (190) \\
audio & audfprint & 0.851 \tiny[0.85, 0.85] & 0.660 \tiny[0.53, 0.74] & 0.008 \tiny[0.00, 0.03] & -- & 21{,}000 (1{,}000) & 3{,}990 (190) \\
audio & ISCC Audio-64 & 0.662 \tiny[0.66, 0.67] & 0.409 \tiny[0.21, 0.41] & 0.017 \tiny[0.00, 0.03] & -- & 21{,}000 (1{,}000) & 3{,}990 (190) \\
audio & ISCC Audio-256 & 0.749 \tiny[0.74, 0.75] & 0.454 \tiny[0.31, 0.51] & 0.019 \tiny[0.00, 0.04] & -- & 21{,}000 (1{,}000) & 3{,}990 (190) \\
audio & CLAP & 0.707 \tiny[0.70, 0.72] & 0.386 \tiny[0.33, 0.45] & 0.017 \tiny[0.01, 0.04] & -- & 21{,}000 (1{,}000) & 3{,}990 (190) \\
audio & NMFP\reftier & 0.905 \tiny[0.90, 0.91] & 0.862 \tiny[0.82, 0.88] & 0.018 \tiny[0.00, 0.04] & -- & 21{,}000 (1{,}000) & 3{,}990 (190) \\
video & vPDQ & 0.628 \tiny[0.62, 0.64] & 0.577 \tiny[0.56, 0.59] & 0.032 \tiny[0.00, 0.08] & -- & 3{,}000 (120) & 500 (20) \\
video & TMK+PDQF & 0.778 \tiny[0.77, 0.79] & 0.631 \tiny[0.61, 0.66] & 0.066 \tiny[0.00, 0.16] & -- & 3{,}000 (120) & 500 (20) \\
video & ISCC Video-64 & 0.731 \tiny[0.72, 0.74] & 0.620 \tiny[0.61, 0.65] & 0.062 \tiny[0.00, 0.16] & -- & 3{,}000 (120) & 500 (20) \\
video & ISCC Video-256 & 0.801 \tiny[0.79, 0.81] & 0.652 \tiny[0.63, 0.67] & 0.068 \tiny[0.00, 0.17] & -- & 3{,}000 (120) & 500 (20) \\
video & DINOv2-S frames & 0.872 \tiny[0.85, 0.89] & 0.437 \tiny[0.41, 0.58] & 0.062 \tiny[0.01, 0.15] & -- & 3{,}000 (120) & 500 (20) \\
video & DINOv2-S mean & 0.947 \tiny[0.94, 0.95] & 0.547 \tiny[0.51, 0.73] & 0.032 \tiny[0.00, 0.16] & -- & 3{,}000 (120) & 500 (20) \\
video & SSCD frames\reftier & 0.921 \tiny[0.90, 0.93] & 0.502 \tiny[0.48, 0.56] & 0.032 \tiny[0.00, 0.10] & -- & 3{,}000 (120) & 500 (20) \\
video & SSCD mean\reftier & 0.958 \tiny[0.95, 0.96] & 0.543 \tiny[0.52, 0.78] & 0.036 \tiny[0.00, 0.14] & -- & 3{,}000 (120) & 500 (20) \\
\bottomrule
\end{tabular}

\end{table*}

\clearpage
\endgroup

\end{document}